\documentclass[11pt,a4paper]{article}

\usepackage[utf8]{inputenc}
\usepackage[T1]{fontenc}
\usepackage[margin=1in]{geometry}
\usepackage{amsmath,amssymb,amsthm,mathtools,bm}
\usepackage{booktabs,array,multirow}
\usepackage{graphicx}
\usepackage{tikz}
\usepackage{tikz}
\usetikzlibrary{arrows.meta,positioning,calc,shapes.geometric,fit,backgrounds,decorations.pathreplacing}
\usepackage{xcolor}
\usepackage{enumitem}
\usepackage[numbers,sort&compress]{natbib}
\usepackage[colorlinks=true,linkcolor=blue!50!black,citecolor=green!40!black,urlcolor=blue!50!black]{hyperref}
\usepackage[capitalise,noabbrev]{cleveref}

\theoremstyle{definition}
\newtheorem{definition}{Definition}[section]

\theoremstyle{plain}
\newtheorem{proposition}{Proposition}[section]
\newtheorem{theorem}{Theorem}[section]

\theoremstyle{remark}
\newtheorem{remark}{Remark}[section]

\newcommand{\R}{\mathbb{R}}

\newcommand{\C}{\mathbb{C}}
\newcommand{\G}{\mathcal{G}}
\newcommand{\V}{\mathcal{V}}
\newcommand{\E}{\mathcal{E}}
\newcommand{\Nb}{\mathcal{N}}                 % neighbourhood
\newcommand{\A}{\mathbf{A}}                   % adjacency matrix
\newcommand{\Dg}{\mathbf{D}}                  % degree matrix
\newcommand{\Lap}{\mathbf{L}}                 % graph Laplacian
\newcommand{\Id}{\mathbf{I}}                  % identity
\newcommand{\X}{\mathbf{X}}                   % node feature matrix
\newcommand{\Hm}{\mathbf{H}}                  % hidden representation matrix
\newcommand{\W}{\mathbf{W}}                   % weight matrix
\newcommand{\hv}{\mathbf{h}}                  % node hidden vector
\newcommand{\xv}{\mathbf{x}}                  % node feature vector
\newcommand{\Sym}{S_n}                        % symmetric group
\newcommand{\Perm}{\mathbf{P}}                % permutation matrix

\newcommand{\softmax}{\mathrm{softmax}}
\newcommand{\MSG}{\mathrm{MSG}}
\newcommand{\AGG}{\mathrm{AGG}}
\newcommand{\UPD}{\mathrm{UPD}}
\newcommand{\READOUT}{\mathrm{READOUT}}
\DeclareMathOperator{\diag}{diag}
\DeclareMathOperator{\tr}{tr}

\newcommand{\ket}[1]{\left|#1\right\rangle}
\newcommand{\bra}[1]{\left\langle#1\right|}

\newcommand{\expval}[2]{\left\langle#1\right\rangle_{#2}}
\newcommand{\Tr}{\operatorname{Tr}}
\newcommand{\Hil}{\mathcal{H}}
\newcommand{\Var}{\operatorname{Var}}

\title{Quantum Graph Neural Networks: \\ Architectures, Trainability, and Applications}
\title{Quantum Graph Neural Networks: \\ Architectures, Trainability, and Applications}
\author{
  Zeinab Sharifi$^{1,*}$ \quad Hossein Shirazi$^{2}$ \\[1.5ex]
  \parbox{0.9\textwidth}{\centering\small
    $^{1}$School of Computer Engineering, Iran University of Science and Technology, Tehran, Iran \\
    $^{2}$Faculty of Information and Communications Technology, Malek Ashtar University of Technology, Tehran, Iran} \\[1ex]
  \small
  Emails: \texttt{z\_sharifi@comp.iust.ac.ir} (Z.~Sharifi), \texttt{shirazi@mut.ac.ir} (H.~Shirazi) \\
  $^{*}$Corresponding author
}
\date{\today}

\begin{document}
\maketitle

\begin{abstract}
Quantum graph neural networks (QGNNs) apply parametrized quantum circuits to graph-structured data, with the aim of overcoming limits of classical graph neural networks in scalability, expressivity, and long-range information flow.
Since the first Hamiltonian-based model in 2019, the field has produced many architectures, applications, and claims, but the evidence behind them has not been assessed systematically.
This survey introduces a unified formalism that describes every QGNN as an encoder, a graph-conditioned circuit, observables, and an optional classical head, together with a circuit-level condition for permutation equivariance and a resource model for comparing costs.
We organize the literature into five architecture families (convolutional, equivariant, message-passing and recurrent, Hamiltonian, and hybrid) and compare them by qubit cost, symmetry, scalability, and evidence.
We then review symmetry and geometric inductive biases (permutation, rotation, and Lorentz), the results on expressivity, trainability, noise, and scalability, and applications in high-energy physics, chemistry, finance, healthcare, earth observation, and transportation.
An audit of datasets and evaluation practices leads to a minimum reporting protocol.
Our main finding is that the design space is rich and symmetry is the best-supported inductive bias, whereas the evidence remains thin: results are mostly simulated, on small graphs, at parity with classical models or with fewer parameters, and no advantage has yet been established on tasks over classical graph data.
We identify five open challenges and a research agenda covering provable advantage, equivariant and geometric designs, fault-tolerant implementations, and shared benchmarks.
The coverage extends through 2026 and emphasizes recent work.

\end{abstract}

\tableofcontents
\newpage

% ===========================================================================
\section{Introduction}\label{sec:intro}
% ===========================================================================
Relational structure is fundamental to scientific and industrial data: molecules are graphs of atoms and chemical bonds, particle-collider events are point clouds of detector hits and reconstructed tracks, financial systems are networks of accounts and transactions, and critical infrastructure is governed by topological grids and communication links~\citep{wu2021comprehensive,bronstein2021geometric}.
Over the past decade, graph neural networks (GNNs) have become the standard paradigm for representation learning on such non-Euclidean domains, combining local message passing across edges with permutation-equivariant aggregation to extract node-, edge-, and graph-level predictions~\citep{gilmer2017neural,battaglia2018relational}.
In parallel, quantum machine learning (QML) has emerged as one of the primary workloads envisioned for both noisy intermediate-scale quantum (NISQ) processors and early fault-tolerant architectures, using parametrized quantum circuits (PQCs) and Hamiltonian evolutions as trainable function approximators in exponentially large Hilbert spaces~\citep{preskill2018quantum,benedetti2019parameterized,cerezo2021variational,bharti2022noisy}.

Quantum graph neural networks (QGNNs) sit at the intersection of these two paradigms.
First formalized by \citet{verdon2019quantum} through alternating graph-Hamiltonian evolutions and anticipated by quantum-walk-based graph models~\citep{bai2023learning,dernbach2019quantum}, QGNNs embed graph topology and node or edge attributes into quantum states and quantum gate layouts.
The guiding hypothesis of the field is that the native properties of quantum mechanics---superposition, multipartite entanglement, and unitary time evolution under sparse interaction Hamiltonians---provide a natural computational substrate for relational data that can alleviate the structural bottlenecks of classical GNNs.
Since 2019, the literature has expanded rapidly into a diverse ecosystem of convolutional, equivariant, message-passing, recurrent, physics-inspired, and hybrid quantum--classical architectures, accompanied by theoretical claims regarding expressivity and trainability as well as application studies spanning high-energy physics, quantum chemistry, finance, healthcare, and earth observation.

Yet the rapid proliferation of architectures has outpaced systematic evaluation.
Different sub-communities define QGNNs differently, rely on incompatible data-encoding assumptions, and evaluate models on disparate, heavily downscaled datasets.
Crucially, the interplay between symmetry, barren plateaus, noise-induced decoherence, classical simulability (dequantization), and data-loading overhead is rarely analyzed end to end.
This survey provides a unified mathematical formalism, a structural taxonomy, a quantitative resource and trainability analysis, and a critical audit of the empirical and theoretical evidence accumulated through 2026, clarifying what has been rigorously established, where current architectures fall short, and what is required to achieve a genuine advantage in quantum graph learning.

\subsection{Motivation: Limits of Classical GNNs and the Promise of Quantum Computing}\label{sec:intro-motivation}
% ===========================================================================
% Section 1.1  Motivation: Limits of Classical GNNs and the Promise of Quantum Computing
% (included from main.tex via \input; heading and \label{sec:intro-motivation}
%  live in main.tex)
% ===========================================================================

Graphs describe molecules, particle detectors, road and communication networks, financial transactions, and many other systems, and graph neural networks (GNNs) have become the standard way to learn from them~\citep{wu2021comprehensive,bronstein2021geometric}.
Modern GNNs update each node by exchanging messages with its neighbours~\citep{gilmer2017neural}, a design that is permutation-equivariant and efficient on moderately sized graphs.
Four limitations of this design motivate the search for alternative computational substrates.

\paragraph{Limits of classical GNNs.}
\emph{Scalability}: the receptive field of a node grows with the number of layers, so memory and time grow quickly on large graphs; neighbour sampling and graph partitioning reduce the cost but do not remove it~\citep{hamilton2017inductive,chiang2019cluster}.
\emph{Over-smoothing}: repeated aggregation drives node representations towards each other, which limits useful depth~\citep{rusch2023survey}.
\emph{Over-squashing}: information from an exponentially growing neighbourhood must pass through fixed-size vectors, which hinders long-range dependencies~\citep{alon2021bottleneck,topping2022understanding}.
\emph{Limited expressivity}: standard message passing cannot distinguish graphs that the 1-dimensional Weisfeiler--Leman test cannot distinguish~\citep{xu2019powerful,morris2019weisfeiler}, and more expressive models are more expensive.
\Cref{sec:classical-gnn} treats these issues in detail.

\paragraph{The promise of quantum computing.}
Quantum hardware offers a different substrate.
A register of $n_q$ qubits has $2^{n_q}$ amplitudes, and entanglement links subsystems in ways that no product structure can represent, which makes it a natural carrier of relational information: edges can be represented by entangling gates, and graph states have an entanglement structure that is exactly the graph (\cref{eq:graphstate}).
Time evolution under a graph Hamiltonian is the canonical workload of quantum simulation~\citep{lloyd1996universal}, and quantum walks give provable, though oracle-based, speedups for some graph problems~\citep{childs2003exponential}.
Since the first QGNN~\citep{verdon2019quantum}, claims have ranged from speedups under data-access assumptions to size-independent circuits and expressivity above the 1-WL limit (\cref{sec:tax-conv,sec:tax-mp}).

\paragraph{A conditional promise.}
The promise is conditional.
Current devices are small and noisy, variational training can suffer from barren plateaus, data loading can cancel a speedup, and many trainable quantum models admit classical surrogates~\citep{cerezo2025does,cerezo2022challenges}.
\Cref{tab:intro-motivation} sets each classical limitation against the quantum hope and the evidence found in this survey.
In short, no QGNN has yet been shown to beat a strong classical GNN on a task over classical graph data, and a critical, structured account of what has and has not been established is needed.

\begin{table}[t]
\centering
\caption{Limitations of classical GNNs, the corresponding quantum hopes, and what the evidence reviewed in this survey shows.}
\label{tab:intro-motivation}
\renewcommand{\arraystretch}{1.3}
\footnotesize
\begin{tabular}{@{}p{2.2cm}p{4.2cm}p{6.0cm}p{1.6cm}@{}}
\toprule
\textbf{Limitation} & \textbf{Quantum hope} & \textbf{What the evidence shows} & \textbf{Where} \\
\midrule
Scalability & compressed encodings; block-encoded graph operators & cost moves to data loading, circuit executions, or classical pre-processing & \cref{sec:scalability,sec:ch-scale} \\
Over-smoothing & unitary, norm-preserving layers & no dedicated study found; not tested against classical remedies & \cref{sec:ch-scale} \\
Over-squashing & entanglement and long-range Hamiltonian interactions & untested; local-gate circuits have the same $L$-hop light cone as message passing & \cref{prop:lightcone,sec:ch-scale} \\
Expressivity & high-dimensional quantum feature maps & universality and WL-level results, but no separation from classical GNNs at equal cost & \cref{sec:universal,sec:ch-theory} \\
\bottomrule
\end{tabular}
\end{table}

\subsection{What Is a QGNN? Definition and Scope}\label{sec:intro-definition}
% ===========================================================================
% Section 1.2  What Is a QGNN? Definition and Scope
% (included from main.tex via \input; heading and \label{sec:intro-definition}
%  live in main.tex)
% ===========================================================================

Informally, a \emph{quantum graph neural network} is a learning model that processes an attributed graph $(\G,\X)$ with parametrized quantum circuits whose parameters are trained against a loss, and in which the graph enters either through the data-encoding step or through the structure of the circuit.
We formalize this in \cref{sec:notation}: a QGNN is a map
\[
  f_{\boldsymbol\theta,\boldsymbol\phi}(\G,\X)=g_{\boldsymbol\phi}\big(\widehat{\expval{O_1}{\psi}},\dots,\widehat{\expval{O_K}{\psi}}\big),\qquad \ket{\psi}=U_{\G}(\boldsymbol\theta)S(\G,\X)\ket{0}^{\otimes n_q},
\]
built from an encoder $S$, a graph-conditioned trainable circuit $U_{\G}$, observables $O_k$ estimated from measurements, and an optional classical head $g_{\boldsymbol\phi}$ (\cref{eq:generic-qgnn}).
The same definition covers fully quantum models, where $g$ is trivial, and hybrid models, where a classical network supplies features or post-processes the measurements.

\paragraph{Scope.}
\begin{figure}[t]
    \centering
    \includegraphics[width=0.55\textwidth]{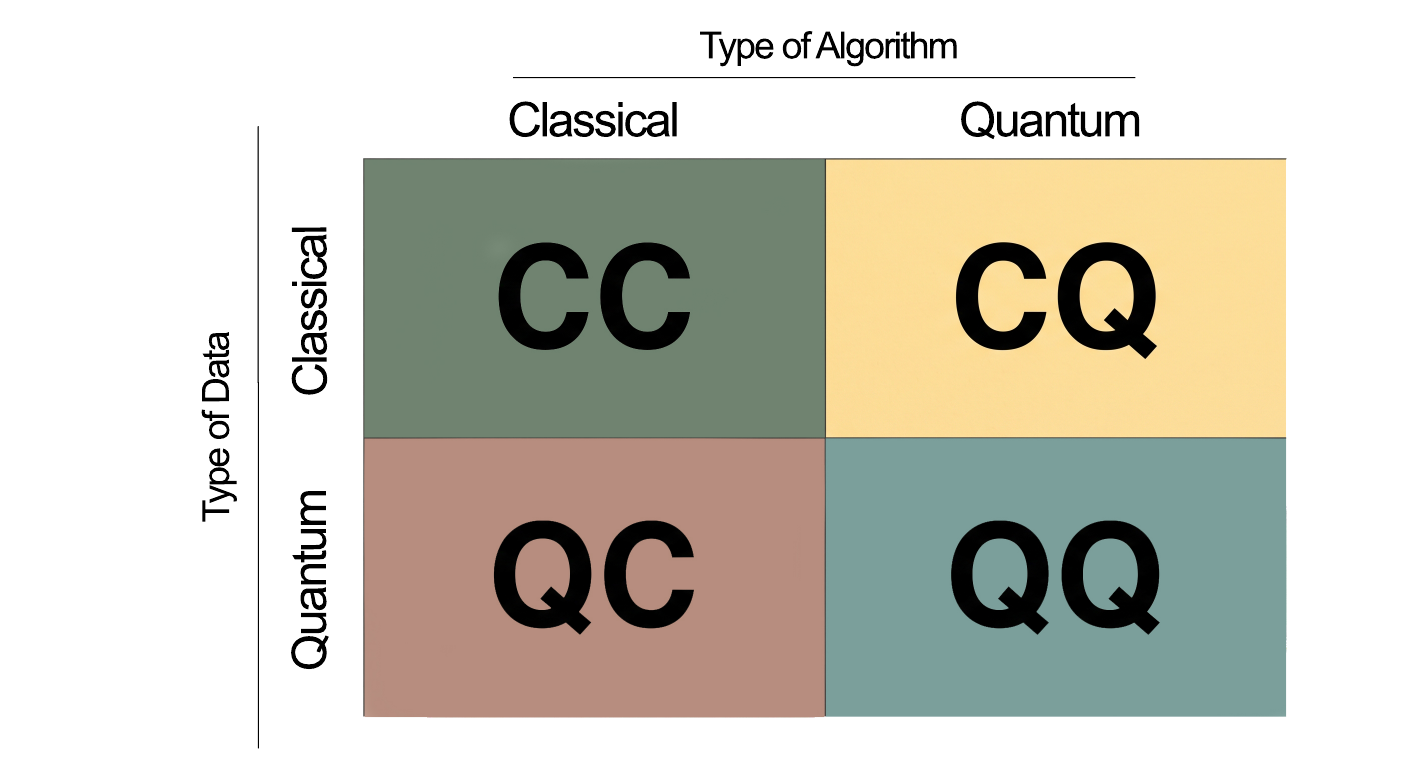}
    \caption{The four paradigms of combining machine learning and quantum computing, categorized by data type and algorithm type. This survey focuses primarily on the \textbf{CQ} regime (classical attributed graphs processed by parametrized quantum circuits), while treating \textbf{CC} (quantum-inspired graph models) and \textbf{QQ} (graph-structured quantum data) as historical and theoretical context (\cref{tab:scope}).}
    \label{fig:qml-taxonomy}
\end{figure}
\Cref{tab:scope} states what the survey covers.
We include models that apply parametrized circuits to attributed graphs, whether fully quantum or hybrid.
We treat three groups as context: quantum-inspired models that borrow quantum dynamics but run on classical hardware, models whose inputs are quantum states on graph vertices, and uses of quantum computers only to produce features for classical models.
We exclude quantum algorithms for classical graph problems that involve no learning, GNNs used to support quantum computing, and quantum neural networks without graph structure; quantum graph kernels are discussed only where they bear on QGNN theory, since earlier surveys cover them~\citep{tang2022from,shayeganfar2025quantum}.
The coverage runs to November 2026 and emphasizes 2025--2026; it is selective, built from prior surveys and targeted searches, and not a systematic review, and we mark preprints as such.

\begin{table}[t]
\centering
\caption{Scope of the survey.}
\label{tab:scope}
\renewcommand{\arraystretch}{1.3}
\footnotesize
\begin{tabular}{@{}p{2.2cm}p{5.6cm}p{5.8cm}@{}}
\toprule
\textbf{Status} & \textbf{Covered} & \textbf{Examples} \\
\midrule
Included & parametrized circuits applied to attributed graphs, fully quantum or hybrid & Hamiltonian, equivariant, convolutional, message-passing, and hybrid QGNNs (\cref{sec:taxonomy}) \\
Context & quantum-inspired models; quantum data on graphs; quantum-computed features & \citet{bai2023learning}; \citet{beer2023quantum}; \citet{thabet2024quantum} (\cref{sec:birth-early}) \\
Excluded & learning-free quantum graph algorithms; GNNs that support quantum computing; graph-agnostic QNNs & walk speedups, QAOA as an optimizer; GNN decoders and compilers; generic variational classifiers \\
\bottomrule
\end{tabular}
\end{table}

\subsection{Contributions of the Survey}\label{sec:intro-contrib}
% ===========================================================================
% Section 1.3  Contributions of the Survey
% (included from main.tex via \input; heading and \label{sec:intro-contrib}
%  live in main.tex)
% ===========================================================================

\paragraph{Relation to earlier surveys.}
Quantum graph learning has been reviewed from several angles (\cref{tab:related-surveys}).
\citet{tang2022from} give a targeted survey from quantum graph computing to quantum graph learning, written before most architectures reviewed here existed.
\citet{ceschini2024graphs} give a critical review of QGNNs that discusses noise, decoherence, and scalability.
\citet{shayeganfar2025quantum} review quantum graph learning with a taxonomy that spans quantum random walks, graph kernels, encoding schemes, and variational QGNNs, with emphasis on materials and image classification.
This survey is complementary: it centres on a unified formalism, on symmetry and resource accounting as organizing themes, and on an audit of the evidence, and it covers the 2025--2026 literature.

\begin{table}[t]
\centering
\caption{Earlier surveys on quantum graph learning and the position of this survey.}
\label{tab:related-surveys}
\renewcommand{\arraystretch}{1.3}
\footnotesize
\begin{tabular}{@{}p{3.1cm}p{1.2cm}p{5.4cm}p{3.7cm}@{}}
\toprule
\textbf{Survey} & \textbf{Year} & \textbf{Focus} & \textbf{Relation to this work} \\
\midrule
\citet{tang2022from} & 2022 & quantum computing for graph tasks, from graph computing to graph learning & predates most architectures reviewed here \\
\citet{ceschini2024graphs} & 2026 (preprint 2024) & critical review of QGNNs: noise, decoherence, scalability & closest in topic; we add formalism, symmetry, evidence audit, 2025--2026 works \\
\citet{shayeganfar2025quantum} & 2025 & taxonomy of quantum graph learning; materials and image classification & broader and application-driven \\
This survey & 2026 & unified formalism, taxonomy, symmetry, evidence audit, research agenda & --- \\
\bottomrule
\end{tabular}
\end{table}

\paragraph{Contributions.}
The main contributions are the following.
\begin{enumerate}[leftmargin=1.8em,itemsep=2pt,topsep=2pt]
  \item \emph{A unified formalism} (\cref{sec:notation}): one definition of a QGNN, a circuit-level condition for permutation equivariance, and a resource model that makes costs comparable across architectures.
  \item \emph{A taxonomy of architectures} (\cref{sec:taxonomy}): five families compared on qubit cost, equivariance, scalability, and the type of evidence reported, with a link back to the earliest models (\cref{sec:birth}).
  \item \emph{A treatment of symmetry} (\cref{sec:symmetry}): permutation, rotation, and Lorentz equivariance, and the strategies for building symmetry into circuits.
  \item \emph{A quantitative analysis} of expressivity, trainability, noise, and scalability (\cref{sec:expr-train-scale}), including budget estimates for graph size under noise.
  \item \emph{An audit of the evidence} (\cref{sec:applications,sec:datasets}): applications across domains, datasets and their sizes against qubit budgets, evaluation practices, and a minimum reporting protocol.
  \item \emph{A research agenda} (\cref{sec:challenges,sec:future}): five open challenges, candidate regimes for provable advantage, and an evidence ladder for judging claims.
\end{enumerate}

\subsection{Organization of the Paper}\label{sec:intro-org}
% ===========================================================================
% Section 1.4  Organization of the Paper
% (included from main.tex via \input; heading and \label{sec:intro-org}
%  live in main.tex)
% ===========================================================================

\begin{figure}[t]
\centering
\begin{tikzpicture}[every node/.style={font=\footnotesize},
  box/.style={draw,rounded corners=2pt,fill=blue!8,text width=3.15cm,align=center,minimum height=1.9cm,inner sep=4pt}]
  \node[box] (a) at (0,0) {\textbf{Foundations}\\[2pt] \ref{sec:background} Background\\ \ref{sec:birth} Origins};
  \node[box] (b) at (3.95,0) {\textbf{Architectures and principles}\\[2pt] \ref{sec:taxonomy} Taxonomy\\ \ref{sec:symmetry} Symmetry\\ \ref{sec:expr-train-scale} Expressivity, training, scale};
  \node[box] (c) at (7.9,0) {\textbf{Evidence}\\[2pt] \ref{sec:applications} Applications\\ \ref{sec:datasets} Datasets and evaluation};
  \node[box,fill=orange!15] (d) at (11.85,0) {\textbf{Outlook}\\[2pt] \ref{sec:challenges} Challenges\\ \ref{sec:future} Future directions\\ \ref{sec:conclusion} Conclusion};
  \foreach \f/\t in {a/b,b/c,c/d}{\draw[->,thick] (\f.east) -- (\t.west);}
\end{tikzpicture}
\caption{Roadmap of the survey. Section numbers refer to the sections of this paper.}
\label{fig:roadmap}
\end{figure}
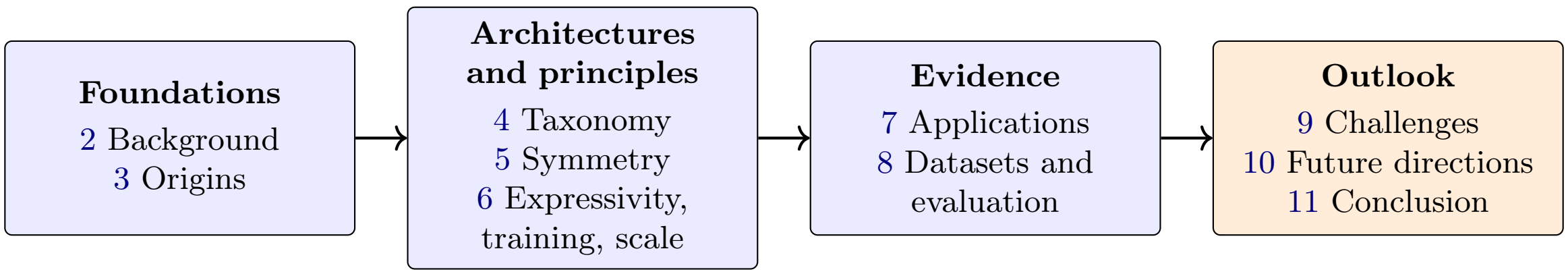

\Cref{fig:roadmap} shows the structure of the paper.
\Cref{sec:background} provides the background: classical GNNs (\cref{sec:classical-gnn}), quantum computing essentials with a graph focus (\cref{sec:qc-essentials}), and the notation and unified formalism used throughout (\cref{sec:notation}).
\Cref{sec:birth} presents the origins of QGNNs, the Hamiltonian-based model and early graph-structured quantum learning.
\Cref{sec:taxonomy} gives the taxonomy of architectures and their comparison, and \cref{sec:symmetry} treats symmetry and geometric inductive biases.
\Cref{sec:expr-train-scale} discusses expressivity, trainability, noise, and scalability.
\Cref{sec:applications} reviews applications, and \cref{sec:datasets} the datasets, evaluation protocols, and reproducibility.
\Cref{sec:challenges} collects the open challenges, \cref{sec:future} the future directions, and \cref{sec:conclusion} concludes.

\paragraph{Reading guide.}
Readers interested in using QGNNs can go to the comparison of architectures (\cref{sec:tax-compare}), the applications (\cref{sec:applications}), and the evaluation protocol (\cref{sec:eval}).
Readers interested in theory can go to symmetry (\cref{sec:symmetry}), the expressivity and trainability results (\cref{sec:universal,sec:bp}), and the theoretical gaps (\cref{sec:ch-theory}).
Readers interested in hardware can go to noise and scalability (\cref{sec:noise,sec:scalability}), the hardware challenges (\cref{sec:ch-hardware}), and the fault-tolerant outlook (\cref{sec:fut-ft}).

% ===========================================================================
\section{Background}\label{sec:background}
% ===========================================================================
\subsection{Classical Graph Neural Networks}\label{sec:classical-gnn}
% ===========================================================================
% Section 2.1  Classical Graph Neural Networks
% (included from main.tex via \input)
% ===========================================================================

Graph neural networks (GNNs) are parametric models that map graph-structured data to vector representations while respecting the symmetries of the underlying relational structure.
Their conceptual roots lie in recursive neural networks for structured data~\citep{sperduti1997supervised,micheli2009neural} and in the first models that iterated a neural transition function over a graph until convergence~\citep{gori2005new,scarselli2009gnn}.
The modern formulation emerged from two converging lines of work: spectral graph convolutions~\citep{bruna2014spectral,henaff2015deep,defferrard2016convolutional,kipf2017semi}, rooted in graph signal processing~\citep{hammond2011wavelets,shuman2013emerging}, and spatial neural message passing~\citep{duvenaud2015convolutional,li2016gated,hamilton2017inductive,gilmer2017neural}, later unified under the umbrella of relational inductive biases~\citep{battaglia2018relational} and geometric deep learning~\citep{bronstein2017geometric,bronstein2021geometric}.
Comprehensive overviews can be found in~\citep{wu2021comprehensive,zhou2020graph,zhang2022deep,hamilton2020graph}, and more recent surveys cover deep graph representation learning~\citep{ju2024comprehensive,khoshraftar2024survey}, graph transformers~\citep{min2022transformer,shehzad2024graph}, expressive power~\citep{sato2020survey,morris2023weisfeiler,zhang2023expressive,morris2024future}, over-smoothing~\citep{rusch2023survey}, over-squashing~\citep{akansha2023over,shi2023exposition,attali2026graph}, and distributed training~\citep{shao2024distributed}.

This subsection provides the classical background on which the quantum models of the later sections build.
We first fix the graph objects and matrix representations (\cref{sec:graph-reps}), then present the main architectural families (\cref{sec:mp-gcn}), formalize the symmetry principle and the resulting expressivity limits (\cref{sec:equivariance}), and finally discuss the three failure modes---over-smoothing, over-squashing, and limited scalability---that motivate much of the interest in alternative computational paradigms (\cref{sec:failure-modes}).

% ---------------------------------------------------------------------------
\subsubsection{Graphs and Representations}\label{sec:graph-reps}
% ---------------------------------------------------------------------------

\paragraph{Graphs and attributes.}
A (finite, simple) graph is a pair $\G=(\V,\E)$ with a set of nodes $\V$, $|\V|=n$, and a set of edges $\E\subseteq\V\times\V$, $|\E|=m$.
Unless stated otherwise, we consider undirected graphs without self-loops, so that $(u,v)\in\E\Leftrightarrow(v,u)\in\E$.
The \emph{neighbourhood} of a node $v$ is $\Nb(v)=\{u\in\V : (u,v)\in\E\}$ and its degree is $d_v=|\Nb(v)|$.
An \emph{attributed} graph additionally carries node features $\xv_v\in\R^{F}$ collected in the matrix $\X\in\R^{n\times F}$, and optionally edge features $\mathbf{e}_{uv}\in\R^{F_e}$.
Weighted, directed, heterogeneous, and temporal graphs, as well as hypergraphs and simplicial complexes, generalize this basic setting; they are treated in the surveys cited above and are mentioned here only when needed.
Learning tasks are classified by their granularity: \emph{node-level} (e.g., classifying papers in a citation network), \emph{edge-level} (link prediction), and \emph{graph-level} (e.g., predicting a molecular property).
Orthogonally, one distinguishes \emph{transductive} settings, in which a single (possibly very large) graph is observed during training and testing, from \emph{inductive} settings, in which the model must generalize to unseen graphs or nodes~\citep{hamilton2017inductive}.

\paragraph{Matrix representations.}
Fixing an arbitrary ordering of the nodes, the structure of $\G$ is encoded by the \emph{adjacency matrix} $\A\in\{0,1\}^{n\times n}$ with $A_{uv}=1$ iff $(u,v)\in\E$, and by the diagonal \emph{degree matrix} $\Dg=\diag(d_1,\dots,d_n)$.
Three Laplacian operators are in common use~\citep{chung1997spectral}:
\begin{equation}\label{eq:laplacians}
  \Lap=\Dg-\A,
  \qquad
  \Lap_{\mathrm{sym}}=\Id-\Dg^{-1/2}\A\,\Dg^{-1/2},
  \qquad
  \Lap_{\mathrm{rw}}=\Id-\Dg^{-1}\A,
\end{equation}
called the combinatorial, symmetric normalized, and random-walk Laplacian, respectively.
All three are positive semi-definite in the appropriate inner product; $\Lap$ and $\Lap_{\mathrm{sym}}$ are real symmetric (hence, trivially, Hermitian) matrices.
The quadratic form of $\Lap$ measures the \emph{smoothness} of a signal $\mathbf{x}\in\R^n$ on the graph,
\begin{equation}\label{eq:dirichlet}
  \mathbf{x}^{\top}\Lap\,\mathbf{x}=\tfrac12\sum_{u,v}A_{uv}\,(x_u-x_v)^2 ,
\end{equation}
and for a feature matrix $\X$ we write $\mathcal{E}(\X)=\tr(\X^{\top}\Lap\,\X)=\tfrac12\sum_{u,v}A_{uv}\|\xv_u-\xv_v\|_2^2$ for the \emph{Dirichlet energy}, which will serve as a diagnostic quantity in \cref{sec:failure-modes}.
The spectrum of $\Lap_{\mathrm{sym}}$ lies in $[0,2]$; the multiplicity of the eigenvalue $0$ equals the number of connected components, and the eigenvalue $2$ occurs iff some component is bipartite~\citep{chung1997spectral}.
With the eigendecomposition $\Lap_{\mathrm{sym}}=\mathbf{U}\boldsymbol{\Lambda}\mathbf{U}^{\top}$, the \emph{graph Fourier transform} of a signal is $\hat{\mathbf{x}}=\mathbf{U}^{\top}\mathbf{x}$, and a spectral filter acts as $g(\Lap_{\mathrm{sym}})\mathbf{x}=\mathbf{U}\,g(\boldsymbol{\Lambda})\,\mathbf{U}^{\top}\mathbf{x}$~\citep{hammond2011wavelets,shuman2013emerging}.
Small eigenvalues correspond to smooth (low-frequency) eigenvectors and large eigenvalues to oscillating (high-frequency) ones.
Throughout, we also use the self-loop-augmented quantities
\begin{equation}\label{eq:renorm}
  \tilde{\A}=\A+\Id,\qquad
  \tilde{\Dg}=\Dg+\Id,\qquad
  \hat{\A}=\tilde{\Dg}^{-1/2}\,\tilde{\A}\,\tilde{\Dg}^{-1/2},
\end{equation}
where $\hat{\A}$ is the renormalized propagation operator of~\citet{kipf2017semi}.

\paragraph{Other data structures.}
In software, sparse graphs are stored as edge lists (COO) or compressed sparse row/column formats, which require $\mathcal{O}(m)$ rather than $\mathcal{O}(n^2)$ memory, a distinction that becomes essential for scalability (\cref{sec:scalability-classical}).
Message-passing implementations such as PyTorch Geometric~\citep{fey2019fast} and the Deep Graph Library~\citep{wang2019deep} operate directly on these sparse formats.
\Cref{tab:graph-reps} summarizes the representations used in the remainder of the survey.

\begin{table}[t]
\centering
\caption{Common representations of a graph $\G$ with $n$ nodes and $m$ edges, and their typical role in GNNs.}
\label{tab:graph-reps}
\renewcommand{\arraystretch}{1.25}
\small
\begin{tabular}{@{}p{3.3cm}p{2.6cm}p{8.4cm}@{}}
\toprule
\textbf{Representation} & \textbf{Memory} & \textbf{Role in (quantum) graph learning} \\
\midrule
Adjacency matrix $\A$ & $\mathcal{O}(n^2)$ dense; $\mathcal{O}(m)$ sparse & Defines neighbourhoods and message routing; basis of spectral methods and of Hamiltonian-type encodings of the graph. \\
Laplacians $\Lap,\Lap_{\mathrm{sym}},\Lap_{\mathrm{rw}}$ & same & Smoothness (Dirichlet energy), spectral filters, diffusion, positional encodings. \\
Edge list / CSR & $\mathcal{O}(m)$ & Efficient sparse message passing and mini-batch sampling. \\
Node/edge feature matrices $\X$, $\mathbf{E}$ & $\mathcal{O}(nF)$, $\mathcal{O}(mF_e)$ & Inputs to message functions; must be embedded (``encoded'') in qubits for quantum models. \\
Positional/structural encodings & $\mathcal{O}(nk)$ & Eigenvectors of $\Lap_{\mathrm{sym}}$, random-walk statistics, shortest-path distances; break symmetries beyond 1-WL (\cref{sec:equivariance}). \\
\bottomrule
\end{tabular}
\end{table}

\paragraph{Positional and structural encodings.}
Because plain message passing is blind to the absolute position of a node in the graph, several works augment the input features with \emph{positional} encodings (PEs) or \emph{structural} encodings (SEs).
Typical choices are the first $k$ non-trivial eigenvectors of $\Lap_{\mathrm{sym}}$~\citep{dwivedi2023benchmarking,dwivedi2021generalization,kreuzer2021rethinking}, random-walk return probabilities and learnable variants~\citep{dwivedi2022graph}, anchor-based distances~\citep{you2019position}, and distance encodings~\citep{li2020distance}.
Laplacian eigenvectors are only defined up to sign (and, for repeated eigenvalues, up to an orthogonal change of basis), which motivated sign- and basis-invariant architectures~\citep{lim2023sign}.

% ---------------------------------------------------------------------------
\subsubsection{Message Passing, Graph Convolution, and Graph Transformers}\label{sec:mp-gcn}
% ---------------------------------------------------------------------------

\paragraph{Early recurrent models.}
The original GNN of~\citet{gori2005new} and~\citet{scarselli2009gnn} associates a state $\hv_v$ with every node and iterates a contractive local transition function to a fixed point, $\hv_v^{(t+1)}=f_{\mathbf{w}}\!\big(\xv_v,\{\mathbf{e}_{uv},\xv_u,\hv_u^{(t)}\}_{u\in\Nb(v)}\big)$, training through the fixed point.
Gated graph sequence networks~\citep{li2016gated} replaced the fixed-point constraint by a fixed number of gated recurrent (GRU) updates, and are the direct ancestors of the recurrent QGNN designs of \cref{sec:tax-mp}.

\paragraph{Spectral graph convolutions.}
Motivated by the convolution theorem, \citet{bruna2014spectral} defined a convolution as multiplication in the graph Fourier domain, $g_{\boldsymbol\theta}\star\mathbf{x}=\mathbf{U}\,g_{\boldsymbol\theta}(\boldsymbol\Lambda)\,\mathbf{U}^{\top}\mathbf{x}$.
This requires an $\mathcal{O}(n^3)$ eigendecomposition, produces non-localized filters, and does not transfer across graphs.
ChebNet~\citep{defferrard2016convolutional} approximates the filter by a truncated Chebyshev expansion,
\begin{equation}\label{eq:cheb}
  g_{\boldsymbol\theta}(\Lap_{\mathrm{sym}})\,\mathbf{x}\;\approx\;\sum_{k=0}^{K}\theta_k\,T_k(\tilde{\Lap})\,\mathbf{x},
  \qquad
  \tilde{\Lap}=\tfrac{2}{\lambda_{\max}}\Lap_{\mathrm{sym}}-\Id,
\end{equation}
with $T_0=1$, $T_1(x)=x$, and $T_k(x)=2xT_{k-1}(x)-T_{k-2}(x)$.
Because $T_k(\tilde\Lap)$ is a degree-$k$ polynomial of the Laplacian, the filter is exactly $K$-hop localized and costs $\mathcal{O}(Km)$ per feature channel, without any eigendecomposition.
The graph convolutional network (GCN)~\citep{kipf2017semi} further sets $K=1$, $\lambda_{\max}\approx 2$, and ties $\theta_0=-\theta_1=\theta$, which yields $\theta(\Id+\Dg^{-1/2}\A\Dg^{-1/2})\mathbf{x}$.
To avoid numerical instability, the \emph{renormalization trick} replaces this operator by $\hat{\A}$ from \cref{eq:renorm}, giving the layer
\begin{equation}\label{eq:gcn}
  \Hm^{(\ell+1)}=\sigma\!\big(\hat{\A}\,\Hm^{(\ell)}\,\W^{(\ell)}\big),\qquad \Hm^{(0)}=\X,
\end{equation}
where $\W^{(\ell)}\in\R^{F_\ell\times F_{\ell+1}}$ is a learnable matrix and $\sigma$ a pointwise nonlinearity (typically ReLU).
Spectrally, \cref{eq:gcn} is a fixed low-pass filter followed by a feature transformation~\citep{nt2019revisiting,wu2019simplifying}.
A line of work has therefore studied \emph{learnable} spectral filters---ARMA filters~\citep{bianchi2022graph}, Bernstein-polynomial filters~\citep{he2021bernnet}, revisited Chebyshev approximations~\citep{he2022convolutional}, and band-pass/high-frequency information~\citep{bo2021beyond}---as well as their expressive power~\citep{balcilar2021analyzing,wang2022powerful}.

\begin{figure}[t]
\centering
\begin{tikzpicture}[
  nd/.style={circle,draw,minimum size=7mm,inner sep=0pt,font=\footnotesize},
  msg/.style={-{Stealth[length=2.2mm]},thick,blue!60!black},
  lab/.style={font=\scriptsize,blue!60!black}]
  \node[nd,fill=blue!15] (v) at (0,0) {$v$};
  \node[nd] (u1) at (-2.4,1.5) {$u_1$};
  \node[nd] (u2) at (-2.4,-1.5) {$u_2$};
  \node[nd] (u3) at (2.4,1.5) {$u_3$};
  \node[nd] (w)  at (-4.6,2.2) {$w$};
  \draw[msg] (u1) -- node[lab,above right,pos=0.45] {$\MSG$} (v);
  \draw[msg] (u2) -- node[lab,below right,pos=0.45] {$\MSG$} (v);
  \draw[msg] (u3) -- node[lab,above left,pos=0.45] {$\MSG$} (v);
  \draw[msg,gray] (w) -- (u1);
  \node[draw,rounded corners,fill=orange!15,font=\footnotesize,inner sep=3pt] (agg) at (3.4,-0.9) {$\hv_v^{(\ell+1)}=\UPD\big(\hv_v^{(\ell)},\AGG(\cdot)\big)$};
  \draw[-{Stealth[length=2mm]},orange!70!black] (v) -- (agg);
  \node[font=\scriptsize,gray,align=center] at (-4.6,0.9) {reaches $v$ only\\after 2 layers};
\end{tikzpicture}
\caption{One round of message passing~\cref{eq:mpnn}. Node $v$ aggregates messages from its neighbours $u_1,u_2,u_3$ and updates its state. Information from the two-hop node $w$ reaches $v$ only after a second layer, which links the number of layers to the receptive field.}
\label{fig:message-passing}
\end{figure}
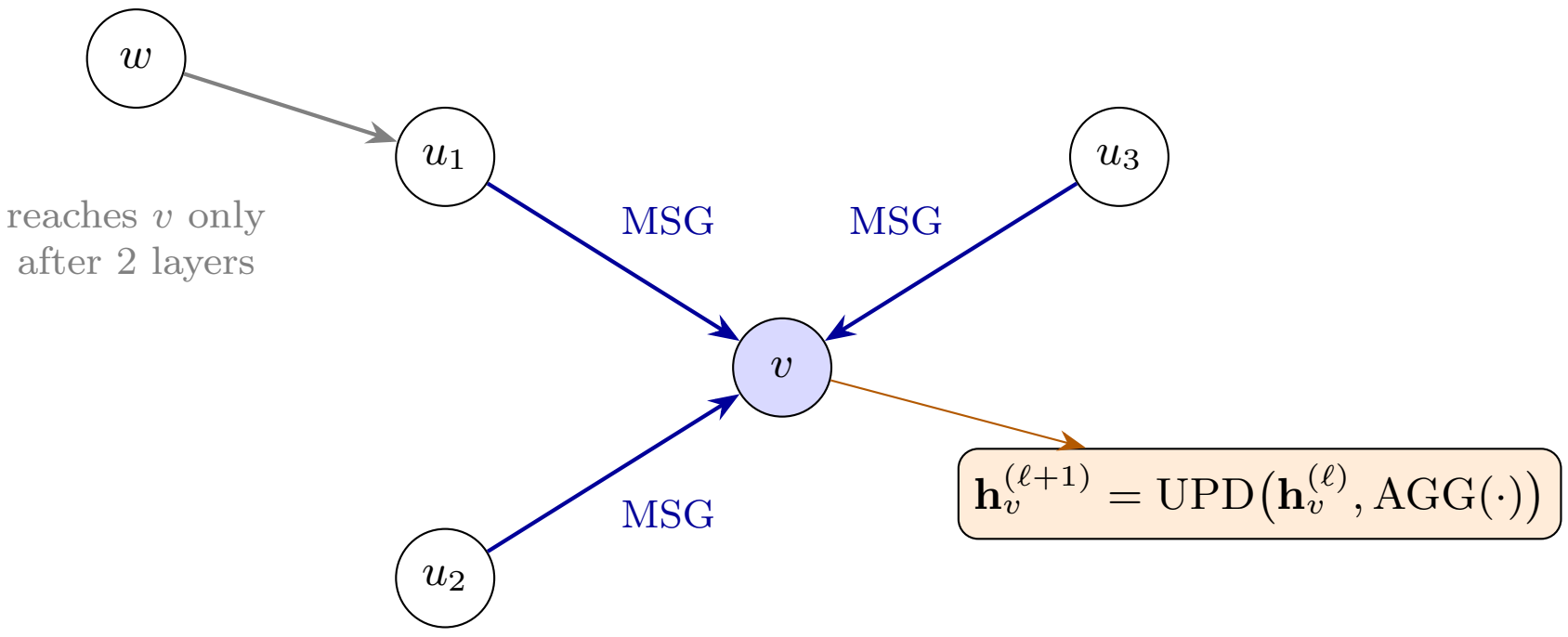

\paragraph{The message-passing framework.}
Most spatial architectures fit the \emph{message-passing neural network} (MPNN) template of~\citet{gilmer2017neural} (see also~\citep{battaglia2018relational}).
Each layer computes, for every node $v$,
\begin{equation}\label{eq:mpnn}
  \mathbf{m}_v^{(\ell+1)}=\AGG^{(\ell)}\!\Big(\big\{\!\!\big\{\MSG^{(\ell)}\!\big(\hv_v^{(\ell)},\hv_u^{(\ell)},\mathbf{e}_{uv}\big):u\in\Nb(v)\big\}\!\!\big\}\Big),
  \qquad
  \hv_v^{(\ell+1)}=\UPD^{(\ell)}\!\big(\hv_v^{(\ell)},\mathbf{m}_v^{(\ell+1)}\big),
\end{equation}
where $\{\!\!\{\cdot\}\!\!\}$ denotes a multiset, $\MSG$ and $\UPD$ are learnable functions (typically MLPs or recurrent cells), and $\AGG$ is a permutation-invariant aggregator such as sum, mean, or max.
After $L$ layers, $\hv_v^{(L)}$ depends on the $L$-hop neighbourhood of $v$ (its \emph{receptive field}).
For graph-level tasks, node embeddings are pooled by a permutation-invariant readout,
\begin{equation}\label{eq:readout}
  \hv_{\G}=\READOUT\big(\{\!\!\{\hv_v^{(L)}:v\in\V\}\!\!\}\big),
\end{equation}
e.g., sum/mean pooling or learned hierarchical pooling~\citep{ying2018hierarchical}.
\Cref{fig:message-passing} illustrates one round of message passing.

\paragraph{Aggregation schemes and notable instances.}
\Cref{eq:gcn} corresponds to a degree-weighted mean over $\Nb(v)\cup\{v\}$.
GraphSAGE~\citep{hamilton2017inductive} introduced inductive aggregators (mean, LSTM, pooling) and fixed-size neighbour sampling, $\hv_v'=\sigma\big(\W\,[\hv_v\,\Vert\,\AGG(\{\hv_u\}_{u\in\Nb(v)})]\big)$.
The \emph{graph isomorphism network} (GIN)~\citep{xu2019powerful} uses sum aggregation and an MLP,
\begin{equation}\label{eq:gin}
  \hv_v^{(\ell+1)}=\mathrm{MLP}^{(\ell)}\!\Big((1+\epsilon^{(\ell)})\,\hv_v^{(\ell)}+\sum_{u\in\Nb(v)}\hv_u^{(\ell)}\Big),
\end{equation}
and is the maximally expressive MPNN in the sense made precise in \cref{sec:equivariance}.
Principal neighbourhood aggregation (PNA)~\citep{corso2020principal} combines several aggregators (mean, max, min, standard deviation) with degree-based scalers to overcome the limits of single aggregators on continuous features.
Earlier convolutional constructions on graphs and manifolds include PATCHY-SAN~\citep{niepert2016learning} and MoNet~\citep{monti2017geometric}.
Jumping-knowledge networks~\citep{xu2018representation} combine representations from all layers, and propagation-based models such as APPNP~\citep{klicpera2019predict} replace repeated transformation by personalized PageRank diffusion, $\mathbf{Z}^{(k+1)}=(1-\alpha)\hat{\A}\mathbf{Z}^{(k)}+\alpha\,\Hm$ with $\Hm=\mathrm{MLP}(\X)$.

\paragraph{Graph attention.}
Graph attention networks (GAT)~\citep{velickovic2018graph} make the aggregation weights data-dependent,
\begin{equation}\label{eq:gat}
  \hv_v'=\sigma\Big(\sum_{u\in\Nb(v)\cup\{v\}}\alpha_{vu}\,\W\hv_u\Big),
  \qquad
  \alpha_{vu}=\frac{\exp\!\big(\mathrm{LeakyReLU}(\mathbf{a}^{\top}[\W\hv_v\,\Vert\,\W\hv_u])\big)}{\sum_{k\in\Nb(v)\cup\{v\}}\exp\!\big(\mathrm{LeakyReLU}(\mathbf{a}^{\top}[\W\hv_v\,\Vert\,\W\hv_k])\big)},
\end{equation}
with multi-head concatenation or averaging.
\citet{brody2022attentive} showed that the attention of \cref{eq:gat} is \emph{static}---the ranking of neighbours is the same for every query node---and proposed GATv2, which applies the nonlinearity after the linear map on the concatenation and thereby realizes dynamic attention.
Attention-based aggregation is a prototype for the quantum attention mechanisms discussed later in the survey.

\paragraph{Graph transformers.}
Transformers~\citep{vaswani2017attention} compute $\hv_v'=\sum_{u}\softmax_u\big(\mathbf{q}_v^{\top}\mathbf{k}_u/\sqrt{d}\big)\,\mathbf{v}_u$ over a set of tokens, which on a graph amounts to message passing on the complete graph with learned weights.
Since self-attention is permutation-equivariant but ignores topology, graph transformers inject structure through positional encodings or attention biases.
Early designs restricted attention to neighbours and added Laplacian PEs~\citep{dwivedi2021generalization}; SAN~\citep{kreuzer2021rethinking} attends over the full graph while using the Laplacian spectrum in a learned way; Graphormer~\citep{ying2021transformers} adds centrality (degree) embeddings, a shortest-path-distance bias on the attention logits, and edge encodings,
\begin{equation}\label{eq:graphormer}
  \mathrm{Attn}_{vu}=\softmax_u\!\Big(\tfrac{\mathbf{q}_v^{\top}\mathbf{k}_u}{\sqrt d}+b_{\phi(v,u)}+c_{vu}\Big),
\end{equation}
where $\phi(v,u)$ is the shortest-path distance and $c_{vu}$ an edge-feature term; and GPS~\citep{rampasek2022recipe} combines a local MPNN branch, a global attention branch, and positional/structural encodings in each layer.
Because dense attention costs $\mathcal{O}(n^2)$, scalable variants rely on kernelized or sparse attention: NodeFormer~\citep{wu2022nodeformer} achieves linear complexity via kernelized Gumbel-softmax message passing, SGFormer~\citep{wu2023sgformer} shows that a single global attention layer can suffice on large graphs, and Exphormer~\citep{shirzad2023exphormer} sparsifies attention using expander graphs and virtual nodes.
The relationship between MPNNs with virtual nodes and transformers is analyzed by~\citet{cai2023connection}, and \citet{muller2024attending} provide a systematic assessment of when and why graph transformers help.
\Cref{tab:gnn-layers} contrasts the architectures discussed in this subsection.

\begin{table}[t]
\centering
\caption{Representative classical GNN layers. $\mathcal{N}^{+}(v)=\Nb(v)\cup\{v\}$; $n$ nodes, $m$ edges, feature width $F$. Complexities are per layer, per forward pass of a full graph.}
\label{tab:gnn-layers}
\renewcommand{\arraystretch}{1.35}
\footnotesize
\begin{tabular}{@{}p{2.7cm}p{6.2cm}p{2.3cm}p{3.4cm}@{}}
\toprule
\textbf{Model} & \textbf{Node update} & \textbf{Time} & \textbf{Key property} \\
\midrule
ChebNet~\citep{defferrard2016convolutional} & $\sum_{k=0}^{K}\theta_kT_k(\tilde\Lap)\Hm$ & $\mathcal{O}(KmF)$ & $K$-hop localized spectral filter \\
GCN~\citep{kipf2017semi} & $\sigma(\hat\A\Hm\W)$ & $\mathcal{O}(mF+nF^2)$ & Fixed low-pass filter \\
GraphSAGE~\citep{hamilton2017inductive} & $\sigma(\W[\hv_v\Vert\AGG_{u\in\Nb(v)}\hv_u])$ & $\mathcal{O}(mF+nF^2)$ & Inductive; sampling \\
GAT~\citep{velickovic2018graph} & $\sigma(\sum_{u\in\mathcal{N}^{+}(v)}\alpha_{vu}\W\hv_u)$ & $\mathcal{O}(mF+nF^2)$ & Learned anisotropic weights \\
GIN~\citep{xu2019powerful} & $\mathrm{MLP}((1+\epsilon)\hv_v+\sum_{u}\hv_u)$ & $\mathcal{O}(mF+nF^2)$ & As expressive as 1-WL \\
PNA~\citep{corso2020principal} & multiple aggregators $\times$ degree scalers & $\mathcal{O}(mF+nF^2)$ & Richer aggregation \\
Graphormer~\citep{ying2021transformers}, GPS~\citep{rampasek2022recipe} & global attention with structural biases (\cref{eq:graphormer}) & $\mathcal{O}(n^2F)$ & Global receptive field \\
\bottomrule
\end{tabular}
\end{table}

% ---------------------------------------------------------------------------
\subsubsection{Permutation Equivariance, Invariance, and Expressivity Limits}\label{sec:equivariance}
% ---------------------------------------------------------------------------

\paragraph{Symmetry.}
A graph has no canonical node ordering: relabeling the nodes by a permutation $\pi\in\Sym$, represented by a permutation matrix $\Perm\in\{0,1\}^{n\times n}$, changes $(\A,\X)$ into $(\Perm\A\Perm^{\top},\Perm\X)$ but not the graph.
A function $f$ acting on graphs must therefore respect this relabeling~\citep{zaheer2017deep,maron2019invariant,bronstein2021geometric}.

\begin{definition}[Permutation invariance and equivariance]\label{def:equiv}
A function $f:\R^{n\times n}\times\R^{n\times F}\to\R^{d}$ is \emph{permutation-invariant} if $f(\Perm\A\Perm^{\top},\Perm\X)=f(\A,\X)$ for all $\Perm$.
A function $F:\R^{n\times n}\times\R^{n\times F}\to\R^{n\times d}$ is \emph{permutation-equivariant} if $F(\Perm\A\Perm^{\top},\Perm\X)=\Perm\,F(\A,\X)$ for all $\Perm$.
\end{definition}

Node-level tasks call for equivariant maps and graph-level tasks for invariant ones; the standard blueprint is a stack of equivariant layers followed by an invariant readout~\cref{eq:readout}~\citep{bronstein2021geometric}.

\begin{proposition}[Message passing is equivariant]\label{prop:mp-equiv}
Every MPNN layer~\cref{eq:mpnn} with a permutation-invariant aggregator is permutation-equivariant, and any composition of such layers followed by the readout~\cref{eq:readout} is permutation-invariant.
\end{proposition}
\begin{proof}[Proof sketch]
Relabeling nodes by $\pi$ maps the multiset $\{\!\!\{\MSG(\hv_v,\hv_u,\mathbf{e}_{uv}):u\in\Nb(v)\}\!\!\}$ to the multiset for node $\pi(v)$ in the relabeled graph, which contains the same elements; because $\AGG$ depends only on the multiset, $\mathbf{m}_{\pi(v)}'=\mathbf{m}_v$ and hence $\hv_{\pi(v)}'=\hv_v$. Induction on $\ell$ extends this to $L$ layers, and the readout is invariant by assumption.
\end{proof}

At the other extreme, one may parametrize \emph{all} linear equivariant maps on tensors of order $k$. \citet{maron2019invariant} showed that the space of linear $\Sym$-equivariant layers from $\R^{n^{k}}$ to $\R^{n^{l}}$ has a basis indexed by set partitions of $\{1,\dots,k+l\}$, so its dimension is the Bell number $\mathrm{Bell}(k+l)$ and is independent of $n$ (for example, $15$ for maps from adjacency-like matrices to matrices, $k=l=2$).
The resulting \emph{invariant and equivariant graph networks} (IGNs) have cost $\mathcal{O}(n^{k})$ in memory, which is prohibitive for large $k$ and anticipates the equivariance--scalability trade-off of quantum circuits (\cref{sec:symmetry}).

\paragraph{The 1-WL test and the power of MPNNs.}
The expressive power of GNNs is best understood through the Weisfeiler--Leman (WL) hierarchy~\citep{weisfeiler1968reduction,morris2023weisfeiler,morris2021power}.
The 1-dimensional WL (1-WL), or colour refinement, assigns initial colours $c^{(0)}(v)$ from the features and iterates
\begin{equation}\label{eq:wl}
  c^{(t+1)}(v)=\mathrm{HASH}\Big(c^{(t)}(v),\{\!\!\{c^{(t)}(u):u\in\Nb(v)\}\!\!\}\Big),
\end{equation}
declaring two graphs non-isomorphic if their colour histograms differ at some iteration~\citep{shervashidze2011weisfeiler}.
Comparing \cref{eq:wl} with \cref{eq:mpnn} reveals that an MPNN is a \emph{continuous relaxation} of 1-WL, and the following equivalence was established independently by~\citet{xu2019powerful} and~\citet{morris2019weisfeiler}.

\begin{theorem}[MPNNs and 1-WL~\citep{xu2019powerful,morris2019weisfeiler}]\label{thm:1wl}
(i) If 1-WL assigns the same colour to nodes $u,v$ after $t$ iterations, then every MPNN with $t$ layers produces the same embedding for $u$ and $v$; consequently, graphs not separated by 1-WL receive identical graph-level embeddings from any MPNN.
(ii) Conversely, if $\AGG$ and $\UPD$ are injective on multisets (as for GIN~\cref{eq:gin} with sufficiently wide MLPs and countable features), the MPNN separates exactly the graphs separated by 1-WL.
\end{theorem}

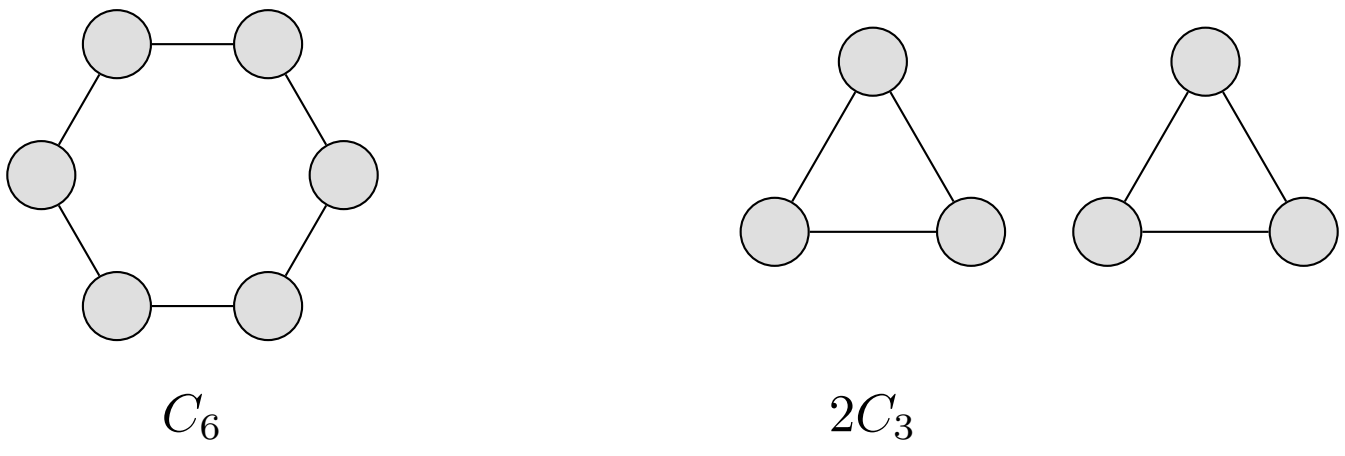
\begin{figure}[t]
\centering
\begin{tikzpicture}[v/.style={circle,draw,fill=gray!25,minimum size=4.5mm,inner sep=0pt}]
  % hexagon
  \begin{scope}[xshift=0cm]
    \foreach \i in {0,...,5}{\node[v] (a\i) at ({60*\i}:1) {};}
    \foreach \i/\j in {0/1,1/2,2/3,3/4,4/5,5/0}{\draw (a\i)--(a\j);}
    \node at (0,-1.6) {\small $C_6$};
  \end{scope}
  % two triangles
  \begin{scope}[xshift=4.5cm]
    \foreach \i in {0,1,2}{\node[v] (b\i) at ({90+120*\i}:0.75) {};}
    \foreach \i/\j in {0/1,1/2,2/0}{\draw (b\i)--(b\j);}
    \node at (0,-1.6) {\small $2C_3$};
  \end{scope}
  \begin{scope}[xshift=6.7cm]
    \foreach \i in {0,1,2}{\node[v] (c\i) at ({90+120*\i}:0.75) {};}
    \foreach \i/\j in {0/1,1/2,2/0}{\draw (c\i)--(c\j);}
  \end{scope}
\end{tikzpicture}
\caption{A failure case of 1-WL and hence of every MPNN (\cref{thm:1wl}): all nodes in both graphs have degree two and identical initial features, so colour refinement~\cref{eq:wl} never splits the colour class, although the graphs are not isomorphic (one is connected, the other is not; they have different numbers of triangles).}
\label{fig:wl-fail}
\end{figure}

Thus, 1-WL is a tight upper bound for standard message passing~\citep{xu2019powerful,morris2019weisfeiler}, and the choice of aggregator matters: sum is injective on multisets of countable features, whereas mean and max lose multiplicity information~\citep{xu2019powerful,corso2020principal}.
The bound has concrete consequences.
MPNNs cannot distinguish regular graphs of the same degree and size, such as the hexagon $C_6$ and two disjoint triangles $2C_3$ (\cref{fig:wl-fail}); they cannot count triangles or other cycles~\citep{chen2020can}, determine whether a graph is bi-connected~\citep{zhang2023rethinking}, or detect many distance-based properties without additional encodings~\citep{li2020distance}.
\citet{loukas2020what} furthermore shows that, under suitable assumptions, MPNNs need depth and width growing polynomially with $n$ for several classical graph problems, and \citet{garg2020generalization} derive complementary limitations together with generalization bounds.
From a logical perspective, MPNNs correspond to fragments of graded modal logic and, with a global readout, to the two-variable fragment of first-order logic with counting~\citep{barcelo2020logical,grohe2021logic}.

\paragraph{Climbing the hierarchy.}
The $k$-WL hierarchy colours $k$-tuples of nodes and is strictly increasing in power~\citep{cai1992optimal}. Higher-order GNNs inherit this hierarchy: $k$-GNNs~\citep{morris2019weisfeiler}, $k$-IGNs~\citep{maron2019invariant}, and provably powerful graph networks (PPGN), which match the 3-WL (2-folklore-WL) test using matrix multiplication~\citep{maron2019provably}.
Their memory cost $\mathcal{O}(n^{k})$ limits applicability to small graphs.
Equivalences between expressive power, graph isomorphism testing, and function approximation were proved by~\citet{chen2019equivalence} and~\citet{azizian2021expressive}; universality of invariant/equivariant networks was studied in~\citep{maron2019universality,keriven2019universal}, and a unified treatment of expressiveness via tensor languages is given by~\citet{geerts2022expressiveness}.
Because full higher-order architectures are expensive, a rich family of cheaper extensions of 1-WL has been developed:
\begin{itemize}[leftmargin=1.5em,itemsep=2pt,topsep=2pt]
  \item \emph{Structural and positional features}: substructure (e.g., cycle or clique) counts~\citep{bouritsas2023improving}, distance encodings~\citep{li2020distance}, anchor-based positions~\citep{you2019position}, and spectral encodings made invariant to sign and basis~\citep{dwivedi2022graph,lim2023sign}; the expressive power of spectral-invariant architectures is characterized by~\citet{zhang2024expressive}.
  \item \emph{Subgraph GNNs}: the graph is processed as a bag of (node-marked or node-deleted) subgraphs~\citep{zhang2021nested,you2021identity,bevilacqua2022equivariant,frasca2022understanding}; closely related is the labeling trick for multi-node tasks~\citep{zhang2021labeling}.
  \item \emph{Randomization}: random node features or colourings~\citep{sato2021random,abboud2021surprising,dasoulas2020coloring}, random dropouts~\citep{papp2021dropgnn}, and relational pooling over permutations~\citep{murphy2019relational} trade exact invariance for expressivity (in expectation or probability).
  \item \emph{Global operations}: virtual nodes, global attention, and graph transformers with structural biases~\citep{puny2020global,ying2021transformers,rampasek2022recipe}.
\end{itemize}
Recent work has moved beyond binary ``can/cannot distinguish'' statements. \citet{zhang2024beyond} introduce \emph{homomorphism expressivity}, which characterizes a model by the set of graph patterns whose homomorphism counts it can compute, \citet{wijesinghe2022new} refine the comparison with 1-WL through local structure, and \citet{franks2024weisfeiler} relate WL-based expressivity to generalization via a margin analysis, showing that higher expressivity helps only when it translates into larger margins.
The broader theory is surveyed in~\citep{jegelka2022theory,morris2023weisfeiler,zhang2023expressive,morris2024future}; in particular, the geometric variants for 3D data are treated in~\citep{joshi2023expressive,satorras2021n}, which we revisit in \cref{sec:sym-rot}.

\begin{remark}[Relevance for quantum models]
The symmetry principle of \cref{def:equiv} is architecture-agnostic: it is independent of whether a layer is implemented classically or as a parametrized quantum circuit.
We return to this point in \cref{sec:tax-eqgc,sec:sym-perm}, where permutation equivariance is imposed on quantum circuits, and in \cref{sec:universal}, where expressivity statements are transferred to the quantum setting.
Importantly, the bounds above are statements about \emph{classical message passing}; they should not be read as saying that quantum architectures are necessarily more or less expressive.
\end{remark}

% ---------------------------------------------------------------------------
\subsubsection{Failure Modes: Over-Smoothing, Over-Squashing, and Scalability}\label{sec:failure-modes}
% ---------------------------------------------------------------------------

Despite their success, message-passing GNNs exhibit three well-documented limitations that become acute for deep models, long-range dependencies, or very large graphs.
Other practical difficulties, such as heterophily~\citep{pei2020geom,zhu2020beyond,platonov2023critical} and unreliable evaluation on long-range benchmarks~\citep{dwivedi2022long,tonshoff2023where}, are related but not treated in detail here.

\paragraph{Over-smoothing.}
\emph{Over-smoothing} refers to the phenomenon that node representations become increasingly similar as the number of layers grows, until they are indistinguishable and performance degrades.
It was first noted by~\citet{li2018deeper}, who interpreted the GCN propagation $\hat{\A}\Hm$ as a form of Laplacian smoothing.
The mechanism is easiest to see for the linear propagation $\hat{\A}^{k}\X$.

\begin{proposition}[Smoothing by repeated propagation~\citep{li2018deeper,oono2020graph,nt2019revisiting}]\label{prop:smoothing}
Let $\G$ be connected with renormalized operator $\hat{\A}$ from~\cref{eq:renorm}. Then $\hat{\A}$ is symmetric with eigenvalues $1=\mu_1>\mu_2\ge\dots\ge\mu_n>-1$, and the eigenvector of $\mu_1$ is $\boldsymbol\varphi=\tilde{\Dg}^{1/2}\mathbf{1}/\|\tilde{\Dg}^{1/2}\mathbf{1}\|_2$. Consequently,
\begin{equation}\label{eq:smoothing}
  \big\|\hat{\A}^{k}\X-\boldsymbol\varphi\boldsymbol\varphi^{\top}\X\big\|_F\le\mu_{*}^{k}\,\|\X\|_F,\qquad \mu_{*}=\max(|\mu_2|,|\mu_n|)<1 .
\end{equation}
\end{proposition}
\begin{proof}[Proof sketch]
$\hat{\A}=\Id-\tilde{\Lap}_{\mathrm{sym}}$ for the normalized Laplacian of the self-loop-augmented graph, whose spectrum lies in $[0,2)$ because the self-loops break bipartiteness; connectivity makes the eigenvalue $0$ simple. Expanding $\X$ in the eigenbasis of $\hat{\A}$ and applying $\hat{\A}^{k}$ multiplies the $i$-th component by $\mu_i^{k}$, so all components except the first decay geometrically.
\end{proof}

Hence, all node representations converge to rows proportional to $\sqrt{\tilde d_v}$---they retain only degree information.
For nonlinear GCNs with weights, \citet{oono2020graph} proved that, under conditions on the singular values of the weights relative to the spectral gap of the graph, features converge exponentially to an invariant subspace, and \citet{cai2020note} showed analogous decay for the Dirichlet energy. Writing $s_\ell$ for the largest singular value of $\W^{(\ell)}$, the decay rate is governed by the product of $s_\ell$ and the largest non-unit eigenvalue magnitude of $\hat{\A}$.
Two measurements are used in practice: the Dirichlet energy $\mathcal{E}(\Hm^{(\ell)})$ (\cref{eq:dirichlet}) and the mean average distance (MAD) between node representations~\citep{chen2020measuring}.
The theory has since been refined in several directions.
Non-asymptotic analyses relate the onset of smoothing to graph structure, weight norms, and depth~\citep{wu2023non}; \citet{keriven2022not} show that a moderate amount of smoothing is beneficial and only excessive smoothing is harmful; \citet{wu2023demystifying} prove that attention-based GNNs also oversmooth, i.e., attention does not by itself prevent the problem; \citet{roth2023rank} connect oversmoothing to rank collapse; \citet{epping2024graph} identify regimes in which GNNs in the infinite-width limit do \emph{not} oversmooth; and \citet{zhang2025measuring} question whether the commonly used measures correctly capture the phenomenon.
Mitigation strategies fall into four groups:
(i) \emph{architectural}: residual and initial-residual connections~\citep{chen2020simple,li2019deepgcns,li2020deepergcn}, jumping knowledge~\citep{xu2018representation}, reversible blocks for very deep networks~\citep{li2021training};
(ii) \emph{normalization and regularization}: PairNorm~\citep{zhao2020pairnorm}, differentiable group normalization~\citep{zhou2020towards}, DropEdge~\citep{rong2020dropedge}, Dirichlet-energy constraints~\citep{zhou2021dirichlet};
(iii) \emph{continuous-time/physics-inspired}: graph diffusion and PDE-based models~\citep{chamberlain2021grand,thorpe2022grand,eliasof2021pde}, oscillator and anti-symmetric networks~\citep{rusch2022graph,gravina2023anti}, and sheaf diffusion~\citep{bodnar2022neural}, see also the energy-based analysis of~\citep{digiovanni2023energies};
(iv) \emph{filter design}, which retains high-frequency components~\citep{bo2021beyond}.
Physics-inspired continuous-time models are of special interest for this survey because they connect to Hamiltonian-based QGNNs (\cref{sec:birth-hamiltonian,sec:tax-physics}).

\begin{figure}[t]
\centering
\begin{tikzpicture}[
  s/.style={circle,draw,fill=gray!25,minimum size=3.2mm,inner sep=0pt},
  h/.style={circle,draw,fill=blue!20,minimum size=6mm,inner sep=0pt,font=\scriptsize}]
  % left cluster
  \node[h] (L) at (0,0) {$a$};
  \foreach \i in {0,...,6}{\node[s] (l\i) at ({-1.9+0.0*\i},{-1.8+0.6*\i}) {}; \draw (l\i)--(L);}
  % right cluster
  \node[h] (R) at (3.6,0) {$b$};
  \foreach \i in {0,...,6}{\node[s] (r\i) at ({5.5},{-1.8+0.6*\i}) {}; \draw (r\i)--(R);}
  % bottleneck edge
  \draw[line width=1.4pt,red!70!black] (L)--node[above,font=\scriptsize,red!70!black]{bottleneck}(R);
  \node[font=\scriptsize,align=center] at (-1.9,-2.5) {$\mathcal{O}(d^{\,r})$ nodes within\\$r$ hops of $a$};
  \node[font=\scriptsize,align=center] at (1.8,-1.4) {all messages squeezed into\\one $F$-dimensional vector};
\end{tikzpicture}
\caption{Over-squashing: information from many nodes on one side of a bottleneck must pass through a single edge and be compressed into the fixed-size state of node $a$ before reaching node $b$. The number of nodes within $r$ hops can grow exponentially while the capacity of the channel does not.}
\label{fig:oversquash}
\end{figure}
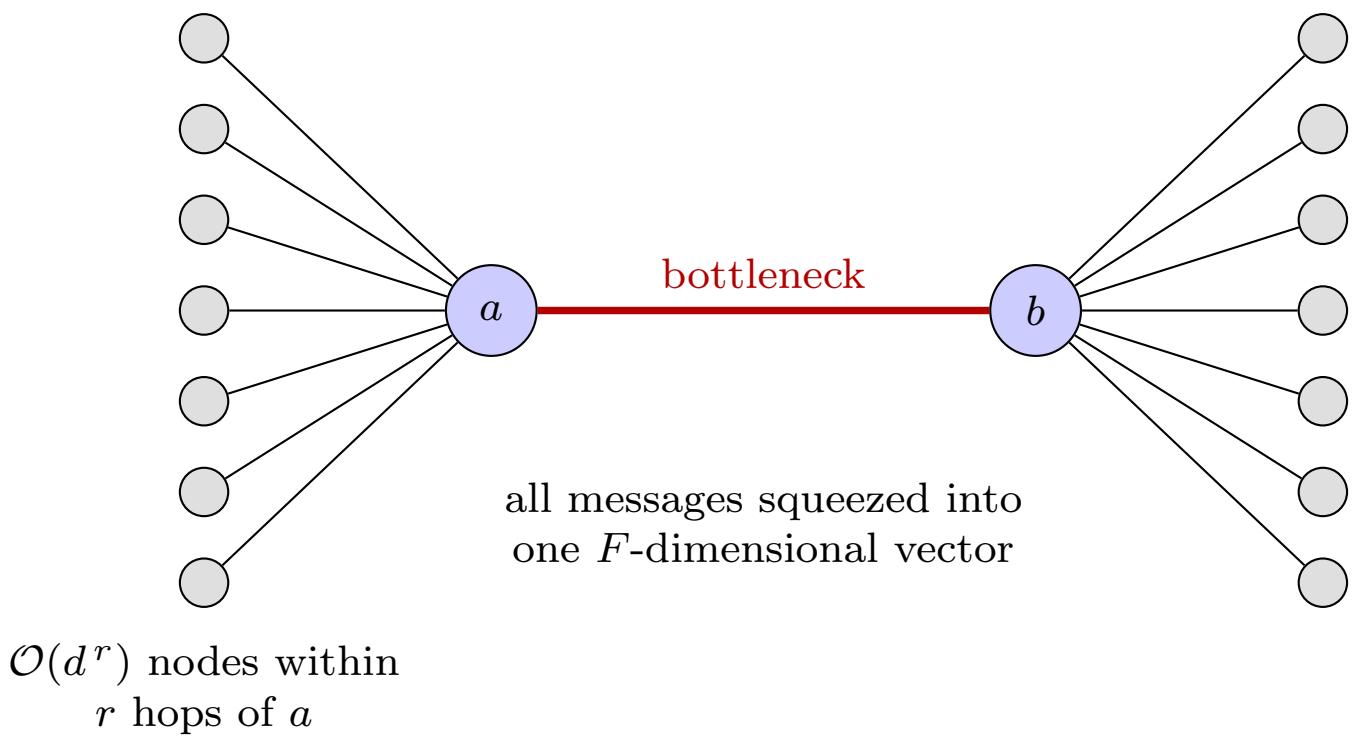

\paragraph{Over-squashing.}
\emph{Over-squashing}~\citep{alon2021bottleneck} is the distortion of information that arises when messages from an exponentially growing receptive field must be compressed into fixed-size node vectors.
In a $d$-regular tree, the number of nodes at distance $r$ from $v$ is $d(d-1)^{r-1}$, yet all of their information reaches $v$ through $d$ edges and a vector of width $F$ (\cref{fig:oversquash}).
The effect can be quantified by the sensitivity of a node representation to the input features of a distant node.

\begin{proposition}[Sensitivity bound~\citep{topping2022understanding}]\label{prop:sensitivity}
Consider an $r$-layer MPNN of the form $\hv_i^{(\ell+1)}=\UPD\big(\hv_i^{(\ell)},\sum_{j}\hat{A}_{ij}\,\MSG(\hv_j^{(\ell)})\big)$, with $\hat{\A}$ the normalized adjacency operator~\cref{eq:renorm}, whose message and update functions are Lipschitz with $|\nabla\MSG|\le\alpha$ and $|\nabla\UPD|\le\beta$. Then for nodes $i,s$ at distance $r$,
\begin{equation}\label{eq:sensitivity}
  \Big\|\frac{\partial\hv_i^{(r)}}{\partial\xv_s}\Big\|\le(\alpha\beta)^{r}\,\big(\hat{\A}^{r}\big)_{is}.
\end{equation}
\end{proposition}

Since $(\hat{\A}^{r})_{is}$ is small when many paths compete for the same bottleneck edges, \cref{eq:sensitivity} shows that influence decays with topological bottlenecks.
This was made precise via the Balanced Forman curvature~\citep{topping2022understanding}, the Cheeger constant and spectral gap~\citep{karhadkar2023fosr}, commute times and effective resistance~\citep{black2023understanding}, and Ollivier--Ricci curvature~\citep{nguyen2023revisiting}.
\citet{digiovanni2023over} show that sensitivity depends jointly on width, depth, and topology, and later work shows that over-squashing also limits expressive power in a quantifiable way~\citep{digiovanni2024how}.

Mitigation techniques include \emph{graph rewiring}---adding edges guided by curvature (SDRF)~\citep{topping2022understanding}, spectral gap (FoSR)~\citep{karhadkar2023fosr}, Ollivier--Ricci curvature~\citep{nguyen2023revisiting}, diffusion~\citep{klicpera2019diffusion}, learnable differentiable rewiring~\citep{arnaizrodriguez2022diffwire}, or locality-preserving schemes~\citep{barbero2024locality}---multi-hop and delayed message passing~\citep{gutteridge2023drew}, virtual nodes, and global attention in graph transformers~\citep{ying2021transformers,rampasek2022recipe,shirzad2023exphormer}.
Over-smoothing and over-squashing are in tension: adding edges to relieve bottlenecks tends to accelerate smoothing, which makes a joint treatment necessary~\citep{giraldo2023trade,nguyen2023revisiting}; see \citep{attali2026graph} for a recent survey of rewiring.

\paragraph{Scalability.}\label{sec:scalability-classical}
Full-batch training of an $L$-layer GCN on a graph with $n$ nodes and $m$ edges requires $\mathcal{O}(LnF)$ memory for the activations (plus $\mathcal{O}(m)$ for the graph and $\mathcal{O}(LF^2)$ for the weights) and $\mathcal{O}(L(mF+nF^2))$ time per epoch.
Mini-batching is non-trivial because of the \emph{neighbourhood explosion}: a single node's $L$-layer computation graph contains up to $\mathcal{O}(d_{\max}^{L})$ nodes, which can cover a large part of the graph on small-world networks.
Benchmark graphs such as those in the Open Graph Benchmark~\citep{hu2020open,hu2021ogb} (up to $10^{8}$ nodes and $10^{9}$ edges, e.g., \texttt{ogbn-papers100M}) make this limitation concrete, and a systematic study of large-scale training is given by~\citet{duan2022comprehensive}.
Scalable approaches can be grouped as follows.
\begin{itemize}[leftmargin=1.5em,itemsep=2pt,topsep=2pt]
  \item \emph{Sampling-based}: node-wise neighbour sampling~\citep{hamilton2017inductive}, layer-wise importance sampling~\citep{chen2018fastgcn}, variance reduction with historical activations~\citep{chen2018stochastic}, graph-partition (cluster) batching~\citep{chiang2019cluster}, and subgraph sampling with bias correction~\citep{zeng2020graphsaint}; see also~\citep{serafini2021scalable}.
  \item \emph{Historical embeddings}: reuse stale embeddings for out-of-batch neighbours~\citep{fey2021gnnautoscale}.
  \item \emph{Decoupling/precomputation}: the propagation is separated from the learnable transformation and precomputed once, e.g., SGC~\citep{wu2019simplifying}, SIGN~\citep{rossi2020sign}, PPRGo~\citep{bojchevski2020scaling}, GBP~\citep{chen2020scalable}, and Correct-and-Smooth~\citep{huang2021combining}. For example, SGC collapses $K$ layers into $\mathrm{softmax}(\hat{\A}^{K}\X\W)$, with $\hat{\A}^{K}\X$ computed once in $\mathcal{O}(KmF)$.
  \item \emph{Distributed training}: partitioning across machines or GPUs, e.g., DistDGL~\citep{zheng2020distdgl}; surveyed in~\citep{shao2024distributed}.
  \item \emph{Efficient transformers}: linear or sparse attention to avoid the $\mathcal{O}(n^2)$ cost of dense attention~\citep{wu2022nodeformer,wu2023sgformer,shirzad2023exphormer}.
\end{itemize}
At the frontier, GNN-based foundation models trained across many graphs push the demand for scalable, expressive graph learning further~\citep{mao2024position}.
Sampling and decoupling reduce cost but typically trade away exactness or expressivity (e.g., decoupled models forgo learned multi-layer nonlinearity), so scalability remains a trade-off rather than a solved problem.

\begin{table}[t]
\centering
\caption{Summary of the three classical failure modes of message-passing GNNs and representative remedies.}
\label{tab:failure-modes}
\renewcommand{\arraystretch}{1.3}
\footnotesize
\begin{tabular}{@{}p{2.2cm}p{4.2cm}p{4.0cm}p{4.0cm}@{}}
\toprule
\textbf{Failure mode} & \textbf{Symptom and mechanism} & \textbf{Diagnostic} & \textbf{Representative remedies} \\
\midrule
Over-smoothing & Representations converge as depth grows; repeated low-pass filtering~\cref{eq:smoothing} & Dirichlet energy, MAD~\citep{chen2020measuring} & Residuals~\citep{chen2020simple}; normalization~\citep{zhao2020pairnorm}; DropEdge~\citep{rong2020dropedge}; PDE/oscillator models~\citep{chamberlain2021grand,rusch2022graph} \\
Over-squashing & Exponentially many messages compressed through bottlenecks~\cref{eq:sensitivity} & Jacobian sensitivity; curvature; effective resistance~\citep{topping2022understanding,black2023understanding} & Rewiring~\citep{karhadkar2023fosr,barbero2024locality}; multi-hop~\citep{gutteridge2023drew}; global attention~\citep{rampasek2022recipe} \\
Scalability & Neighbourhood explosion; $\mathcal{O}(n^2)$ attention & Memory and time vs.\ $n,m,L$ & Sampling~\citep{hamilton2017inductive,zeng2020graphsaint}; decoupling~\citep{wu2019simplifying,rossi2020sign}; distributed~\citep{zheng2020distdgl}; linear attention~\citep{wu2022nodeformer} \\
\bottomrule
\end{tabular}
\end{table}

\paragraph{Summary.}
Classical GNNs combine local, permutation-equivariant message passing with learned transformations; their power is capped by the 1-WL test unless costly higher-order or structural extensions are used; and they suffer from over-smoothing with depth, over-squashing across bottlenecks, and cost growth with graph size (\cref{tab:failure-modes}).
These limitations are the classical counterpart of the motivations given in \cref{sec:intro-motivation}.
Whether quantum circuits can offer a practical or provable advantage on any of them is an open question that we examine critically in \cref{sec:ch-theory,sec:fut-advantage}.

\subsection{Quantum Computing Essentials}\label{sec:qc-essentials}
% ===========================================================================
% Section 2.2  Quantum Computing Essentials  (included from main.tex)
% Written with the QGNN use case in mind: every concept is tied to graphs.
% ===========================================================================

This subsection collects the quantum-computing concepts needed to read the rest of the survey, with a deliberate bias toward what matters for graph learning: how a graph or its features become a quantum state (\cref{sec:vqc}), which circuit primitives mimic message passing (\cref{sec:gates}), and why training such circuits at the size of realistic graphs is hard (\cref{sec:trainability-bg}).
We follow standard references for the fundamentals~\citep{nielsen2010quantum} and for the variational/NISQ viewpoint~\citep{preskill2018quantum,bharti2022noisy,cerezo2021variational,benedetti2019parameterized}.

% ---------------------------------------------------------------------------
\subsubsection{Qubits, States, Entanglement, and Measurement}\label{sec:qubits}
% ---------------------------------------------------------------------------

\paragraph{Qubits and states.}
A qubit is a two-level quantum system with state $\ket{\psi}=\alpha\ket{0}+\beta\ket{1}\in\C^2$, $|\alpha|^2+|\beta|^2=1$, defined up to a global phase.
A register of $n$ qubits lives in $\Hil=(\C^2)^{\otimes n}\cong\C^{2^n}$, with computational basis $\{\ket{x}:x\in\{0,1\}^n\}$, and a pure state is
\begin{equation}\label{eq:state}
  \ket{\psi}=\sum_{x\in\{0,1\}^n}c_x\ket{x},\qquad\sum_x|c_x|^2=1 .
\end{equation}
The exponential size of $\Hil$ is the source of the hoped-for advantages, and also of the difficulty of simulating, reading out, and training quantum models classically.
Mixed states, which arise from noise or from discarding part of a system, are described by density operators $\rho\succeq0$, $\Tr\rho=1$.
For graph learning, the most common design choice is a \emph{node--qubit correspondence}: a graph with $n$ nodes is mapped to (at least) $n$ qubits, so that qubit $v$ carries the information of node $v$.
Alternatives use $\mathcal{O}(\log n)$ qubits to index nodes in superposition, which is more compact but costlier to prepare and read out (\cref{sec:vqc}).

\paragraph{Superposition, interference, and entanglement.}
Superposition lets \cref{eq:state} carry amplitudes on exponentially many basis states, while interference---the addition of complex amplitudes---is what a circuit exploits to shape output probabilities.
A pure state of two subsystems $A,B$ is \emph{entangled} if it cannot be written as $\ket{\psi_A}\otimes\ket{\psi_B}$.
The Bell state $\tfrac{1}{\sqrt2}(\ket{00}+\ket{11})$ is the canonical example, and bipartite entanglement is quantified by the entanglement entropy $S(\rho_A)=-\Tr(\rho_A\log\rho_A)$ of the reduced state $\rho_A=\Tr_B\ket{\psi}\bra{\psi}$.
Entanglement is the natural carrier of \emph{relational} information, which is why it is central to QGNNs: edges between nodes can be represented by entangling gates between the corresponding qubits.
A particularly direct link is provided by \emph{graph states}~\citep{hein2006entanglement}.
For a graph $\G=(\V,\E)$ one prepares
\begin{equation}\label{eq:graphstate}
  \ket{\G}=\prod_{(u,v)\in\E}\mathrm{CZ}_{uv}\ket{+}^{\otimes n},
\end{equation}
where $\mathrm{CZ}$ is the controlled-$Z$ gate (\cref{sec:gates}) and $\ket{+}=(\ket{0}+\ket{1})/\sqrt2$.
Since all $\mathrm{CZ}$ gates commute, the preparation is independent of edge ordering, and \cref{eq:graphstate} is a state whose entanglement structure is exactly the graph.
Graph states are stabilizer states, hence efficiently simulable by themselves~\citep{gottesman1998heisenberg}, which is why practical QGNNs interleave them with non-Clifford, trainable gates.

\paragraph{Measurement.}
Information is read out by measuring an observable, a Hermitian operator $O$.
Measuring in the computational basis returns $x$ with probability $|c_x|^2$ (Born rule), and for a general observable the expectation value is
\begin{equation}\label{eq:expval}
  \expval{O}{\psi}=\bra{\psi}O\ket{\psi}=\Tr\!\big(O\,\ket{\psi}\bra{\psi}\big).
\end{equation}
Measurement is probabilistic and destructive, so \cref{eq:expval} is not accessed directly; it is \emph{estimated} from $S$ repeated runs (``shots''), with statistical error $\mathcal{O}(1/\sqrt{S})$.
Estimating a quantity to additive precision $\varepsilon$ therefore needs $\mathcal{O}(\varepsilon^{-2})$ shots, which makes gradient-based training, with many parameters and many observables, expensive.
In quantum graph models, typical readouts are single-qubit Pauli expectations $\expval{Z_v}{\psi}\in[-1,1]$ for node-level tasks, and global or averaged observables such as $\tfrac1n\sum_v Z_v$ for graph-level tasks; the latter is permutation-invariant by construction (\cref{sec:equivariance}).
Because the full state $\ket{\psi}$ cannot be extracted efficiently (tomography costs exponentially many shots), a QGNN output is necessarily a low-dimensional classical summary, which constrains what can be read out per circuit execution and is often a bottleneck in practice.

% ---------------------------------------------------------------------------
\subsubsection{Quantum Gates and Circuits}\label{sec:gates}
% ---------------------------------------------------------------------------

\paragraph{Gates.}
Computation on $\Hil$ is performed by unitary operators ($U^\dagger U=\Id$).
Single-qubit gates include the Pauli matrices $X,Y,Z$, the Hadamard gate $H$, and the parametrized rotations
\begin{equation}\label{eq:rot}
  R_P(\theta)=e^{-i\theta P/2}=\cos\tfrac\theta2\,\Id-i\sin\tfrac\theta2\,P,\qquad P\in\{X,Y,Z\},
\end{equation}
whose angles $\theta$ are the trainable (or data-dependent) parameters in variational models.
Two-qubit gates create entanglement, e.g., $\mathrm{CNOT}$, $\mathrm{CZ}$, and the Ising-type rotations $R_{ZZ}(\theta)=e^{-i\theta Z\otimes Z/2}$, $R_{XX}(\theta)$, $R_{YY}(\theta)$.
Single-qubit rotations together with a single entangling gate form a universal gate set, i.e., they approximate any unitary to arbitrary precision~\citep{nielsen2010quantum}.
Many gates are Clifford gates ($H$, $S$, $\mathrm{CNOT}$, $\mathrm{CZ}$), which map Pauli operators to Pauli operators; circuits made only of Clifford gates can be simulated efficiently on a classical computer~\citep{gottesman1998heisenberg}, so non-Clifford rotations like \cref{eq:rot} are essential.

\paragraph{Circuits and cost.}
A quantum circuit is a sequence of gates applied to an initial state (usually $\ket{0}^{\otimes n}$), followed by measurement.
Its \emph{depth} (number of sequential layers) and \emph{gate count} determine the runtime and the accumulated noise; on hardware with restricted qubit connectivity, additional SWAP gates are needed to implement interactions between non-adjacent qubits (transpilation), which inflates depth.
This is directly relevant for graphs: an edge $(u,v)$ in the input graph requires interaction between qubits $u$ and $v$, but the physical coupling map of the device is generally different from the input graph.
Dense or irregular graphs therefore incur a large routing overhead, while sparse, low-degree graphs can be embedded more cheaply.

\begin{figure}[t]
\centering
\begin{tikzpicture}[
  every node/.style={font=\footnotesize},
  g/.style={draw,fill=blue!10,minimum width=8mm,minimum height=6mm,inner sep=1pt},
  ctrl/.style={circle,fill,inner sep=1.3pt},
  wire/.style={thick}]
  % wires
  \foreach \y/\l in {0/{v_1},-0.9/{v_2},-1.8/{v_3},-2.7/{v_4}}{
    \node[left] at (-0.2,\y) {$\ket{0}_{\l}$};
    \draw[wire] (0,\y) -- (10.2,\y);}
  % Hadamards
  \foreach \y in {0,-0.9,-1.8,-2.7}{\node[g] at (0.7,\y) {$H$};}
  % edge layer: edges (1,2),(2,3),(3,4),(1,4)? keep a path + one chord (2,4)
  \begin{scope}[xshift=2.0cm]
    \draw[orange!70!black,thick] (0,0)--(0,-0.9); \node[ctrl,orange!70!black] at (0,0){}; \node[ctrl,orange!70!black] at (0,-0.9){};
    \draw[orange!70!black,thick] (0.7,-0.9)--(0.7,-1.8); \node[ctrl,orange!70!black] at (0.7,-0.9){}; \node[ctrl,orange!70!black] at (0.7,-1.8){};
    \draw[orange!70!black,thick] (1.4,-1.8)--(1.4,-2.7); \node[ctrl,orange!70!black] at (1.4,-1.8){}; \node[ctrl,orange!70!black] at (1.4,-2.7){};
    \node[orange!70!black] at (0.7,0.55) {edge layer $U_\E(\gamma)$};
  \end{scope}
  % trainable rotations
  \foreach \y in {0,-0.9,-1.8,-2.7}{\node[g,fill=green!12] at (5.4,\y) {$R_X(\beta)$};}
  \node[green!40!black] at (5.4,0.55) {mixer / trainable};
  % second block (repeat)
  \foreach \y in {0,-0.9,-1.8,-2.7}{\node[g,fill=gray!15] at (7.0,\y) {$\cdots$};}
  % measurement
  \foreach \y in {0,-0.9,-1.8,-2.7}{
    \node[draw,fill=white,minimum width=7mm,minimum height=6mm,inner sep=1pt] at (9.2,\y) {$\langle Z\rangle$};}
  \draw[decorate,decoration={brace,amplitude=4pt,mirror}] (0.35,-3.15) -- (7.5,-3.15) node[midway,below=4pt] {$L$ repeated layers (shared parameters)};
\end{tikzpicture}
\caption{A schematic graph-structured parametrized circuit on $n=4$ qubits (one per node): Hadamard layer, an edge-wise entangling layer following the graph connectivity as in \cref{eq:edgelayer}, trainable single-qubit rotations, and per-qubit measurements of $\expval{Z_v}{}$. Shared parameters across all edges and nodes keep the circuit independent of node ordering.}
\label{fig:circuit}
\end{figure}
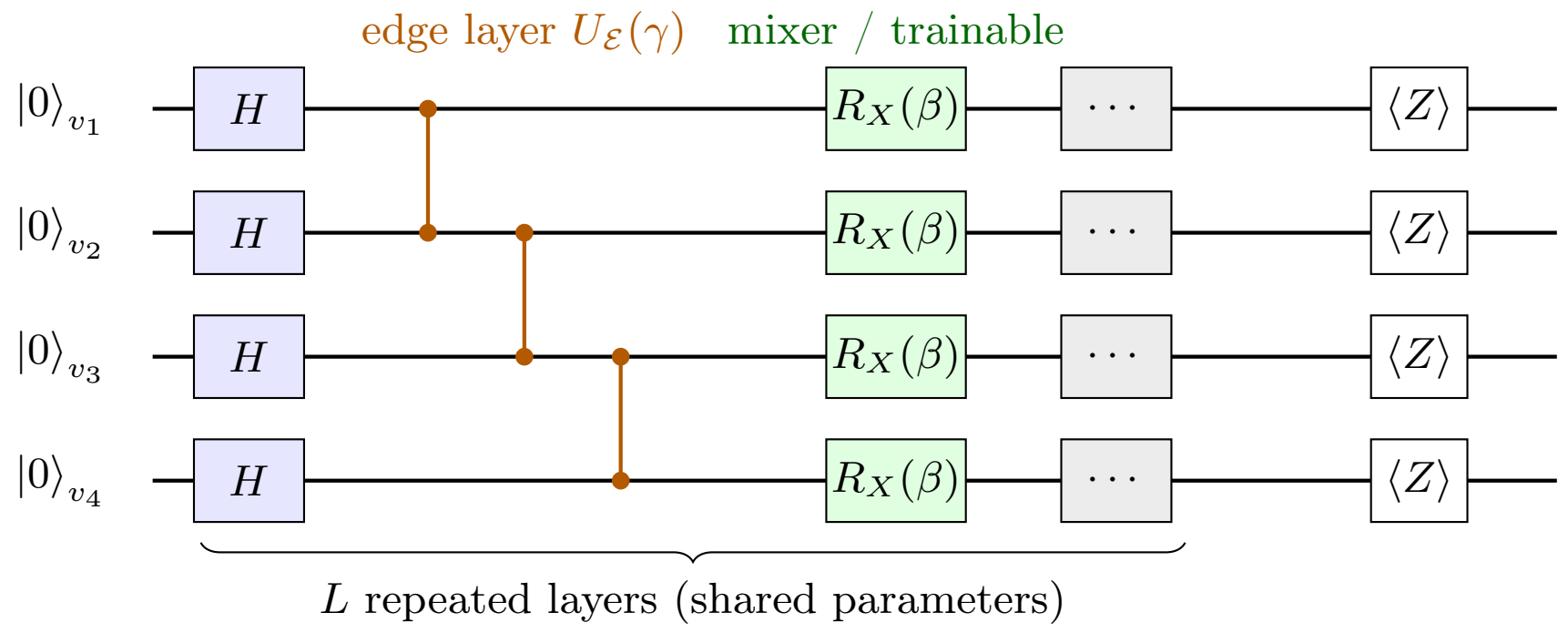

\paragraph{Graph-structured gates.}
Two gate patterns recur throughout the QGNN literature and are useful to introduce now.
The first is the \emph{edge-wise interaction layer},
\begin{equation}\label{eq:edgelayer}
  U_{\E}(\gamma)=\prod_{(u,v)\in\E}e^{-i\gamma\,Z_uZ_v},
\end{equation}
which couples neighbouring qubits with a shared parameter and is invariant under relabeling of the nodes (a shared parameter over all edges), a property exploited by the equivariant designs of \cref{sec:tax-eqgc}.
This is the same cost-Hamiltonian layer used in the quantum approximate optimization algorithm (QAOA)~\citep{farhi2014quantum}.
The second is the evolution under a graph-dependent Hamiltonian $H_{\G}$ (e.g., an Ising or Heisenberg Hamiltonian whose couplings are the adjacency matrix, $H_{\G}=\sum_{(u,v)\in\E}A_{uv}\,Z_uZ_v$ plus single-qubit terms):
\begin{equation}\label{eq:hamevo}
  U(t)=e^{-itH_{\G}},\qquad e^{-it(H_1+H_2)}\approx\Big(e^{-i\frac{t}{r}H_1}e^{-i\frac{t}{r}H_2}\Big)^{r}\ \text{(Trotter)}.
\end{equation}
Such continuous-time evolution is the quantum counterpart of graph diffusion and of the continuous-time GNNs of \cref{sec:failure-modes}, and underlies the Hamiltonian-based QGNNs of \cref{sec:birth-hamiltonian}.
Product-formula (Trotter) approximations implement it with gate count growing with the number of Hamiltonian terms and the required accuracy~\citep{lloyd1996universal}.
\Cref{fig:circuit} sketches a simple graph-structured circuit combining these ingredients.

% ---------------------------------------------------------------------------
\subsubsection{Variational Quantum Circuits and Data Encoding}\label{sec:vqc}
% ---------------------------------------------------------------------------

\paragraph{Variational quantum circuits.}
On near-term (noisy, shallow-depth) hardware, quantum learning relies on \emph{variational quantum algorithms} (VQAs)~\citep{cerezo2021variational,bharti2022noisy}: a parametrized circuit $U(\boldsymbol\theta)$ is executed on a quantum device and its parameters are optimized by a classical optimizer.
The first instances were the variational quantum eigensolver (VQE)~\citep{peruzzo2014variational,mcclean2016theory} and QAOA~\citep{farhi2014quantum}; quantum machine learning models use the same pattern~\citep{benedetti2019parameterized,mitarai2018quantum,schuld2019quantum}.
A VQC-based model for input $\xv$ is
\begin{equation}\label{eq:vqc}
  f_{\boldsymbol\theta}(\xv)=\Tr\!\Big[O\;U(\boldsymbol\theta)\,\rho(\xv)\,U^\dagger(\boldsymbol\theta)\Big],\qquad \rho(\xv)=S(\xv)\ket{0}\bra{0}^{\otimes n}S(\xv)^\dagger,
\end{equation}
where $S(\xv)$ is the data-encoding circuit, $U(\boldsymbol\theta)=\prod_{\ell=1}^{L}W_\ell(\boldsymbol\theta_\ell)$ is the trainable ansatz, and $O$ is the measured observable.
Training minimizes an empirical loss $\mathcal{L}(\boldsymbol\theta)=\frac1N\sum_i\ell\big(f_{\boldsymbol\theta}(\xv_i),y_i\big)$ with a classical optimizer.
For rotation gates of the form \cref{eq:rot}, gradients can be obtained on hardware with the \emph{parameter-shift rule}~\citep{mitarai2018quantum,schuld2019evaluating},
\begin{equation}\label{eq:psr}
  \partial_{\theta_k}f_{\boldsymbol\theta}=\tfrac12\Big[f_{\boldsymbol\theta+\frac\pi2\mathbf{e}_k}-f_{\boldsymbol\theta-\frac\pi2\mathbf{e}_k}\Big],
\end{equation}
which requires two circuit evaluations per parameter, so a full gradient costs $\mathcal{O}(P)$ circuit executions for $P$ parameters, each with $\mathcal{O}(\varepsilon^{-2})$ shots.
The choice of ansatz $U(\boldsymbol\theta)$ ranges from \emph{hardware-efficient} layers of single-qubit rotations and native entanglers~\citep{kandala2017hardware}, which are easy to run but unstructured, to \emph{problem-inspired} ansatzes whose structure follows the physics or symmetries of the task~\citep{farhi2014quantum,larocca2022group,meyer2023exploiting}.
In QGNNs, the graph provides exactly such structure, as in \cref{eq:edgelayer}.
The expressive capacity of an ansatz can be quantified, for example, by the expressibility measure of~\citet{sim2019expressibility}, and, for the \emph{data} dependence, by the Fourier-series view discussed below.
Learning-theoretic questions---generalization bounds that scale with the number of trainable gates~\citep{caro2022generalization}, the role of the data~\citep{huang2021power}, and whether standard generalization arguments suffice for quantum models~\citep{gilfuster2024understanding}---are active research topics and we return to them in \cref{sec:ch-theory}.
Training itself can be NP-hard in the worst case~\citep{bittel2021training}.

\paragraph{Data encoding.}
Encoding classical data $\xv$ into a quantum state is the interface between classical graphs and quantum circuits, and it largely determines both the cost and the expressivity of the model~\citep{schuld2019quantum,havlicek2019supervised,schuld2021effect}.
The recent survey of~\citet{sammartino2026feature} compares feature-encoding methods in terms of resource cost, expressivity, and noise robustness; we summarize the families most used for graphs.
\begin{itemize}[leftmargin=1.5em,itemsep=2pt,topsep=2pt]
  \item \emph{Basis encoding.} A bit string $x\in\{0,1\}^n$ is mapped to $\ket{x}$. For graphs, this encodes the adjacency structure bit by bit: $\binom n2$ qubits represent the edge indicators of a simple undirected graph, which is exact but inefficient and rarely used in learning models.
  \item \emph{Angle (rotation) encoding.} Each feature $x_j$ sets a rotation angle, $S(\xv)=\bigotimes_{j}R_Y(x_j)$ (cf. \cref{eq:rot}), producing a product state with one qubit per feature.
  For node features $\xv_v\in\R^F$, one uses $F$ qubits per node, or one qubit per node when $F=1$. The circuit is shallow (depth $\mathcal{O}(1)$), hence noise-robust, but the qubit count grows linearly with $nF$, which limits graph size.
  \item \emph{Dense angle encoding.} Two features per qubit, e.g., via $R_Y(x_{2j})R_Z(x_{2j+1})$, halving the qubit count at the price of a slightly deeper encoder.
  \item \emph{Amplitude encoding.} A normalized vector $\mathbf{x}\in\R^{2^n}$ is stored in the amplitudes, $\ket{\mathbf{x}}=\|\mathbf{x}\|^{-1}\sum_{i}x_i\ket{i}$, so that $\mathcal{O}(\log d)$ qubits hold $d$ numbers.
  For graphs, the adjacency matrix ($n^2$ entries) needs only $\mathcal{O}(\log n)$ qubits and the node-feature matrix $\mathcal{O}(\log(nF))$ qubits. This is the main source of the claimed exponential compression, but exact preparation of a generic state costs $\mathcal{O}(2^n)$ gates~\citep{mottonen2005transformation,plesch2011quantum}, and loading classical data efficiently usually assumes a quantum RAM~\citep{aaronson2015read}. Without such assumptions, the cost of state preparation can cancel the advantage, and noise limits the usable depth, which is why shallow angle-based encodings are preferred on current hardware.
  \item \emph{Data re-uploading.} The data are encoded repeatedly, interleaved with trainable layers, $U(\boldsymbol\theta;\xv)=W_L\,S(\xv)\cdots W_1\,S(\xv)\,W_0$~\citep{perezsalinas2020data}. This increases the expressivity of small circuits, and it is the standard way to introduce nonlinearity in the input, since \cref{eq:vqc} is otherwise linear in $\rho(\xv)$.
  \item \emph{Graph-structure encodings.} Besides the features, the \emph{topology} can enter the circuit through (i) the pattern of entangling gates, as in graph states \cref{eq:graphstate} and \cref{eq:edgelayer}; (ii) the coefficients of a Hamiltonian built from $\A$ or $\Lap$, as in \cref{eq:hamevo}; or (iii) block-encoding or quantum-walk representations of $\A$, discussed in \cref{sec:scalability}.
  Options (i)--(ii) fix the circuit to a specific graph, so the circuit must be recompiled for each input graph, whereas parametrized encodings of node features can be shared across graphs.
\end{itemize}
\Cref{tab:encodings} compares these options for a graph with $n$ nodes and $F$ features per node.

\begin{table}[t]
\centering
\caption{Data-encoding strategies for graph inputs. Qubit counts are for storing $\X\in\R^{n\times F}$ (features) or the topology; ``depth'' refers to the encoding circuit only. Amplitude encoding assumes a generic dense input without quantum RAM.}
\label{tab:encodings}
\renewcommand{\arraystretch}{1.3}
\footnotesize
\begin{tabular}{@{}p{2.6cm}p{2.9cm}p{2.8cm}p{3.0cm}p{2.5cm}@{}}
\toprule
\textbf{Encoding} & \textbf{Qubits} & \textbf{Depth} & \textbf{Noise robustness} & \textbf{Typical use} \\
\midrule
Basis & $\binom{n}{2}$ (edges) & $\mathcal{O}(1)$ & high & exact structure \\
Angle & $\mathcal{O}(nF)$ & $\mathcal{O}(1)$ & high & node features, NISQ QGNNs \\
Dense angle & $\mathcal{O}(nF/2)$ & $\mathcal{O}(1)$ & high & compact node features \\
Amplitude & $\mathcal{O}(\log nF)$ & $\mathcal{O}(nF)$ (generic) & low & small graphs; compression \\
Re-uploading & as angle & $\mathcal{O}(L)$ & medium & expressivity \\
Graph-state / Hamiltonian & $\mathcal{O}(n)$ & $\mathcal{O}(|\E|)$ & medium & topology \\
\bottomrule
\end{tabular}
\end{table}

\paragraph{Expressivity through the encoding.}
A model of the form \cref{eq:vqc} with Pauli-rotation encoding gates is a truncated Fourier series in the data, $f_{\boldsymbol\theta}(x)=\sum_{\omega\in\Omega}c_\omega(\boldsymbol\theta)e^{i\omega x}$, where the accessible frequencies $\Omega$ are determined by the eigenvalues of the encoding generators and the number of re-uploadings, while the coefficients are controlled by the trainable circuit~\citep{schuld2021effect}.
Equivalently, encoding maps the data into a high-dimensional Hilbert space and the model is a linear function there, which relates variational models to kernel methods~\citep{schuld2019quantum,havlicek2019supervised}.
This viewpoint will be used in \cref{sec:universal} to discuss universality of QGNNs.

% ---------------------------------------------------------------------------
\subsubsection{Trainability: Barren Plateaus, Noise, and NISQ Constraints}\label{sec:trainability-bg}
% ---------------------------------------------------------------------------

\paragraph{Barren plateaus.}
The central obstacle to scaling VQCs is the \emph{barren plateau} (BP) phenomenon: for many circuits, the variance of the loss gradient vanishes exponentially in the number of qubits,
\begin{equation}\label{eq:bp}
  \Var_{\boldsymbol\theta}\big[\partial_{\theta_k}\mathcal{L}\big]\in\mathcal{O}\!\big(b^{-n}\big),\qquad b>1,
\end{equation}
so that the landscape is exponentially flat almost everywhere, and an exponentially large number of shots is needed to resolve a descent direction~\citep{mcclean2018barren}.
\citet{mcclean2018barren} showed this for random circuits that form $2$-designs; $\Var\propto 2^{-n}$ in that case.
The phenomenon has since been characterized in several forms, which are important for QGNN design:
\begin{itemize}[leftmargin=1.5em,itemsep=2pt,topsep=2pt]
  \item \emph{Cost locality and depth.} Global cost functions (observables acting on all qubits) exhibit BPs even for shallow circuits, whereas local costs yield at most polynomially vanishing gradients for logarithmic depth~\citep{cerezo2021cost}. Graph-level readouts such as $\tfrac1n\sum_vZ_v$ are sums of single-qubit terms and therefore local, which is favourable.
  \item \emph{Expressivity.} Highly expressive ansatzes, those approaching a unitary design, concentrate the cost~\citep{holmes2022connecting}, and entanglement between the visible and hidden parts of the circuit also induces vanishing gradients (\citealp{mcclean2018barren}; see the review below).
  \item \emph{A unifying theory.} Gradient variances are governed by the dimension of the dynamical Lie algebra generated by the circuit's gates: small algebras avoid BPs, large ones suffer them~\citep{ragone2024lie,fontana2024characterizing}. The recent review of~\citet{larocca2025review} collects these sources (expressivity, entanglement, locality, noise) into a single picture.
  \item \emph{Noise-induced barren plateaus.} Hardware noise can itself flatten the landscape: for local Pauli noise the gradient decays exponentially in the circuit depth~\citep{wang2021noise}, independent of the ansatz.
  \item \emph{Local minima and optimization hardness.} Even without BPs, landscapes can contain many poor local minima and training can be computationally hard~\citep{bittel2021training}.
\end{itemize}
Two classes of countermeasures are used.
(i) \emph{Initialization and training heuristics}: near-identity or structured initializations that start in a region with non-vanishing gradients~\citep{grant2019initialization}, and layer-wise training~\citep{skolik2021layerwise}.
(ii) \emph{Structured architectures}: models with restricted expressivity, such as quantum convolutional networks, which provably avoid BPs~\citep{pesah2021absence}, and models that encode the symmetries of the task~\citep{larocca2022group,meyer2023exploiting,schatzki2024theoretical}.
The latter is the geometric/equivariant route central to the QGNN literature: a circuit whose gates commute with the permutation action explores a much smaller subspace of $\Hil$ than a generic circuit and, under suitable conditions, has a gradient variance decaying only polynomially~\citep{schatzki2024theoretical}.
Permutation-equivariant constructions for graphs and images have been proposed in~\citep{skolik2023equivariant,das2024permutation} and, for molecular learning, compared against rotation-equivariant and classical networks in~\citep{biswas2025permeq}.

\paragraph{Trainability versus classical simulability.}
A caveat that has become prominent recently is that \emph{avoiding BPs may make a model easy to simulate classically}.
Many mechanisms that give polynomially large gradients---restricting the circuit to a small Lie algebra, using local observables with shallow depth, or limiting entanglement---also admit efficient classical algorithms for estimating the loss~\citep{cerezo2025does}.
Cerezo et al.\ argue that, for broad families of BP-free models, the loss landscape can be classically simulated, possibly after a data-acquisition phase on a quantum device; this does not prove that every trainable model is simulable, but it shifts the burden of proof onto claims of advantage.
Further work gives regimes in which trainability is provably preserved where no classical surrogate is known~\citep{meyer2025trainability}, and demonstrates efficient classical predictive surrogates for a 20-qubit chain on a programmable superconducting processor~\citep{liao2025demonstration}.
For graph learning, a recent preprint combines an implementation and trainability analysis of quantum graph convolutional networks with a dequantization argument based on low-rank simulability, and stresses that any end-to-end advantage also depends on data-loading, block-encoding, and readout assumptions~\citep{lo2026quantum}.
% VERIFY: confirm the exact statements of lo2026quantum (stable-rank condition, gradient-variance scaling) against the full text before adding them.
Because this work is very recent, we treat it as an indication of the type of analysis needed rather than as a settled result.
The relationship between dequantization and efficient classical sampling goes back to~\citet{tang2019quantum}; see also~\citet{schuld2022quantum} for the broader point that quantum advantage should not be the sole criterion for QML research, and~\citet{bowles2024better} for a large-scale benchmark showing that simple classical models often outperform their quantum counterparts on small-scale tasks.

\paragraph{NISQ constraints.}
Present devices are in the noisy intermediate-scale quantum (NISQ) regime~\citep{preskill2018quantum}: tens to a few hundred qubits with two-qubit gate errors of the order $10^{-3}$--$10^{-2}$, limited connectivity, and short coherence times.
A circuit of $G$ gates with average error $\epsilon$ retains a fidelity of roughly $e^{-\epsilon G}$, so the useful depth is $\mathcal{O}(1/\epsilon)$, which shapes QGNN design: shallow layers, few qubits, sparse graphs, and shared parameters.
For graph models, three practical consequences follow.
\begin{enumerate}[leftmargin=1.8em,itemsep=2pt,topsep=2pt]
  \item \emph{Qubit budget.} With the node--qubit mapping, only graphs with $n\lesssim$ the number of reliable qubits can be processed directly. Larger graphs require sampling, partitioning into subgraphs (ego-graphs), or compressed encodings.
  \item \emph{Connectivity and routing.} The input graph must be embedded on the device's coupling map (\cref{sec:gates}). Dense graphs incur SWAP overhead.
  \item \emph{Shot cost.} Each loss evaluation needs $\mathcal{O}(\varepsilon^{-2})$ shots, each gradient entry needs two evaluations \cref{eq:psr}, and each mini-batch needs $N_{\mathrm{batch}}$ graphs, so wall-clock time grows rapidly compared with GPU training.
\end{enumerate}
These constraints explain why most reported QGNN results are on small graphs, simulated circuits, or hybrid classical--quantum pipelines (\cref{sec:tax-hybrid}) and why fair comparison with classical GNNs is difficult (\cref{sec:ch-fair}).

\paragraph{Error mitigation and error correction.}
Two strategies address noise.
\emph{Error mitigation} reduces the bias of expectation values at the cost of additional circuit executions and without encoding logical qubits, using techniques such as zero-noise extrapolation and probabilistic error cancellation~\citep{temme2017error,endo2018practical}; the review of~\citet{cai2023quantum} covers these methods and their sampling overhead, which grows exponentially with the circuit size in the worst case.
Mitigation has enabled experiments on 127-qubit devices at a scale beyond exact classical simulation of the specific task~\citep{kim2023evidence}, but this does not by itself establish an advantage for learning tasks.
\emph{Error correction} encodes logical qubits in many physical qubits so that errors can be detected and removed.
A recent landmark experiment showed that increasing the surface-code distance reduced the logical error rate, i.e., operation below the error-correction threshold~\citep{acharya2025quantum}.
Roadmaps for the next stage---early fault-tolerant machines with only a few logical qubits and tight limits on connectivity, latency, and control---stress that overheads and decoding remain the main bottlenecks and caution against over-reading small demonstrations~\citep{palsberg2026computer}.
Hence, for the foreseeable future, QGNNs are best viewed as NISQ-era models whose practical value must be demonstrated under realistic noise, and as candidates for early fault-tolerant hardware afterwards (\cref{sec:fut-ft}).

\begin{table}[t]
\centering
\caption{Main obstacles to scaling variational quantum graph models and the mitigation strategies discussed in this subsection.}
\label{tab:obstacles}
\renewcommand{\arraystretch}{1.3}
\footnotesize
\begin{tabular}{@{}p{2.4cm}p{4.6cm}p{4.0cm}p{3.7cm}@{}}
\toprule
\textbf{Obstacle} & \textbf{Cause} & \textbf{Mitigation} & \textbf{Caveat} \\
\midrule
Barren plateaus & Expressive/global/entangled circuits; large Lie algebra~\citep{mcclean2018barren,cerezo2021cost,ragone2024lie} & Local costs; symmetric/equivariant ansatz; structured initialization~\citep{schatzki2024theoretical,grant2019initialization} & BP-free may be classically simulable~\citep{cerezo2025does} \\
Noise & Gate errors; decoherence~\citep{wang2021noise} & Shallow circuits; error mitigation~\citep{cai2023quantum} & Mitigation overhead grows with size \\
Data loading & Amplitude encoding costs $\mathcal{O}(2^n)$~\citep{mottonen2005transformation} & Shallow angle or dense-angle encoding~\citep{sammartino2026feature} & More qubits for angle encoding \\
Shot cost & $\mathcal{O}(\varepsilon^{-2})$ samples per estimate & Gradient-free or Fisher-aware optimizers; batching & Slow wall-clock training \\
Connectivity & Mismatch of graph and device coupling map & Sparse graphs; subgraph partitioning & Routing overhead \\
\bottomrule
\end{tabular}
\end{table}

\paragraph{Summary and link to the rest of the survey.}
In summary, a quantum graph model is a parametrized unitary acting on qubits that represent nodes (or nodes in superposition), whose graph awareness enters through the encoding and through the entangling pattern, and whose scalability is limited by qubit count, noise, shot cost, and barren plateaus.
Symmetry-restricted architectures offer the clearest route to trainability, but a growing body of results warns that restriction may also bring classical simulability.
These two threads---equivariance and the trainability--simulability tension---are developed in \cref{sec:symmetry,sec:expr-train-scale,sec:ch-theory}.

\subsection{Notation and Unified Formalism}\label{sec:notation}
% ===========================================================================
% Section 2.3  Notation and Unified Formalism
% (included from main.tex via \input; the \subsection heading and
%  \label{sec:notation} live in main.tex)
% ===========================================================================

\Cref{sec:classical-gnn,sec:qc-essentials} introduced the classical and quantum ingredients separately.
This subsection joins them.
We first fix the conventions that connect the two vocabularies (\cref{sec:notation-conv}), then give a single definition of a quantum graph neural network that all architectures of \cref{sec:taxonomy} instantiate (\cref{sec:unified-qgnn}), transfer the symmetry principle of \cref{def:equiv} to quantum circuits (\cref{sec:notation-sym}), and finally fix the resource parameters used in every complexity statement of the survey (\cref{sec:notation-cost}).
The formalism follows the equivariant-circuit viewpoint of~\citet{mernyei2022equivariant}, which was the first to treat several earlier proposals~\citep{verdon2019quantum} as instances of one class, and the taxonomy of the review of~\citet{ceschini2024graphs}; we do not repeat the definitions of \cref{sec:graph-reps,sec:equivariance,sec:qubits,sec:gates}.

% ---------------------------------------------------------------------------
\subsubsection{Conventions and Notational Bridge}\label{sec:notation-conv}
% ---------------------------------------------------------------------------

\paragraph{Conventions.}
Graph symbols ($\G,\V,\E,\Nb,\A,\Dg,\Lap,\X,\xv_v,\hv_v^{(\ell)}$) keep their meaning from \cref{sec:graph-reps,sec:mp-gcn}, and $\Sym$, $\Perm$ their meaning from \cref{sec:equivariance}.
Quantum symbols follow \cref{sec:qubits,sec:gates}: states are kets, operators are plain capitals, and $Z_v$ denotes the Pauli-$Z$ operator on the qubit associated with node $v$ (identity on all other qubits).
Two clarifications avoid clashes between the two parts of the background.
First, \cref{sec:qc-essentials} used $n$ for the number of qubits of a generic register; from here on $n=|\V|$ always counts \emph{nodes} and $n_q$ counts \emph{qubits}, with $n_q=n$ in the node--qubit mapping, $n_q=\mathcal{O}(\log n)$ in index (amplitude-type) encodings, and $n_q=\mathcal{O}(nF)$ with several qubits per node (\cref{tab:encodings}).
Second, $\tr$ is the trace of a classical matrix and $\Tr$ the trace of an operator on $\Hil$; $\Id$ is the identity of whatever dimension the context fixes.
Estimates obtained from $S$ measurement shots carry a hat, $\widehat{\expval{O}{\psi}}$, while exact expectation values do not.

\paragraph{A classical--quantum dictionary.}
\Cref{tab:bridge} lists how each component of the message-passing template~\cref{eq:mpnn,eq:readout} is realized in the quantum models reviewed later.
The correspondence is functional, not literal: a quantum layer does not expose per-node hidden vectors, because measuring destroys the state; the quantity that plays the role of $\hv_v^{(\ell)}$ is the reduced state $\rho_v^{(\ell)}=\Tr_{\V\setminus\{v\}}\big[\rho^{(\ell)}\big]$ of node $v$, which can be accessed only through expectation values.

\begin{table}[t]
\centering
\caption{Dictionary between the classical message-passing template and its quantum counterparts in this survey.}
\label{tab:bridge}
\renewcommand{\arraystretch}{1.3}
\footnotesize
\begin{tabular}{@{}p{3.1cm}p{4.9cm}p{5.5cm}@{}}
\toprule
\textbf{Component} & \textbf{Classical (\cref{eq:mpnn,eq:readout})} & \textbf{Quantum (this survey)} \\
\midrule
Node input & $\xv_v\in\R^F$ & angle/amplitude-encoded qubit(s) of node $v$ via $S(\G,\X)$ \\
Node state & $\hv_v^{(\ell)}$ & reduced state $\rho_v^{(\ell)}$; accessed via $\expval{Z_v}{}$ \\
Topology & $\A$, $\Nb(v)$, $\Lap$ & entangling pattern $U_{\E}(\gamma)$ \cref{eq:edgelayer}; couplings of $H_{\G}$ \cref{eq:hamevo} \\
$\MSG$, $\AGG$ & learnable message, invariant aggregator & shared-parameter two-qubit gates over $\E$ (commuting or symmetrized) \\
$\UPD$ & MLP / recurrent cell & trainable single-qubit rotations $R_P(\theta)$ \cref{eq:rot} \\
$\READOUT$ & sum / mean / learned pooling & symmetric observable, e.g.\ $\tfrac1{n}\sum_v Z_v$, then classical head $g_{\boldsymbol\phi}$ \\
Receptive field & $L$-hop neighbourhood & $L$-hop light cone of the observable (\cref{prop:lightcone}) \\
Training & backpropagation & parameter-shift rule \cref{eq:psr}, or backpropagation in simulation \\
\bottomrule
\end{tabular}
\end{table}

% ---------------------------------------------------------------------------
\subsubsection{A Unified Definition of QGNNs}\label{sec:unified-qgnn}
% ---------------------------------------------------------------------------

\begin{definition}[Quantum graph neural network]\label{def:qgnn}
Let $(\G,\X)$ be an attributed graph with $n$ nodes. A \emph{QGNN} is a map
\begin{equation}\label{eq:generic-qgnn}
  f_{\boldsymbol\theta,\boldsymbol\phi}(\G,\X)
  \;=\;
  g_{\boldsymbol\phi}\Big(\widehat{\expval{O_1}{\psi}},\dots,\widehat{\expval{O_K}{\psi}}\Big),
  \qquad
  \ket{\psi}=U_{\G}(\boldsymbol\theta)\,S(\G,\X)\,\ket{0}^{\otimes n_q},
\end{equation}
consisting of (i) a data-encoding circuit $S(\G,\X)$ (\cref{sec:vqc}); (ii) a graph-conditioned trainable circuit $U_{\G}(\boldsymbol\theta)=W_{L,\G}(\boldsymbol\theta_L)\cdots W_{1,\G}(\boldsymbol\theta_1)$ whose layers use the graph through the pattern of entangling gates or the couplings of a Hamiltonian (\cref{eq:edgelayer,eq:hamevo}); (iii) observables $O_1,\dots,O_K$ whose expectation values are estimated from $S$ shots; and (iv) a classical post-processing map $g_{\boldsymbol\phi}$, possibly the identity.
\end{definition}

\Cref{eq:generic-qgnn} extends the variational model~\cref{eq:vqc} by letting both $S$ and $U$ depend on the graph.
Task levels are obtained through the choice of observables: node-level outputs use $O_v=Z_v$ (one per node), edge-level outputs use $O_{uv}=Z_uZ_v$, and graph-level outputs use a symmetric combination such as $\tfrac1n\sum_vZ_v$.
The three regimes used to organize \cref{sec:taxonomy} are special cases of \cref{def:qgnn}:
\begin{itemize}[leftmargin=1.5em,itemsep=2pt,topsep=2pt]
  \item \emph{Fully quantum}: $g_{\boldsymbol\phi}$ is fixed or trivial, and all learning happens in $\boldsymbol\theta$ (e.g., the Hamiltonian-based models of \cref{sec:birth-hamiltonian,sec:tax-physics}).
  \item \emph{Hybrid}: $g_{\boldsymbol\phi}$ is a classical network trained jointly with $\boldsymbol\theta$, and/or $\X$ is replaced by learned features from a classical encoder, so that a quantum layer substitutes the aggregation or update of a classical GNN (\cref{sec:tax-hybrid}).
  \item \emph{Quantum-inspired}: the structure of \cref{eq:generic-qgnn} is kept but the circuit is evaluated on classical hardware; such models are excluded from the claims of quantum advantage but are included when they are used as baselines.
\end{itemize}
Where a statement depends on whether $\boldsymbol\phi$ is trained jointly with $\boldsymbol\theta$, we say so explicitly.

\paragraph{Receptive field.}
A classical $L$-layer MPNN depends on the $L$-hop neighbourhood of a node (\cref{sec:mp-gcn}).
The same locality holds for graph-conditioned circuits whose two-qubit gates act only on edges of $\G$, which is what makes the node--qubit mapping a faithful analogue of message passing.

\begin{proposition}[Light cone]\label{prop:lightcone}
Let $U_{\G}(\boldsymbol\theta)=W_L\cdots W_1$ in the node--qubit mapping, where each layer $W_\ell$ is a product of single-qubit gates and two-qubit gates acting only on pairs $(u,v)\in\E$. Then $U_{\G}^\dagger Z_v U_{\G}$ is supported on the qubits of nodes at graph distance at most $L$ from $v$. Consequently $\expval{Z_v}{\psi}$ depends on $\X$ only through the $L$-hop neighbourhood of $v$ when $S$ is a product of single-qubit encodings.
\end{proposition}
\begin{proof}[Proof sketch]
Conjugating an operator supported on a set $T$ of qubits by a layer $W_\ell$ enlarges the support at most to $T$ together with the neighbours of $T$ in $\G$, because gates that do not touch $T$ cancel. Applying this for $\ell=L,\dots,1$ gives support in the $L$-hop ball of $v$. A product-state encoding then ensures that only the features of qubits in that ball enter the expectation value.
\end{proof}

\noindent The converse does not hold: with non-local encodings or with Hamiltonian evolution under dense couplings, the effective receptive field can be larger, which is relevant for over-squashing (\cref{sec:failure-modes}).

% ---------------------------------------------------------------------------
\subsubsection{Symmetry for Quantum Graph Circuits}\label{sec:notation-sym}
% ---------------------------------------------------------------------------

\Cref{def:equiv} is stated for functions of $(\A,\X)$ and so applies to \cref{eq:generic-qgnn} as it stands; what is new in the quantum case is a condition on the \emph{circuit} that guarantees it.
In the node--qubit mapping, a permutation $\pi\in\Sym$ acts on $\Hil$ through the unitary $R_\pi\ket{x_1\cdots x_n}=\ket{x_{\pi^{-1}(1)}\cdots x_{\pi^{-1}(n)}}$ that moves the content of qubit $v$ to qubit $\pi(v)$; it satisfies $R_\pi Z_vR_\pi^\dagger=Z_{\pi(v)}$ and leaves $\ket{0}^{\otimes n}$ unchanged.

\begin{figure}[t]
    \centering
    \includegraphics[width=\textwidth]{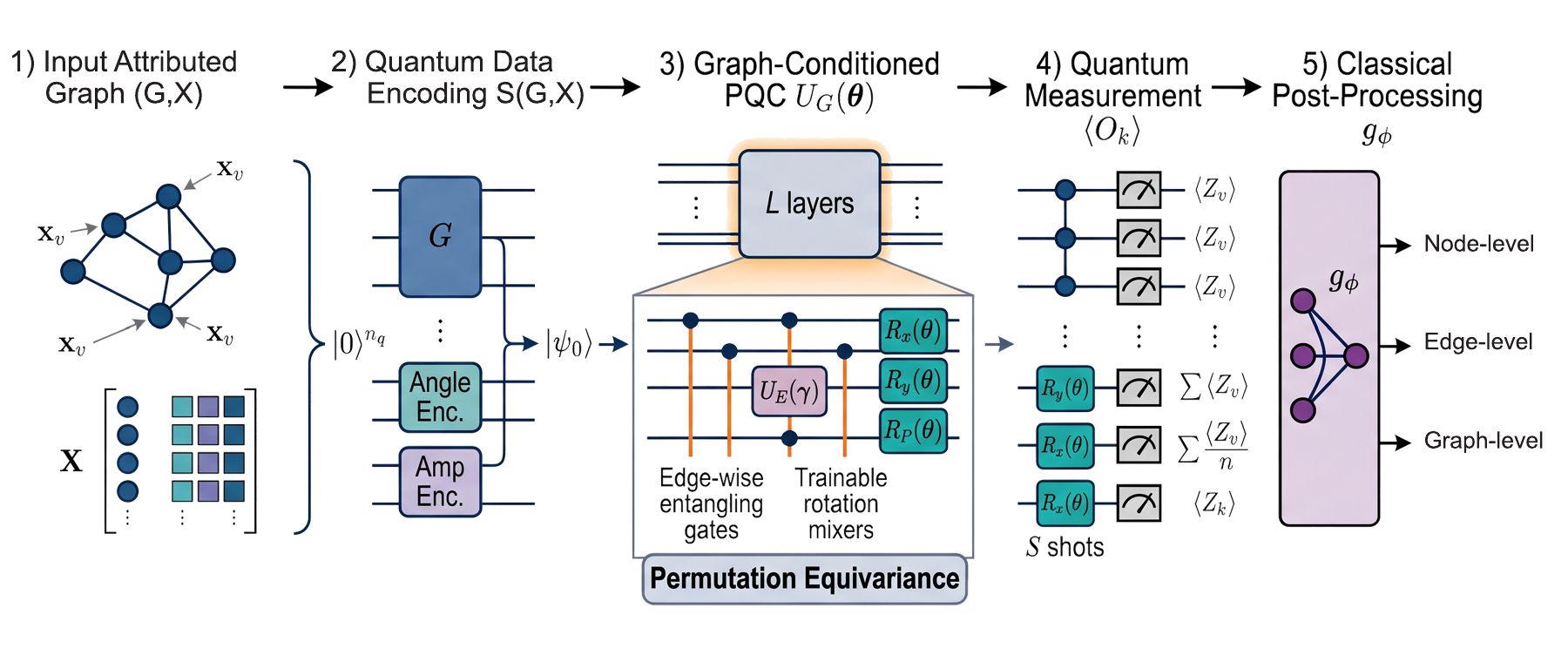}
    \caption{Unified architecture of a Quantum Graph Neural Network (\cref{def:qgnn}, \cref{eq:generic-qgnn}). (1) An input attributed graph $(\G,\X)$ is mapped via (2) a data-encoding circuit $S(\G,\X)$ into an $n_q$-qubit register. (3) An $L$-layer graph-conditioned variational circuit $U_{\G}(\boldsymbol\theta)$ alternates edge-wise entangling gates $U_{\E}(\gamma)$ with trainable single-qubit rotations. (4) Observables $\expval{O_k}{\psi}$ are estimated from $S$ shots for node-, edge-, or graph-level tasks and passed to (5) an optional classical head $g_{\boldsymbol\phi}$, while respecting permutation equivariance (\cref{def:qequiv}).}
    \label{fig:unified-qgnn-formalism}
\end{figure}
\begin{definition}[Equivariant quantum graph circuit]\label{def:qequiv}
A pair $(S,U)$ of encoder and graph-conditioned circuit is \emph{permutation-equivariant} if, writing $\pi\cdot(\G,\X)=(\Perm\A\Perm^{\top},\Perm\X)$,
\begin{equation}\label{eq:circ-equiv}
  S(\pi\cdot(\G,\X))\,\ket{0}^{\otimes n}=R_\pi\,S(\G,\X)\,\ket{0}^{\otimes n},
  \qquad
  U_{\pi\cdot\G}(\boldsymbol\theta)=R_\pi\,U_{\G}(\boldsymbol\theta)\,R_\pi^\dagger
  \qquad\forall\pi\in\Sym,
\end{equation}
up to a global phase.
\end{definition}

\begin{proposition}[Equivariance of QGNN outputs]\label{prop:qgnn-equiv}
If $(S,U)$ satisfies \cref{eq:circ-equiv}, then $\ket{\psi(\pi\cdot(\G,\X))}=R_\pi\ket{\psi(\G,\X)}$ and hence $\expval{Z_{\pi(v)}}{\psi(\pi\cdot(\G,\X))}=\expval{Z_v}{\psi(\G,\X)}$ for all $v$. Node-level outputs built from $\{Z_v\}$ are therefore permutation-equivariant, and graph-level outputs built from a symmetric function of $\{Z_v\}$ (e.g.\ $\tfrac1n\sum_vZ_v$) are permutation-invariant, in the sense of \cref{def:equiv}.
\end{proposition}
\begin{proof}[Proof sketch]
The first claim is \cref{eq:circ-equiv} applied to $\ket{\psi}=U_{\G}S\ket{0}$. For the second, $\bra{\psi}R_\pi^\dagger Z_{\pi(v)}R_\pi\ket{\psi}=\bra{\psi}Z_v\ket{\psi}$ using $R_\pi^\dagger Z_{\pi(v)}R_\pi=Z_v$. The post-processing $g_{\boldsymbol\phi}$ must itself be equivariant or invariant for the property to survive end to end.
\end{proof}

\noindent The edge layer \cref{eq:edgelayer} satisfies the second part of \cref{eq:circ-equiv} because all its factors commute and share one parameter, so $R_\pi U_{\E}(\gamma)R_\pi^\dagger=U_{\pi\cdot\E}(\gamma)$ regardless of edge order.
Layers built from \emph{non-commuting} edge terms, for instance Trotterized evolution~\cref{eq:hamevo} with non-commuting couplings, depend on the ordering of the factors and are equivariant only if the construction is symmetrized or the terms commute; this is a recurring design constraint in \cref{sec:tax-eqgc,sec:sym-perm}.
Two caveats apply to \cref{prop:qgnn-equiv}.
First, it concerns exact expectation values; with finite $S$ the estimates are equivariant in distribution, not sample by sample.
Second, in index encodings with $n_q=\mathcal{O}(\log n)$ the permutation acts on the computational basis as $\ket{v}\mapsto\ket{\pi(v)}$ rather than by moving qubits, and \cref{eq:circ-equiv} must be restated with that representation; the group-theoretic form of this idea is developed in \cref{sec:symmetry}.
As in \cref{sec:equivariance}, equivariance says nothing about expressive power: the limits of 1-WL stated there apply to the \emph{classical} MPNN family and should not be transferred to \cref{def:qgnn} without proof (\cref{sec:universal}).

% ---------------------------------------------------------------------------
\subsubsection{Resource Parameters and Cost Model}\label{sec:notation-cost}
% ---------------------------------------------------------------------------

All complexity statements in the survey use the quantities of \cref{tab:notation}.
Unless stated otherwise, cost is counted in circuit executions.
For a dataset of $N$ graphs, a model with $P$ trainable parameters, $K$ distinct measurement bases, and $S$ shots per basis, one epoch of gradient evaluation with the parameter-shift rule \cref{eq:psr} costs
\begin{equation}\label{eq:epoch-cost}
  \mathcal{C}_{\mathrm{epoch}} \;=\; 2\,P\,K\,S\,N\ \text{circuit executions},
\end{equation}
while the circuit itself has depth growing linearly in $L$ and gate count $\mathcal{O}(L(n+m))$ for a node--qubit model with one single-qubit gate per node and one two-qubit gate per edge per layer, before routing overhead on restricted hardware (\cref{sec:gates}).
Because observables diagonal in the same basis, such as all $Z_v$, can be estimated from the same shots, node-level readout over all nodes still has $K=1$; global parameter sharing across nodes and edges, as in \cref{eq:edgelayer}, makes $P=\mathcal{O}(L)$ independent of $n$.
Both properties are used when the scalability of specific architectures is compared in \cref{sec:tax-compare,sec:scalability}.

\begin{table}[t]
\centering
\caption{Main notation used in the survey (graph and quantum symbols are defined in \cref{sec:graph-reps,sec:qubits,sec:gates}).}
\label{tab:notation}
\renewcommand{\arraystretch}{1.25}
\footnotesize
\begin{tabular}{@{}p{3.3cm}p{10.2cm}@{}}
\toprule
\textbf{Symbol} & \textbf{Meaning} \\
\midrule
$\G=(\V,\E)$, $n$, $m$ & Graph, number of nodes, number of edges \\
$\A$, $\Dg$, $\Lap$, $\Nb(v)$ & Adjacency, degree, Laplacian, neighbourhood of node $v$ \\
$\X\in\R^{n\times F}$, $\xv_v$ & Node features, feature vector of node $v$ \\
$\Sym$, $\Perm$, $R_\pi$ & Permutation group, permutation matrix, qubit-permuting unitary on $\Hil$ \\
$\hv_v^{(\ell)}$, $\rho_v^{(\ell)}$ & Classical hidden vector; reduced quantum state of node $v$ after layer $\ell$ \\
$n_q$ & Number of qubits ($n_q=n$ in the node--qubit mapping) \\
$\ket{\psi}$, $\rho$, $\Hil$ & Pure state, density operator, Hilbert space of $n_q$ qubits \\
$X_v,Y_v,Z_v$ & Pauli operators on the qubit of node $v$ \\
$R_P(\theta)$ & Pauli rotation $e^{-i\theta P/2}$ \\
$S(\G,\X)$ & Data-encoding circuit \\
$U_{\G}(\boldsymbol\theta)$ & Graph-conditioned trainable circuit with $L$ layers and parameters $\boldsymbol\theta\in\R^P$ \\
$O_k$, $\widehat{\expval{O_k}{\psi}}$ & Observable and its shot-based estimate \\
$g_{\boldsymbol\phi}$ & Classical post-processing network with parameters $\boldsymbol\phi$ \\
$L$, $P$, $K$, $S$, $N$ & Layers, trainable parameters, measurement bases, shots per basis, number of graphs \\
$\mathcal{L}$ & Empirical loss, $\frac1N\sum_i\ell(f(\G_i,\X_i),y_i)$ \\
\bottomrule
\end{tabular}
\end{table}

\paragraph{Summary.}
\Cref{def:qgnn} describes every model of \cref{sec:taxonomy} as an encoder, a graph-conditioned circuit, symmetric observables, and an optional classical head; \cref{def:qequiv,prop:qgnn-equiv} give the circuit-level condition for \cref{def:equiv}; and \cref{eq:epoch-cost} fixes how costs are compared.
These three elements recur in \cref{sec:taxonomy,sec:symmetry,sec:expr-train-scale}.

% ===========================================================================
\section{The Birth of Quantum Graph Neural Networks}\label{sec:birth}
% ===========================================================================
This section covers the first proposals that combined graph-structured data with quantum learning, from the quantum-inspired models of 2018 to the circuit-based and quantum-data models of 2021--2022, before equivariance became the organizing principle of the field (\cref{sec:taxonomy}).
We read ``first'' functionally: a work is included if it fixes a way for the graph to enter a quantum model---through Hamiltonian couplings, gate patterns, amplitudes, or the data itself---that later architectures reuse.
\Cref{sec:birth-hamiltonian} presents the Hamiltonian-based QGNN of \citet{verdon2019quantum}, the first model to carry the name, together with its recent follow-ups; \cref{sec:birth-early} places it among contemporaneous quantum-inspired, circuit-based, and quantum-data approaches and extracts the design lessons that the taxonomy builds on.

\subsection{Hamiltonian-Based QGNN: The Foundational Model}\label{sec:birth-hamiltonian}
% ===========================================================================
% Section 3.1  Hamiltonian-Based QGNN: The Foundational Model
% (included from main.tex via \input; heading and \label{sec:birth-hamiltonian}
%  live in main.tex)
% ===========================================================================

The first model to be named a \emph{quantum graph neural network} is due to \citet{verdon2019quantum}.
Its starting point is physical rather than data-driven: a graph is read as the interaction pattern of a \emph{quantum network}, i.e., one quantum subsystem per vertex, and learning means tuning the Hamiltonians whose evolution acts on that network.
Three ideas introduced there recur throughout the survey: the graph enters the circuit only through the \emph{interaction topology} of a Hamiltonian; tying parameters across edges, nodes, or time steps yields convolutional and recurrent variants whose parameter count does not grow with the graph; and the same ansatz serves both quantum data with graph structure and classical graph data.
This subsection first states the general ansatz (\cref{sec:ham-ansatz}), then its three specializations (\cref{sec:ham-variants}) and the tasks on which they were illustrated (\cref{sec:ham-tasks}), and finally places the model in the formalism of \cref{sec:unified-qgnn} and traces its influence on later work, including recent follow-ups (\cref{sec:ham-legacy}).

% ---------------------------------------------------------------------------
\subsubsection{The General Ansatz}\label{sec:ham-ansatz}
% ---------------------------------------------------------------------------

Assign to every node $v\in\V$ a Hilbert space $\Hil_v$ (a qubit in the node--qubit mapping of \cref{sec:notation-conv}) and let $\Hil=\bigotimes_{v}\Hil_v$.
The \emph{general QGNN ansatz} is an alternating sequence of evolutions under $Q$ graph-structured Hamiltonians, repeated $T$ times~\citep{verdon2019quantum} (written here in the notation of this survey, following the restatement of~\citet{larocca2022group}):
\begin{equation}\label{eq:verdon-ansatz}
  U_{\G}(\boldsymbol\eta,\boldsymbol\theta)\;=\;\prod_{t=1}^{T}\Big[\prod_{q=1}^{Q}e^{-i\,\eta_{tq}\,H_q(\boldsymbol\theta)}\Big],
  \qquad
  H_q(\boldsymbol\theta)\;=\;\sum_{(j,k)\in\E}\sum_{r}W_{qrjk}\,O_j^{(qr)}\!\otimes O_k'^{(qr)}\;+\;\sum_{v\in\V}\sum_{r}B_{qrv}\,M_v^{(qr)} .
\end{equation}
Here $O_j^{(qr)},O_k'^{(qr)},M_v^{(qr)}$ are Hermitian operators acting on the single node indicated by the subscript, $W_{qrjk}$ and $B_{qrv}$ are variational coefficients of two-body (edge) and one-body (node) terms, and $\eta_{tq}$ are variational evolution times.
The interaction graph of every $H_q$ is exactly $\G$, which is how topology enters.
In the node--qubit mapping the operators are Pauli operators, and the edge layer \cref{eq:edgelayer} is the special case $O=O'=Z$, $M=0$, $W\equiv\gamma/\eta$; the Hamiltonian $H_{\G}$ of \cref{eq:hamevo} is one choice of $H_q$.
Within \cref{def:qgnn}, \cref{eq:verdon-ansatz} is a graph-conditioned circuit $U_{\G}(\boldsymbol\theta)$ with $L=TQ$ layers, each layer being the exponential of a graph Hamiltonian.

\paragraph{Compilation.}
The exponentials in \cref{eq:verdon-ansatz} are not native gates and are implemented by product formulas as in \cref{eq:hamevo}~\citep{lloyd1996universal}, with an approximation error that decreases with the number of Trotter steps, or directly as analog evolution on hardware whose native interactions match $\G$.
A first-order product formula over the $n$ node terms and $m$ edge terms uses $\mathcal{O}(n+m)$ gates per step; if all terms of a layer commute (for example, only $Z_uZ_v$ couplings), the edge terms can be grouped by an edge colouring into at most $\Delta+1$ rounds of disjoint gates, where $\Delta$ is the maximum degree, provided the device coupling map contains $\G$.
Otherwise the routing overhead of \cref{sec:gates} applies.

\paragraph{Symmetry of tied layers.}
Because each $H_q$ sums over \emph{all} edges and nodes, the ansatz is permutation-covariant as soon as its coefficients do not depend on node or edge identity.

\begin{proposition}[Tied Hamiltonian layers are equivariant]\label{prop:ham-equiv}
Let $\Hil_v\cong\Hil_1$ for all $v$, let $W_{qrjk}=W_{qr}$ and $B_{qrv}=B_{qr}$ for all $j,k,v$, and let the edge operators satisfy $O^{(qr)}=O'^{(qr)}$ (so that the edge term is symmetric in its endpoints for undirected graphs). Then $e^{-i\eta H_q(\G)}$ satisfies the second condition of \cref{eq:circ-equiv}, i.e., $e^{-i\eta H_q(\pi\cdot\G)}=R_\pi\,e^{-i\eta H_q(\G)}R_\pi^\dagger$ for all $\pi\in\Sym$.
\end{proposition}
\begin{proof}[Proof sketch]
Conjugation by $R_\pi$ relabels the tensor factors, $R_\pi O_j R_\pi^\dagger=O_{\pi(j)}$, so $R_\pi H_q(\G)R_\pi^\dagger=\sum_{(j,k)\in\E}W\,O_{\pi(j)}\otimes O_{\pi(k)}+\sum_vB\,M_{\pi(v)}=H_q(\pi\cdot\G)$, because $\pi$ maps $\E$ bijectively to the edge set of $\pi\cdot\G$. Equivariance of the exponential follows since $R_\pi$ is unitary.
\end{proof}

\noindent The statement concerns the exact exponential; once compiled with a product formula, equivariance holds only if the factors commute or the formula is symmetrized (\cref{sec:notation-sym}).
Tying also fixes the parameter count: it depends on $T$, $Q$, and the number of operator types, but not on $n$ or $m$, which is what \cref{eq:epoch-cost} rewards.

% ---------------------------------------------------------------------------
\subsubsection{Recurrent, Convolutional, and Spectral Specializations}\label{sec:ham-variants}
% ---------------------------------------------------------------------------

\Cref{eq:verdon-ansatz} is the most general member of the family, and generic parametrized circuits of this size are prone to barren plateaus (\cref{sec:trainability-bg}).
\citet{verdon2019quantum} therefore add structure by tying parameters; \cref{tab:verdon} summarizes the resulting variants.

\paragraph{Quantum graph recurrent neural networks (QGRNN).}
Here the temporal parameters are shared across repetitions, $\eta_{tq}\mapsto\eta_q$, in analogy with weight sharing in recurrent networks.
Each repetition is then a Trotter step of the evolution $e^{-i\Delta H_{\mathrm{eff}}}$ with $H_{\mathrm{eff}}=\Delta^{-1}\sum_q\eta_qH_q$ and $\Delta=\sum_q|\eta_q|$, so the ansatz is specialized to learning effective Hamiltonian dynamics of systems living on a graph.

\paragraph{Quantum graph convolutional neural networks (QGCNN).}
Here the parameters are tied \emph{spatially}, $W_{qrjk}\mapsto W_{qr}$ and $B_{qrv}\mapsto B_{qr}$, so that all edges and all nodes share one set of coefficients, as the filter of a classical GCN is shared across the graph.
The resulting family is a graph-structured instance of the quantum alternating operator ansatz~\citep{hadfield2019quantum}, itself a generalization of QAOA~\citep{farhi2014quantum}, and by \cref{prop:ham-equiv} it respects the permutation symmetry that makes classical message passing well defined (\cref{prop:mp-equiv}).

\paragraph{Quantum spectral graph convolutional neural networks (QSGCNN).}
A third variant takes inspiration from a continuous-variable form of the quantum approximate optimization ansatz and, in the Heisenberg picture, reproduces the layer structure of Laplacian-based graph convolutions: alternating message-passing, node-update, and nonlinear steps~\citep{verdon2019quantum}.
This variant is the quantum counterpart of the spectral view of \cref{sec:mp-gcn}.
(It should not be confused with the quantum-inspired QSGCNN of \citet{bai2023learning}, discussed in \cref{sec:early-inspired}, which shares the abbreviation.)

\begin{table}[t]
\centering
\caption{Members of the Hamiltonian-based QGNN family of~\citet{verdon2019quantum} in the notation of \cref{eq:verdon-ansatz}. ``Use'' lists the kind of task the family is aimed at, not benchmark results.}
\label{tab:verdon}
\renewcommand{\arraystretch}{1.3}
\footnotesize
\begin{tabular}{@{}p{2.5cm}p{3.7cm}p{3.5cm}p{4.3cm}@{}}
\toprule
\textbf{Variant} & \textbf{Parameter tying} & \textbf{Classical analogue} & \textbf{Use} \\
\midrule
General QGNN & none; $\eta_{tq},W_{qrjk},B_{qrv}$ free & general message passing with per-node weights & quantum processes on a network \\
QGRNN & $\eta_{tq}\mapsto\eta_q$ (over repetitions) & recurrent network, weights shared over time & learning effective graph Hamiltonian dynamics \\
QGCNN & $W_{qrjk}\mapsto W_{qr}$, $B_{qrv}\mapsto B_{qr}$ & GCN / MPNN, filter shared over the graph & permutation-invariant learning, e.g.\ graph classification \\
QSGCNN & spectral form, CV-QAOA-inspired & Laplacian-based spectral convolution & spectral tasks, e.g.\ clustering \\
\bottomrule
\end{tabular}
\end{table}

% ---------------------------------------------------------------------------
\subsubsection{Illustrated Tasks}\label{sec:ham-tasks}
% ---------------------------------------------------------------------------

\citet{verdon2019quantum} illustrate the framework on four tasks, using numerical simulation: learning the Hamiltonian dynamics of a quantum system (the QGRNN is demonstrated on the effective real-time dynamics of an Ising model on a graph), learning to create multipartite entanglement in a quantum network, unsupervised spectral clustering, and supervised graph-isomorphism classification.
Two features of this list matter for the rest of the survey.
First, the first two tasks have \emph{quantum} inputs or outputs, so the model is an object for quantum information processing before it is a competitor of classical GNNs; only the latter two are classical graph-learning tasks in the sense of \cref{sec:graph-reps}.
Second, the graph is fixed inside the circuit: changing the graph means changing the Hamiltonian and hence recompiling, so there is no single circuit that is applied to graphs of different size or topology (cf.\ the remark on graph-specific encodings in \cref{sec:vqc}).

% ---------------------------------------------------------------------------
\subsubsection{Place in the Formalism and Influence on Later Work}\label{sec:ham-legacy}
% ---------------------------------------------------------------------------

\paragraph{Place in the formalism.}
In the terms of \cref{def:qgnn}, the Hamiltonian-based QGNN is a \emph{fully quantum} model with a graph-conditioned circuit built from $e^{-i\eta H}$ layers and local observables as readout.
By \cref{prop:lightcone}-type arguments, the influence of a node on a local observable spreads at a finite speed set by the evolution times and couplings, so that the total evolution time plays the role that depth plays in the receptive field of a message-passing network; dense couplings or long evolution times enlarge it, which is the double-edged feature relevant to over-squashing (\cref{sec:failure-modes}).
Later authors classify this line of work as an \emph{analog} approach, based on global unitary dynamics, in contrast to digital variational circuits~\citep{faria2025inductive}.

\paragraph{Trainability and simulability.}
\citet{verdon2019quantum} motivate parameter tying by barren plateaus~\citep{mcclean2018barren}.
The modern picture makes this precise: restricting the circuit to a small dynamical Lie algebra avoids barren plateaus~\citep{ragone2024lie,larocca2025review}, and permutation-symmetric constructions such as \cref{prop:ham-equiv} are examples of such restrictions~\citep{larocca2022group,schatzki2024theoretical}.
The same restriction is also what raises the question of classical simulability~\citep{cerezo2025does}, which we take up for QGNNs in \cref{sec:bp,sec:ch-theory}.

\paragraph{Recent follow-ups.}
The Hamiltonian idea has continued to develop on three fronts.
\citet{thabet2024quantum} map graph topology onto qubit interactions and read positional encodings for classical graph transformers from the resulting quantum correlations; the quantum part is not trained, and the authors prove that some of these features are more expressive than relative random-walk probabilities for certain graphs.
\citet{wang2025qghnn} proposed the quantum graph Hamiltonian neural network (QGHNN), which treats the adjacency matrix as the coupling coefficients of a Pauli Hamiltonian and builds the circuit by decomposing the corresponding time-evolution operator into rotation and CNOT gates, trained on the expectation of that Hamiltonian.
Finally, \citet{faria2025inductive} point to the graph-specific circuit as the main obstacle of this lineage and propose digital, GraphSAGE-style aggregators whose qubit count depends on the feature dimension rather than on the graph; we return to this design in \cref{sec:tax-mp,sec:tax-hybrid}.

\paragraph{Summary.}
The Hamiltonian-based QGNN supplies the survey with its first template: topology as interaction pattern, tied parameters as inductive bias, and symmetry through shared coefficients (\cref{prop:ham-equiv}).
Its limitations, namely graph-specific compilation, dependence on connectivity or analog hardware, and simulation-level evidence, motivate the digital and equivariant designs of \cref{sec:taxonomy}, and the physics-inspired branch is developed further in \cref{sec:tax-physics}.

\subsection{Early Graph-Structured Quantum Learning}\label{sec:birth-early}
% ===========================================================================
% Section 3.2  Early Graph-Structured Quantum Learning
% (included from main.tex via \input; heading and \label{sec:birth-early}
%  live in main.tex)
% ===========================================================================

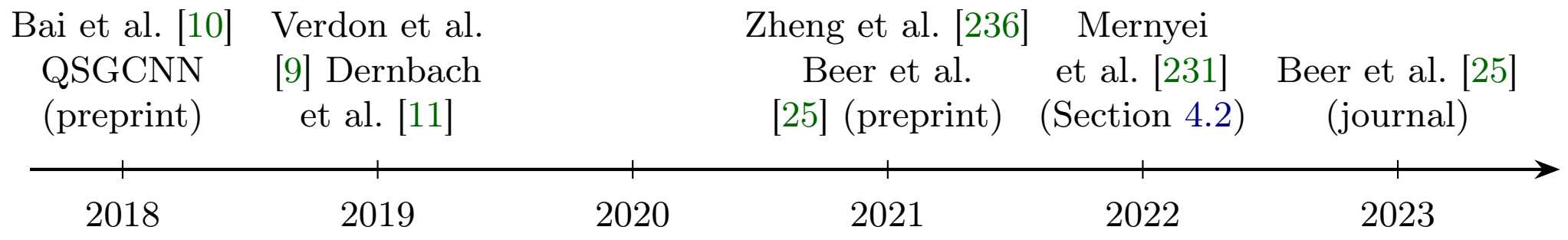
\begin{figure}[t]
\centering
\begin{tikzpicture}[every node/.style={font=\footnotesize}]
  \draw[thick,-{Stealth[length=2.2mm]}] (0,0) -- (13.2,0);
  % ticks at years 2018 ... 2023, 2.2 cm apart
  \foreach \x/\y in {0.8/2018, 3.0/2019, 5.2/2020, 7.4/2021, 9.6/2022, 11.8/2023}{
    \draw (\x,0.08) -- (\x,-0.08); \node[below=2pt] at (\x,-0.08) {\y};}
  \node[above=3pt,align=center,text width=2.6cm] at (0.8,0.05) {\citet{bai2023learning}\\ QSGCNN (preprint)};
  \node[above=3pt,align=center,text width=2.6cm] at (3.0,0.05) {\citet{verdon2019quantum} \citet{dernbach2019quantum}};
  \node[above=3pt,align=center,text width=2.6cm] at (7.4,0.05) {\citet{zheng2021quantum}\\ \citet{beer2023quantum} (preprint)};
  \node[above=3pt,align=center,text width=2.6cm] at (9.6,0.05) {\citet{mernyei2022equivariant}\\ (\cref{sec:tax-eqgc})};
  \node[above=3pt,align=center,text width=2.6cm] at (11.8,0.05) {\citet{beer2023quantum}\\ (journal)};
\end{tikzpicture}
\caption{Timeline of the early works discussed in \cref{sec:birth}, by first public appearance (journal versions for the last entry); the equivariant circuits of \citet{mernyei2022equivariant} open the taxonomy of \cref{sec:taxonomy}.}
\label{fig:timeline}
\end{figure}

The Hamiltonian-based QGNN of \cref{sec:birth-hamiltonian} did not appear in isolation.
Between 2018 and 2022 graph-structured data entered quantum learning along three routes that differ in \emph{what the graph is} and in \emph{where the quantum resources sit}: classical graphs processed by quantum-inspired operators that run on classical hardware (\cref{sec:early-inspired}), classical graphs processed by parametrized quantum circuits (\cref{sec:early-circuits}), and graphs whose vertices carry quantum states (\cref{sec:early-qdata}).
\Cref{tab:early} contrasts these works with the Hamiltonian-based model, and \cref{fig:timeline} places them in time.
We use the regimes of \cref{def:qgnn}: quantum-inspired models are included as historical context and as baselines, and no claim of quantum computational advantage applies to them.

% ---------------------------------------------------------------------------
\subsubsection{Quantum-Inspired Models: Quantum Walks and Vertex Information Propagation}\label{sec:early-inspired}
% ---------------------------------------------------------------------------

The earliest graph-learning models with a quantum label borrow the dynamics of quantum walks, which spread amplitude over a graph and replace the diffusion of a classical random walk (\cref{sec:mp-gcn}).

\paragraph{Quantum spatial graph convolution.}
\citet{bai2023learning} propose the quantum spatial graph convolutional neural network (QSGCNN) for graph classification, first posted in 2018.
The model first identifies transitive alignments between the vertices of different graphs, which maps graphs of arbitrary size to fixed-size vertex grids; a \emph{quantum spatial graph convolution} then extracts multi-scale vertex features through quantum information propagation between grid vertices, and a standard convolutional network is applied to the resulting grid.
The convolution leaves the vertex order of the grid unchanged, which is what allows the conventional CNN on top, and the authors argue that this reduces the information loss of SortPooling and SumPooling readouts and the tottering problem of spatial GCNs.
The model is described as quantum-inspired: it requires no quantum hardware.

\paragraph{Quantum walk neural networks.}
\citet{dernbach2019quantum} define a graph neural network whose diffusion operator is the evolution of a discrete-time quantum walk.
The walk's coin operators are not fixed but learned, and in the feature-dependent version they are generated from the features of neighbouring nodes, so that structurally identical graphs with different features produce different walks; the model addresses node-level and graph-level tasks.
In the vocabulary of \cref{tab:bridge}, the learned coin plays the role of $\MSG$ and the walk step that of $\AGG$.

Both models foreshadow ideas that reappear in quantum circuits: quantum-walk encodings of $\A$, taken up in \cref{sec:scalability}, and learned, feature-dependent propagation, taken up in \cref{sec:tax-mp}.
\citet{zheng2024quantum} later realized the QSGCNN of \citet{bai2023learning} as a model implementable on quantum circuits and analysed its trainability in terms of barren plateaus, which makes the lineage a bridge between the quantum-inspired and the circuit-based regimes.

% ---------------------------------------------------------------------------
\subsubsection{Quantum Circuits for Classical Graph Data}\label{sec:early-circuits}
% ---------------------------------------------------------------------------

The first circuit-based graph models after \citet{verdon2019quantum} copied the architecture of classical GCNs onto parametrized quantum circuits.
\citet{zheng2021quantum} encode graph data into quantum states by amplitude encoding (\cref{sec:vqc}), build the convolution layer from parametrized universal gates arranged according to the graph structure, and implement pooling by measurement, which reduces the feature dimension in the way classical pooling does; the model is evaluated on graph classification.
Compared with the Hamiltonian-based family, these designs are \emph{digital}: they use gate sequences rather than evolution under a Hamiltonian.
They also inherit the costs of the encoding.
As \citet{faria2025inductive} observe, encoding the whole graph (adjacency, nodes, and edges) into a single circuit makes circuit width and depth grow quickly with graph size, and the amplitude encoding on which these designs rely is expensive to prepare for generic data (\cref{tab:encodings}); the same authors make the corresponding point about the later circuit constructions of \citet{liao2024graph}, which lack numerical benchmarks.
These costs shaped the later design space: compact, shared-parameter circuits in the equivariant line (\cref{sec:tax-eqgc}), and sampled or ego-graph aggregators that decouple circuit size from graph size (\cref{sec:tax-mp}).

% ---------------------------------------------------------------------------
\subsubsection{Graph-Structured Quantum Data}\label{sec:early-qdata}
% ---------------------------------------------------------------------------

A third early route reverses the roles.
\citet{beer2023quantum} consider data in which a quantum state is associated with each vertex of a graph and the edges encode correlations between the states; the training data are pairs of input and output quantum states on these vertices.
They train a quantum neural network on such data and show how to exploit the graph structure to improve learning with an arbitrary network ansatz, including a self-supervised objective that captures the information-theoretic closeness of quantum states, and show that the procedure can also infer unobserved vertex labellings.
The motivation is the characterization of noisy quantum devices and quantum data sources, for which full tomography scales exponentially with the number of qubits.

This setting lies at the boundary of our scope.
In \cref{def:qgnn} the input is a classical attributed graph $(\G,\X)$ that is loaded by $S(\G,\X)$; here the input is already a quantum state on $\bigotimes_v\Hil_v$, and the encoder $S$ is replaced by a given state preparation.
The graph acts as a constraint on the learning procedure and on which correlations to expect, not as a data structure to be loaded.
We therefore treat quantum-data models as a related line and focus on classical-graph inputs from \cref{sec:taxonomy} onwards, returning to quantum inputs where the symmetry arguments of \cref{sec:symmetry} apply.

% ---------------------------------------------------------------------------
\subsubsection{Synthesis}\label{sec:early-synthesis}
% ---------------------------------------------------------------------------

\begin{table}[t]
\centering
\caption{Early graph-structured quantum learning compared with the Hamiltonian-based QGNN, using the regimes of \cref{def:qgnn}. ``Quantum data'' marks inputs that are quantum states rather than classical graphs.}
\label{tab:early}
\renewcommand{\arraystretch}{1.3}
\footnotesize
\begin{tabular}{@{}p{3.1cm}p{2.5cm}p{4.3cm}p{2.7cm}p{2.0cm}@{}}
\toprule
\textbf{Work} & \textbf{Graph data} & \textbf{Quantum ingredient} & \textbf{Execution} & \textbf{Regime} \\
\midrule
\citet{bai2023learning} & classical graphs & quantum information propagation on aligned vertex grids & classical hardware & quantum-inspired \\
\citet{dernbach2019quantum} & classical graphs & learned, feature-dependent quantum-walk coins & classical hardware & quantum-inspired \\
\citet{verdon2019quantum} & quantum network / classical graphs & Hamiltonian evolution with graph interaction topology (\cref{sec:birth-hamiltonian}) & numerical simulation & fully quantum \\
\citet{zheng2021quantum} & classical graphs & amplitude encoding, graph-structured gates, measurement pooling & circuit simulation & fully quantum \\
\citet{zheng2024quantum} & classical graphs & circuit realization of the QSGCNN & circuit simulation & fully quantum \\
\citet{beer2023quantum} & quantum data on vertices & QNN with graph-informed learning objective & simulation & quantum data \\
\bottomrule
\end{tabular}
\end{table}

Three lessons from this period shape the remaining sections.
First, \emph{where the graph enters} was settled early and remains the main design axis: through the pattern of entangling gates or Hamiltonian couplings (\citet{verdon2019quantum}), through amplitudes of an encoded state (\citet{zheng2021quantum}), through classical pre- and post-processing around a quantum-inspired core (\citet{bai2023learning}), or through the correlations of the data itself (\citet{beer2023quantum}).
Second, the graph-specific circuit and the amplitude encoding are the two recurring \emph{scalability} bottlenecks, which anticipates the discussion of \cref{sec:scalability,sec:ch-scale}.
Third, almost all evidence in these works is numerical, on small problems and in the absence of standard benchmarks, which anticipates \cref{sec:ch-fair}.
What was still missing was a general criterion for graph-structured circuits to respect the symmetry of the domain; it was supplied by the equivariant quantum graph circuits of \citet{mernyei2022equivariant}, with which \cref{sec:taxonomy} begins, and which \cref{sec:notation-sym} has already formalized.

% ===========================================================================
\section{A Taxonomy of QGNN Architectures}\label{sec:taxonomy}
\begin{figure}[t]
\centering
\begin{tikzpicture}[every node/.style={font=\footnotesize},
  box/.style={draw,rounded corners=2pt,fill=blue!8,minimum height=6.5mm,text width=4.3cm,align=left,inner sep=3pt},
  root/.style={draw,rounded corners=2pt,fill=gray!15,minimum height=8mm,align=center,inner sep=3pt}]
  \node[root] (r) at (0,0) {QGNN\\architectures};
  \node[box] (a) at (5.4,2.4) {\ref{sec:tax-conv}\; Convolutional: copy a GCN layer};
  \node[box] (b) at (5.4,1.2) {\ref{sec:tax-eqgc}\; EQGC: symmetry by construction};
  \node[box] (c) at (5.4,0) {\ref{sec:tax-mp}\; Message passing, ego-graph, recurrent};
  \node[box] (d) at (5.4,-1.2) {\ref{sec:tax-physics}\; Hamiltonian / physics-inspired};
  \node[box] (e) at (5.4,-2.4) {\ref{sec:tax-hybrid}\; Hybrid quantum--classical};
  \foreach \n in {a,b,c,d,e}{\draw (r.east) -- ++(0.7,0) |- (\n.west);}
\end{tikzpicture}
\caption{Taxonomy of QGNN architectures used in this survey, with the subsection that treats each family.}
\label{fig:taxonomy}
\end{figure}
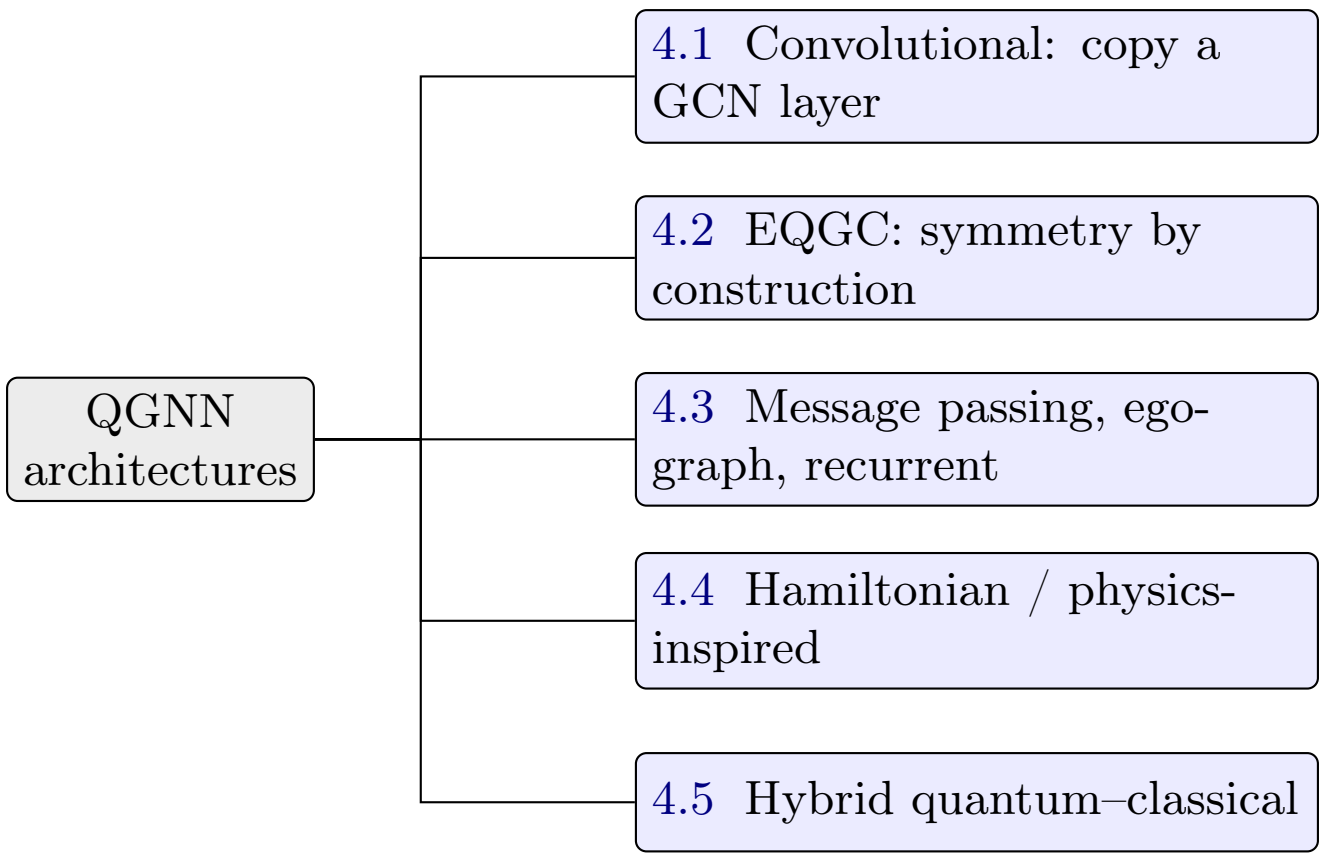

% ===========================================================================
Building on the unified definition of \cref{def:qgnn}, we organize QGNN architectures by \emph{what the layer is} and \emph{which part of the computation is quantum}, since these two choices determine qubit cost, symmetry, and scalability (\cref{fig:taxonomy}).
The families are not mutually exclusive: a model is assigned to the family that best describes how it processes the graph, and cross-memberships are noted in the text.
The subsections proceed from models that copy a classical layer (\cref{sec:tax-conv}), through models defined by a symmetry principle (\cref{sec:tax-eqgc}), by the form of the update (\cref{sec:tax-mp}), and by a Hamiltonian (\cref{sec:tax-physics}), to designs that split the work between classical and quantum parts (\cref{sec:tax-hybrid}); \cref{sec:tax-compare} compares them.
Quantum-inspired models and models whose inputs are quantum states (\cref{sec:birth-early}) are outside the taxonomy and appear only as context.

\subsection{Convolutional QGNNs}\label{sec:tax-conv}
% ===========================================================================
% Section 4.1  Convolutional QGNNs
% (included from main.tex via \input; heading and \label{sec:tax-conv}
%  live in main.tex)
% ===========================================================================

Convolutional QGNNs transfer the layer of a graph convolutional network (\cref{sec:mp-gcn}) to the quantum setting: node features are encoded in a state, a graph-dependent operation mixes them, and a measurement or a further quantum layer reduces the result.
They are the most direct reading of \cref{def:qgnn}, and the family splits by how the mixing operation is realized.
\emph{Variational} designs build it from parametrized gates arranged along the graph and target near-term devices (\cref{sec:conv-var}); \emph{algorithmic} designs implement the matrix operations of a GCN (block encodings, phase estimation) with complexity guarantees that rest on data-access assumptions (\cref{sec:conv-alg}).
This distinction, rather than the common name, determines what each family can claim.

% ---------------------------------------------------------------------------
\subsubsection{Variational Spatial Designs}\label{sec:conv-var}
% ---------------------------------------------------------------------------

\paragraph{Models.}
\citet{zheng2021quantum} encode graph data by amplitude encoding, build the convolution layer from parametrized universal gates arranged according to the graph, and pool by measurement; this is the template of the family (\cref{sec:early-circuits}).
\citet{hu2022design} instead use Givens rotations to encode graph information and a variational circuit for the learnable part, giving a fully quantum design for semi-supervised node classification, evaluated on a Cora subset; their analysis points to a potential advantage as the number of features grows.
\citet{zheng2024quantum} realize the quantum-inspired QSGCNN of \citet{bai2023learning} on circuits and study its trainability through barren plateaus (\cref{sec:early-inspired}).
\citet{zeng2025spatial} separate the circuit into quantum \emph{aggregation} circuits (the graph-convolutional layer) and quantum \emph{classification} circuits (a dense layer), report node-classification results comparable to or better than classical GNNs with fewer parameters, and, notably, run the circuits on a superconducting quantum computer.

\paragraph{Cost and symmetry.}
Amplitude-encoded designs need $\mathcal{O}(\log nF)$ qubits but pay for state preparation (\cref{tab:encodings}), and the circuit is built for a given graph, so the gate pattern must be rebuilt for each new graph.
Permutation equivariance (\cref{def:qequiv}) is in general not imposed as an explicit design criterion in this family; it holds only when parameters are shared across nodes and the encoding does not depend on node order, which should be checked per model.
Evidence is almost entirely numerical on small graphs, with the hardware run of \citet{zeng2025spatial} the exception among the works above.

% ---------------------------------------------------------------------------
\subsubsection{Algorithmic and Spectral Designs}\label{sec:conv-alg}
% ---------------------------------------------------------------------------

A second line implements the linear algebra of graph convolutions directly.
\citet{liao2024graph} give quantum circuits for the three classical layer types (convolution, attention, and message passing) based on block encodings, and a complexity analysis of a quantum simplified graph convolution that can reach logarithmic depth in the input sizes at the cost of linear space, a trade-off between depth and qubits that can be tuned to the hardware.
In the spectral setting, \citet{ye2025spectral} construct the graph Laplacian quantumly, extract its eigenvectors by phase estimation, and compute the convolution between eigenvectors and node features with the swap test; their analysis claims an exponential speedup in the number of nodes.
The variational spectral model of \citet{verdon2019quantum} (\cref{sec:ham-variants}) addresses the same idea without phase estimation.

These results are of a different kind from the variational ones: they are statements about asymptotic cost, they assume efficient access to classical data (quantum random access memory or efficient state preparation, \cref{sec:vqc}) and, for phase estimation, fault-tolerant hardware.
\citet{ye2025spectral} report no numerical evaluation, and the generic caveats on data loading and on dequantization of such speedups (\cref{sec:trainability-bg}) apply; we return to them in \cref{sec:scalability}.

\paragraph{Summary.}
Convolutional QGNNs are the oldest and most numerous family, but the label hides two regimes: NISQ-oriented variational circuits, whose claims are empirical, and fault-tolerant algorithms, whose claims are asymptotic and conditional on data access.
Both inherit the cost of encoding the graph into a quantum state, which the equivariant and sampled designs of \cref{sec:tax-eqgc,sec:tax-mp} try to avoid.

\subsection{Equivariant Quantum Graph Circuits (EQGC)}\label{sec:tax-eqgc}
% ===========================================================================
% Section 4.2  Equivariant Quantum Graph Circuits (EQGC)
% (included from main.tex via \input; heading and \label{sec:tax-eqgc}
%  live in main.tex)
% ===========================================================================

Equivariant quantum graph circuits make the symmetry condition of \cref{def:qequiv} the organizing principle of the architecture.
Instead of copying a classical layer, \citet{mernyei2022equivariant} ask which parametrized circuits respect the invariances of the graph domain and characterize the resulting class; its extended journal version \citep{mernyei2023constructions} adds constructions for universal approximation.
The proposal also serves as a unifying framework: several earlier models, including the Hamiltonian-based family of \cref{sec:birth-hamiltonian}, become special cases.

\paragraph{The class and its subclasses.}
An EQGC is a parametrized circuit on the node--qubit register whose action commutes with the relabeling of nodes in the sense of \cref{eq:circ-equiv}.
Two practical subclasses are defined through one-node and two-node operators.
\emph{Equivariant Hamiltonian QGCs} (EH-QGCs) evolve under Hamiltonians that are sums of one-node and two-node terms, which is the form of \cref{eq:verdon-ansatz} with tied coefficients (\cref{prop:ham-equiv}).
\emph{Equivariantly diagonalizable unitary QGCs} (EDU-QGCs) alternate node layers $V^{\otimes n}$, the same unitary on every node, with edge layers that apply to every edge a unitary diagonalizable in a node-local basis; as we read the construction, the factors of such a layer commute, so their order does not matter, which avoids the ordering issue noted after \cref{prop:qgnn-equiv}.
Every EDU-QGC can be written as a sequence of EH-QGC layers, so the EDU class is the more restricted of the two~\citep{mernyei2022equivariant}.

\paragraph{Expressivity.}
The characteristic result is theoretical.
EH-QGCs and EDU-QGCs are universal approximators of functions on graphs of bounded size, shown via a correspondence between EDU-QGCs and message-passing networks enhanced with random node initialization; because quantum circuits are probabilistic, the circuit size needed depends on a confidence parameter~\citep{mernyei2022equivariant,mernyei2023constructions}.
Empirically, EDU-QGCs with one qubit per node, trained in classical simulation on a toy task (distinguishing graphs consisting of one cycle from graphs consisting of two cycles, 6 to 10 nodes), separate graph pairs that standard message-passing networks cannot, which demonstrates expressivity beyond 1-WL on small instances~\citep{mernyei2022equivariant}.
\citet{skolik2023equivariant} apply the same symmetry idea to weighted graphs with QAOA-type circuits.

\paragraph{Cost and limits.}
The node--qubit mapping gives $n_q=n$ and one two-qubit interaction per edge per layer, so width and depth grow with the graph; this is the price of exact equivariance and it limits experiments to small graphs, as \citet{faria2025inductive} observe for the earlier evaluations on small synthetic data.
Universality is an existence statement that does not bound the size of the circuit needed, and the guarantee is probabilistic.
More recent work asks how to keep the symmetry while scaling: in-circuit message passing with an equivariant adjacency layer, positioned at a chosen level of the Weisfeiler--Leman hierarchy above 1-WL and simulated up to 56 qubits, is discussed in \cref{sec:mp-incircuit}.

\paragraph{Summary.}
EQGCs are the only family in this section whose defining property is a symmetry guarantee proved at the circuit level.
They set the reference against which the symmetry of the other families is judged (\cref{tab:tax-compare}) and connect directly to the trainability arguments of \cref{sec:trainability-bg} and to the group-theoretic treatment of \cref{sec:symmetry}.

\subsection{Message-Passing, Ego-Graph, and Recurrent QGNNs}\label{sec:tax-mp}
% ===========================================================================
% Section 4.3  Message-Passing / Ego-Graph and Recurrent QGNNs
% (included from main.tex via \input; heading and \label{sec:tax-mp}
%  live in main.tex)
% ===========================================================================

The designs of this subsection keep the shape of a classical message-passing network (\cref{eq:mpnn}) and ask a different question from the convolutional family: which part of the update runs in a quantum circuit, and how large must that circuit be?
Three answers have been explored.
The graph can be cut into small local pieces so that a modest circuit serves graphs of any size (\cref{sec:mp-ego}); message passing can be carried out \emph{inside} the circuit, with guarantees (\cref{sec:mp-incircuit}); or the update can be iterated with shared parameters, as in recurrent models (\cref{sec:mp-recurrent}).

% ---------------------------------------------------------------------------
\subsubsection{Ego-Graph and Sampled Aggregators}\label{sec:mp-ego}
% ---------------------------------------------------------------------------

The ego-graph QGNN of \citet{ai2024ego} is a hybrid method that processes the ego-graphs of the input graph on a modest quantum device, controlled by a classical computer, so that graphs of arbitrary size can be handled.
It represents the GNN computation by tensor products and unity matrices, which greatly reduces the number of parameters, relies on a mapping from data to Hilbert space that maintains distance relations, and is reported to outperform competing models with a small fraction (1.68\%) of their parameters.
\citet{faria2025inductive} follow the same decoupling with a GraphSAGE-style design: neighbours are sampled, a quantum circuit built from convolution and pooling layers acts as the aggregator, and the number of qubits depends only on the node-feature dimension, not on the sampled nodes or edges.
On a QM9 node-regression task their single shared circuit trains worse than a parameter-matched classical GNN but generalizes better as molecules grow, using distinct circuits per hop improves training, and numerical gradient-variance scaling shows no barren plateau as qubits increase; the authors state that the contribution lies in generalization and not in a runtime advantage.
The same decomposition idea has been applied to wireless networks by \citet{giang2026scalable}, who process each local subgraph with a shared-parameter circuit that includes node and edge attributes.

\paragraph{Cost and symmetry.}
Circuit width is set by the feature dimension, so it is independent of $n$; the total number of circuit executions still grows with the number of nodes or ego-graphs, like the cost of a classical GNN.
Relabeling symmetry follows from sharing the aggregator across nodes and, in \citet{faria2025inductive}, from assigning aggregators by hop depth rather than by node identity.
Because the graph is handled classically (sampling, ego-extraction), global entanglement across the graph is given up, which is the reason these models scale.

% ---------------------------------------------------------------------------
\subsubsection{In-Circuit Message Passing}\label{sec:mp-incircuit}
% ---------------------------------------------------------------------------

Most circuit-based models derive the gate layout from the graph but leave the aggregation of neighbour information to a classical readout; a recent preprint targets exactly this gap~\citep{raj2026scalable}.
It builds a QGNN from four blocks (a data loader, an equivariant adjacency layer, a trainable evolution, and a mixing layer between registers), re-uploads the graph at every layer to introduce nonlinearity, and shows that the model is permutation equivariant and sits at a level of the Weisfeiler--Leman hierarchy set by the model, above the 1-WL limit of ordinary message passing.
The authors validate it in simulation up to 56 qubits on Cai--F\"urer--Immerman graphs, QM9 property prediction, and the Euclidean travelling-salesperson problem, and argue that the trainable dynamics do not scale with graph size, which avoids cost concentration and allows pre-training on small graphs.
Earlier, \citet{liao2024graph} had given circuits for quantum attention and message-passing layers, with the caveats on data access of \cref{sec:conv-alg}.
Compared with the EQGCs of \cref{sec:tax-eqgc}, which obtain expressivity through random node initialization, this line ties expressivity to an explicit position in the Weisfeiler--Leman hierarchy; as a 2026 preprint, its claims await independent replication.

% ---------------------------------------------------------------------------
\subsubsection{Recurrent QGNNs}\label{sec:mp-recurrent}
% ---------------------------------------------------------------------------

Recurrent QGNNs iterate one graph-structured update with shared parameters; the QGRNN of \citet{verdon2019quantum} (\cref{sec:ham-variants}) is the prototype, and its parameter sharing over repetitions yields depth without growth in the number of parameters.
\citet{choi2021tutorial} present the QGRNN from a software point of view, with the Ising model, the variational quantum eigensolver for preparing quantum data, and the training of an Ising Hamiltonian.
Recent work moves the model from quantum data to classical graph data: \citet{kaldari2025feature} use a QGRNN to reconstruct node features of classical datasets and to hide messages inside a graph, and \citet{kaldari2025resqgrnn} add residual connections to improve the trainability of QGRNNs while respecting the no-cloning constraint on quantum states.
The remaining open point is evidence: these recurrent models are demonstrated on small instances in simulation, and their trainability over many repetitions is not yet characterized (\cref{sec:trainability-bg}).

\paragraph{Summary.}
This family is where QGNNs most resemble their classical counterparts in structure, and where the scaling question is addressed head on: size-independent circuits trade global entanglement for locality, while in-circuit message passing aims to keep the quantum character with provable symmetry and expressivity.
Both options appear in the comparison of \cref{sec:tax-compare}.

\subsection{Hamiltonian and Physics-Inspired QGNNs}\label{sec:tax-physics}
% ===========================================================================
% Section 4.4  Hamiltonian / Physics-Inspired QGNNs
% (included from main.tex via \input; heading and \label{sec:tax-physics}
%  live in main.tex)
% ===========================================================================

The foundational Hamiltonian model and its specializations were presented in \cref{sec:birth-hamiltonian}; here we place the family in the taxonomy and review how it has been used since.
A QGNN belongs to this category when its layer is \emph{time evolution under a graph-dependent Hamiltonian}, either executed natively as analog dynamics or compiled into gates (\cref{eq:hamevo}), so that the graph acts through the couplings of the Hamiltonian and parameters are evolution times and coefficients.
Two senses of ``physics'' should be kept apart.
\emph{Physics-inspired} models use a Hamiltonian as an architectural device for graph data; \emph{physics-targeting} models apply quantum graph circuits to physical systems.

\paragraph{Hamiltonian architectures.}
The quantum graph Hamiltonian neural network (QGHNN) of \citet{wang2025qghnn} treats the adjacency matrix as the coupling coefficients of a Pauli Hamiltonian and builds a layered circuit from the decomposition of the corresponding time-evolution operator into rotation and CNOT gates, trained on the expectation of the graph-derived Hamiltonian; it is motivated as graph learning through Hamiltonian learning on near-term devices.
At the level of the taxonomy, such models are instances of \cref{eq:verdon-ansatz} and, when their coefficients are tied, of the EH-QGC subclass of \cref{sec:tax-eqgc}.

\paragraph{Physics simulation.}
\citet{collis2023physics} develop two quantum realizations of the interaction network used in learned physics simulators and test them on the simulation of particles interacting under gravity in a box; the first is a speculative design and the second an implementable hybrid model that propagates particle information through the parameters of rotation gates, both compared with a classical GNN.
This is the physics-targeting use: the graph is a system of interacting bodies and the quantum circuit replaces the interaction network of an encoder--processor--decoder pipeline.

\paragraph{Strengths and limits.}
The family has three structural advantages.
The graph enters without data loading, since the topology fixes the interaction pattern.
Parameter tying yields a size-independent parameter count and, by \cref{prop:ham-equiv}, exact symmetry.
Analog implementations can bypass compilation entirely on hardware whose interactions can be programmed to match the graph.
Its limits mirror these advantages: the circuit is tied to one graph, a native analog implementation requires matching connectivity, a gate-based implementation inherits Trotter error and depth (\cref{sec:ham-ansatz}), and the symmetry restriction that makes training feasible is the same one that raises the question of classical simulability (\cref{sec:ham-legacy}).
In terms of cost, this family sits with the node--qubit models: $n_q=n$ and a gate count growing with $n+m$ per step.

\paragraph{Summary.}
The Hamiltonian family offers the cleanest link between graphs and quantum hardware, but evidence of benefit on graph-learning benchmarks is thin, and its natural domain remains systems in which the graph is a physical interaction pattern.
Its role in hybrid pipelines, where the Hamiltonian only generates features, is covered next.

\subsection{Hybrid Quantum--Classical Designs}\label{sec:tax-hybrid}
% ===========================================================================
% Section 4.5  Hybrid Quantum--Classical Designs
% (included from main.tex via \input; heading and \label{sec:tax-hybrid}
%  live in main.tex)
% ===========================================================================

Hybrid designs keep a classical backbone for graph processing and insert a small quantum component, which makes them the dominant choice for applications today.
In \cref{def:qgnn} they correspond to a non-trivial classical map $g_{\boldsymbol\phi}$ and/or classical features entering $S(\G,\X)$.
We distinguish three patterns by the position of the quantum component.

\paragraph{Quantum head on a classical GNN.}
A classical GNN produces a graph or node embedding that a variational quantum classifier consumes.
\citet{ray2023hybrid} combine a GNN with a variational quantum classifier for breast-cancer subtyping on tumour-microenvironment graphs and study two questions: whether the quantum model helps in higher feature dimensions, and which compression scheme is better when qubits are scarce.
They find that amplitude encoding, which stores the embedding in a logarithmic number of qubits, outperforms classical compression, that end-to-end training of both parts improves over a fixed GNN, and that the hybrid model is on par with state-of-the-art classical GNNs on weighted precision, recall, and F1.

\paragraph{Quantum layer inside the message-passing pipeline.}
Here a quantum circuit replaces a classical sub-network such as the MLP of the update or the readout.
\citet{mauro2024hybrid} replace the dense part of a graph convolutional network with a quantum multilayer perceptron using amplitude encoding for Oceanic Ni\~no Index prediction; the motivation is that quantum convolutional networks capture spatial features well but are limited for global dependencies such as El Ni\~no fluctuations, which the graph-based part is meant to supply.
The NISQ-oriented circuit design of \citet{hu2022design}, built from Givens rotations to encode graph information and a variational circuit for learning, serves as a reference for how few qubits and which gates such a component needs (\cref{sec:conv-var}); it is itself a fully quantum design, so it marks the boundary between this pattern and \cref{sec:tax-conv}.
The sampled aggregators of \cref{sec:mp-ego} are the most developed instance, as the quantum circuit is the aggregation function of a GraphSAGE-type network.

\paragraph{Quantum-computed features for a classical model.}
In the third pattern the quantum computer is used only to compute structural features that a classical network consumes.
\citet{thabet2024quantum} map graph topology to qubit interactions and take positional encodings from the long-range correlations of the resulting quantum states; they prove that some of these features are more expressive than relative random-walk probabilities for certain graphs, and show that tractable (classically computable) versions improve state-of-the-art models on standard benchmarks.
The quantum part is not trained, so the barren-plateau obstacle (\cref{sec:trainability-bg}) does not arise, and the interface to the classical model is a feature matrix.

\paragraph{Cost, symmetry, and evaluation.}
The quantum component acts on a compressed or low-dimensional input, so qubit count is set by the embedding or feature dimension and is independent of the graph size; scalability is inherited from the classical backbone.
Permutation equivariance is also inherited, provided the quantum head is applied after a symmetric readout or node-wise with shared parameters.
The main methodological risk is attribution: with a classical network upstream, a gain from the quantum part must be separated from the effect of the encoding or of added capacity, which requires ablations such as classical feature maps and parameter-matched classical baselines (\cref{sec:ch-fair}); \citet{faria2025inductive} include a control of this kind.

\paragraph{Summary.}
Hybrid designs trade the possibility of quantum advantage for feasibility: small circuits, a mature classical backbone, and proven application domains (medicine, earth observation, physics), but no clear mechanism by which the quantum component should help and, so far, results that are mostly on par with classical models rather than beyond them.

\subsection{Comparison of Architectures}\label{sec:tax-compare}
% ===========================================================================
% Section 4.6  Comparison of Architectures
% (included from main.tex via \input; heading and \label{sec:tax-compare}
%  live in main.tex)
% ===========================================================================

\Cref{tab:tax-compare} compares the families of \cref{sec:tax-conv,sec:tax-eqgc,sec:tax-mp,sec:tax-physics,sec:tax-hybrid} along the four axes that matter for the questions of this survey: the number of qubits $n_q$ as a function of the input, whether permutation equivariance (\cref{def:qequiv}) is guaranteed by construction, what limits scalability, and what kind of evidence has been reported.
The qubit counts are class-level scalings implied by the encoding (\cref{tab:encodings}) and the node--qubit mapping, not figures measured in individual papers, and the equivariance column states what the cited works establish or state, not what a variant might achieve.

\begin{table}[t]
\centering
\caption{Comparison of QGNN architecture families. $n$: nodes, $m$: edges, $F$: features per node. ``By construction'' means proven at the circuit level (\cref{def:qequiv}); ``not imposed'' means not stated as a design criterion in the cited works. Evidence is as reported by the authors.}
\label{tab:tax-compare}
\renewcommand{\arraystretch}{1.3}
\footnotesize
\begin{tabular}{@{}p{2.2cm}p{2.4cm}p{2.2cm}p{2.0cm}p{2.4cm}p{2.3cm}@{}}
\toprule
\textbf{Family} & \textbf{Examples} & \textbf{Qubits $n_q$} & \textbf{Equivariance} & \textbf{Scalability limit} & \textbf{Evidence} \\
\midrule
Convolutional, variational & \citet{zheng2021quantum,hu2022design,zeng2025spatial} & $\mathcal{O}(\log nF)$ (amplitude) & not imposed & state preparation; circuit per graph & simulation; one hardware run~\citep{zeng2025spatial} \\
Convolutional, algorithmic/spectral & \citet{liao2024graph,ye2025spectral} & $\mathcal{O}(\log n)$ plus ancillas & not applicable (algorithmic) & data access (QRAM); fault tolerance & complexity analysis \\
EQGC (EH/EDU) & \citet{mernyei2022equivariant,mernyei2023constructions} & $\mathcal{O}(n)$ & by construction & width $n$, $m$ edge gates per layer & universality proof; small simulations \\
Ego-graph / sampled aggregators & \citet{ai2024ego,faria2025inductive} & $\mathcal{O}(F)$, independent of $n$ & shared aggregator; stated in \citep{faria2025inductive} & circuit runs grow with nodes; local view only & simulation (QM9 and others) \\
In-circuit message passing & \citet{raj2026scalable} & node register plus embedding register; simulated to 56 qubits & by construction (stated) & simulation cost; preprint & proofs; simulation \\
Recurrent / Hamiltonian & \citet{verdon2019quantum,wang2025qghnn} & $\mathcal{O}(n)$ & by construction if tied (\cref{prop:ham-equiv}) & graph-specific circuit; Trotter depth & small simulations \\
Hybrid (head, layer, features) & \citet{ray2023hybrid,mauro2024hybrid,thabet2024quantum} & set by feature or embedding dimension & inherited from the classical part & classical backbone & applications; parity with classical models \\
\bottomrule
\end{tabular}
\end{table}

Four observations follow from the table and the preceding subsections.

\paragraph{Symmetry and qubit cost are coupled.}
Guaranteed equivariance appears together with the node--qubit mapping ($n_q=\mathcal{O}(n)$) and its width, while the families with size-independent circuits (sampled aggregators, hybrid heads) obtain scalability by treating the graph classically and giving up global entanglement across it.
Compressed encodings reduce qubit count but are not equivariant by default, since permutations then act on basis-state indices and must be handled explicitly (\cref{sec:notation-sym}).

\paragraph{Claims differ in kind.}
Algorithmic designs claim asymptotic speedups under data-access assumptions; variational designs claim empirical parity or better generalization at small scale; the EQGC and in-circuit message-passing lines claim expressivity results.
These are not comparable on one axis, which is why \cref{sec:ch-fair} treats evaluation protocols separately.

\paragraph{Hardware evidence is scarce.}
Among the works reviewed in this section, a single hardware implementation is reported~\citep{zeng2025spatial}; most results come from noiseless simulation, in line with the NISQ limitations of \cref{sec:trainability-bg}.

\paragraph{Recent work targets the gaps.}
The 2025--2026 contributions in the table concentrate on the weaknesses of earlier families: hardware execution~\citep{zeng2025spatial}, size-independent circuits and generalization~\citep{faria2025inductive}, and symmetry together with expressivity beyond 1-WL at larger scale~\citep{raj2026scalable}.
\Cref{sec:symmetry,sec:expr-train-scale} examine symmetry, expressivity, and scaling in depth, and \cref{sec:ch-fair} returns to evaluation.

% ===========================================================================
\section{Symmetry, Equivariance, and Geometric Inductive Biases}\label{sec:symmetry}
% ===========================================================================
Symmetry is the most developed inductive bias of QGNNs: it fixes what a valid model is, supports trainability arguments, and drives the application-specific designs of high-energy physics and 3D molecular learning.
We treat relabeling symmetry (\cref{sec:sym-perm}), rotations in 3D (\cref{sec:sym-rot}), and the Lorentz group (\cref{sec:sym-lorentz}), and compare the design strategies in \cref{sec:sym-compare}.

\subsection{Permutation Equivariance and Invariance}\label{sec:sym-perm}
% ===========================================================================
% Section 5.1  Permutation Equivariance and Invariance
% (included from main.tex via \input; heading and \label{sec:sym-perm}
%  live in main.tex)
% ===========================================================================

Permutation symmetry is the baseline symmetry of graph learning, and \cref{def:qequiv,prop:qgnn-equiv} state when a QGNN respects it.
All symmetries in this section follow the same pattern.
For a group $\mathsf{G}$ acting on inputs by $g\cdot x$ and on the register through a unitary representation $R_g$, a model is $\mathsf{G}$-equivariant if
\begin{equation}\label{eq:g-equiv}
  f_{\boldsymbol\theta}(g\cdot x)=\rho(g)\,f_{\boldsymbol\theta}(x)\qquad\forall g\in\mathsf{G},
\end{equation}
with $\rho$ the representation on the output ($\rho=\mathrm{id}$ for invariants), and a sufficient condition is \cref{eq:circ-equiv} with $R_\pi$ replaced by $R_g$ together with observables that are closed under conjugation by $R_g$.
For $\mathsf{G}=\Sym$, $R_\pi$ permutes qubits.

\paragraph{Three ways to obtain it.}
(i) \emph{Shared-parameter layers}: tied node and edge terms, as in the Hamiltonian family (\cref{prop:ham-equiv}) and the EQGCs of \cref{sec:tax-eqgc}, are equivariant by construction.
(ii) \emph{Symmetrization}: a non-equivariant circuit is averaged over the group; the permutation-equivariant quantum convolutional network of \citet{das2024permutation} applies all possible QCNNs with equal probability, which can be read as a dropout-like strategy, at the price of many circuit evaluations.
(iii) \emph{Representation-theoretic restriction}: gates are chosen from the commutant of the group action, the route of geometric quantum machine learning~\citep{larocca2022group,meyer2023exploiting,nguyen2024theory}.
Permutation-equivariant circuits have also been built for weighted graphs~\citep{skolik2023equivariant} and compared with rotation-equivariant and classical networks in molecular learning~\citep{biswas2025permeq}.

\paragraph{Why it matters beyond correctness.}
Equivariant circuits explore a much smaller subspace of $\Hil$ and, under suitable conditions, have gradient variance that decays only polynomially~\citep{schatzki2024theoretical}, which is the main trainability argument for symmetric QGNNs (\cref{sec:trainability-bg}).
The same restriction is what invites classical-simulation arguments~\citep{cerezo2025does}, so symmetry is both a route to trainability and a reason to test claims of advantage carefully (\cref{sec:ch-theory}).
In practice, exact equivariance is a statement about noiseless expectation values; finite shots and hardware noise break it sample by sample (\cref{sec:notation-sym}).

\subsection{Rotation and 3D Equivariance}\label{sec:sym-rot}
% ===========================================================================
% Section 5.2  Rotation and 3D Equivariance
% (included from main.tex via \input; heading and \label{sec:sym-rot}
%  live in main.tex)
% ===========================================================================

Graphs with coordinates, such as molecules and point clouds, carry geometric symmetry in addition to relabeling symmetry: rotating the input should rotate vector outputs and leave scalar ones unchanged.
Classical E($n$)-equivariant GNNs satisfy translation, rotation, reflection, and permutation constraints together~\citep{satorras2021n}; the quantum counterparts are recent.

\paragraph{Models.}
\citet{liu2025rotation} observe that earlier equivariant QGNNs consider only permutation symmetry and therefore underuse geometric information on 3D data.
They derive the constraints that a quantum graph network must satisfy to be rotation- and permutation-equivariant and propose the RP-EQGNN, evaluated on molecular property prediction and on point-cloud classification, where it is reported to improve over quantum baselines on ModelNet40.
A follow-up \citep{liu2026rotation} adds a trainable compression encoder and entanglement-enhanced aggregation to a rotation-equivariant QGNN; as the title indicates, the encoder addresses the qubit cost of encoding features and the aggregation uses entanglement to combine neighbours.
\citet{park2026hyqurp} take a hybrid route, HyQuRP, with four quantum stages and a classical head, each designed to preserve rotation and permutation, and report that it generally surpasses quantum 3D point-cloud baselines, including the RP-EQGNN, in a controlled sparse-point regime.

\paragraph{Design choice.}
Two strategies recur: encode rotation-invariant scalars (distances, angles) and keep the circuit unconstrained, or constrain the gates so that vector features transform correctly.
The first is cheap and exact but discards directional information; the second needs group-aware encodings and costs qubits, which is why compression encoders appear.
\Cref{sec:sym-compare} compares these strategies.

\subsection{Lorentz Equivariance in High-Energy Physics}\label{sec:sym-lorentz}
% ===========================================================================
% Section 5.3  Lorentz Equivariance in High-Energy Physics
% (included from main.tex via \input; heading and \label{sec:sym-lorentz}
%  live in main.tex)
% ===========================================================================

Collision events at the Large Hadron Collider are naturally represented as graphs, with jets viewed as particle clouds, and the relevant symmetry is the Lorentz group.
Classical networks such as LorentzNet impose it layer by layer.
The Lorentz-equivariant QGNN of \citet{jahin2025lorentz} replaces the multilayer-perceptron components of LorentzNet's equivariant block with dressed quantum circuits and obtains Lorentz symmetry by scalarizing the particle four-vectors with Minkowski dot products before they enter the circuits.
On quark--gluon jet tagging the authors report 74.00\% test accuracy and an AUC of 87.38\%, above the classical and quantum GNN baselines they compare with, using about 5.5 times fewer parameters and only four qubits; on a data-scarce electron--photon task they report 67.00\% accuracy with 800 training samples.

Two points should be kept in view.
As we read the construction, the quantum circuits act on Lorentz-invariant scalars, so the symmetry is carried by the classical equivariant structure and by the choice of inputs, not by the gates; it is an instance of the ``invariant inputs'' strategy of \cref{sec:sym-compare}.
Second, the reported comparisons are in simulation on limited data, so the parameter and qubit savings are evidence of efficiency in that regime, not of advantage.
HEP is, nonetheless, the application where symmetric QGNNs are most developed, since small, symmetry-constrained models suit the data-scarce settings typical of the field.

\subsection{Invariant versus Equivariant Designs Compared}\label{sec:sym-compare}
% ===========================================================================
% Section 5.4  Invariant versus Equivariant Designs Compared
% (included from main.tex via \input; heading and \label{sec:sym-compare}
%  live in main.tex)
% ===========================================================================

\citet{forestano2024comparison} give the most direct comparison for QGNNs: classical GNNs and equivariant GNNs against their quantum counterparts, with invariant inputs in the plain networks and equivariant layers in the equivariant ones, on the binary task of identifying the parton that initiated a jet, with identical training, validation, and test sets.
By AUC the quantum networks outperform the classical ones; the authors add that seeing a computational advantage in practice may have to wait for further hardware and software development.
\Cref{tab:sym-compare} summarizes the design patterns met in this section.

\begin{table}[t]
\centering
\caption{Strategies for building symmetry into QGNNs.}
\label{tab:sym-compare}
\renewcommand{\arraystretch}{1.3}
\footnotesize
\begin{tabular}{@{}p{2.6cm}p{3.5cm}p{3.6cm}p{3.6cm}@{}}
\toprule
\textbf{Strategy} & \textbf{Where symmetry lives} & \textbf{Examples} & \textbf{Cost / limit} \\
\midrule
Invariant inputs & features (distances, Minkowski products) & \citet{jahin2025lorentz}; invariant baselines of \citet{forestano2024comparison} & cheap; loses directional information \\
Equivariant gates & circuit (commutant of the group) & EQGC~\citep{mernyei2022equivariant}; \citet{liu2025rotation} & group-aware encoding; more qubits \\
Symmetrization & average over the group & \citet{das2024permutation} & many circuit evaluations \\
Classical equivariant backbone & classical layers; quantum part small & \citet{jahin2025lorentz}; \citet{park2026hyqurp} & symmetry of the quantum part unchecked \\
\bottomrule
\end{tabular}
\end{table}

No pattern dominates.
Equivariant gates give the strongest guarantee and the trainability argument of \cref{sec:sym-perm}, but at the qubit cost seen in \cref{tab:tax-compare}; invariant inputs and classical backbones are cheaper and are the choice of most application-driven work.
Reported gains over classical baselines come from small simulated settings, so the comparison must control for parameter count and training data before the benefit of the symmetry itself can be separated from that of the quantum model (\cref{sec:ch-fair}), and the interaction between symmetry, expressivity, and scaling is taken up in \cref{sec:expr-train-scale}.

% ===========================================================================
\section{Expressivity, Trainability, and Scalability}\label{sec:expr-train-scale}
% ===========================================================================
Three properties decide whether a QGNN is useful beyond toy graphs: what it can represent, whether it can be trained, and how large a graph it can handle.
They are coupled: restricting circuits helps trainability but may remove expressive power or invite classical simulation, compression helps scale but discards information, and noise limits depth for all of them.
This section collects the results for QGNN families; the general background is in \cref{sec:trainability-bg}, and open problems are in \cref{sec:challenges}.

\subsection{Universal Approximation Results}\label{sec:universal}
% ===========================================================================
% Section 6.1  Universal Approximation Results
% (included from main.tex via \input; heading and \label{sec:universal}
%  live in main.tex)
% ===========================================================================

The classical reference is the 1-WL limit of message passing, which holds for any permutation-equivariant MPNN (\cref{sec:equivariance}) and can be exceeded by adding randomness, subgraph information, or higher-order structure.
Results for QGNNs fall into four kinds (\cref{tab:expr}).

\paragraph{Universality.}
The equivariant circuits of \cref{sec:tax-eqgc} are universal approximators of functions on graphs of bounded size; the proof goes through a correspondence between EDU-QGCs and message-passing networks with random node initialization, and the circuit size needed depends on a confidence parameter~\citep{mernyei2022equivariant,mernyei2023constructions}.

\paragraph{Position in the WL hierarchy.}
The in-circuit message-passing QGNN of \citet{raj2026scalable} is permutation equivariant and sits at a level of the Weisfeiler--Leman hierarchy set by the model, above 1-WL, and is validated in simulation on graphs designed to defeat 1-WL.

\paragraph{Feature-level separation.}
\citet{thabet2024quantum} prove that quantum positional encodings are more expressive than relative random-walk probabilities for certain graphs, an argument about features supplied to a classical model, not about a trained quantum network (\cref{sec:tax-hybrid}).

\paragraph{Function-class view.}
Independently of graphs, a model with Pauli-rotation encoding is a truncated Fourier series in the data whose accessible frequencies are set by the encoding generators and the number of re-uploadings~\citep{schuld2021effect}; this transfers to QGNNs through the encoder $S(\G,\X)$ of \cref{def:qgnn}.

\begin{table}[t]
\centering
\caption{Expressivity results for QGNNs and what they do and do not establish.}
\label{tab:expr}
\renewcommand{\arraystretch}{1.3}
\footnotesize
\begin{tabular}{@{}p{2.9cm}p{3.7cm}p{3.1cm}p{3.5cm}@{}}
\toprule
\textbf{Result} & \textbf{Mechanism} & \textbf{Source} & \textbf{Scope and caveat} \\
\midrule
Universality on bounded graphs & equivalence to MPNN with random node initialization & \citet{mernyei2022equivariant} & existence statement; circuit size not bounded; probabilistic \\
Beyond 1-WL, equivariant & in-circuit message passing at a chosen WL level & \citet{raj2026scalable} & preprint; simulation up to 56 qubits \\
Quantum PE $>$ RRWP & correlations of graph-Hamiltonian dynamics & \citet{thabet2024quantum} & quantum part untrained; classical versions also help \\
Fourier-series model & encoding generators fix the frequencies & \citet{schuld2021effect} & function-class statement, not graph-specific \\
\bottomrule
\end{tabular}
\end{table}

\paragraph{Reading the results.}
Three cautions apply.
First, universality is an existence result with no bound on circuit size, and its mechanism, as the correspondence suggests, is the same randomness that lets classical networks with random node initialization separate graphs; it shows that QGNNs are not weaker than that class, not that they are stronger.
Second, no result yet identifies a graph-learning task that a QGNN solves efficiently and a classical model provably cannot; the dequantization argument for quantum graph convolutions~\citep{lo2026quantum} points the other way for low-rank settings (\cref{sec:trainability-bg}).
Third, expressivity trades against trainability: highly expressive ansatzes concentrate the cost~\citep{holmes2022connecting} and large dynamical Lie algebras give barren plateaus~\citep{ragone2024lie}, which is the subject of \cref{sec:bp} and of the open theoretical questions in \cref{sec:ch-theory}.

\subsection{Barren Plateaus and Initialization Strategies}\label{sec:bp}
% ===========================================================================
% Section 6.2  Barren Plateaus and Initialization Strategies
% (included from main.tex via \input; heading and \label{sec:bp}
%  live in main.tex)
% ===========================================================================

The general theory of barren plateaus, noise-induced plateaus, and the trainability--simulability tension is summarized in \cref{sec:trainability-bg}; here we collect what is known \emph{for QGNN families}.
Four design features matter.
Graph-level readouts such as $\tfrac1n\sum_vZ_v$ and node-level readouts $Z_v$ are local observables, which is favourable.
Parameter tying makes the parameter count independent of $n$, but plateaus are governed by the observables and the dynamical Lie algebra, not by the parameter count.
Symmetry restricts the algebra, and this is the main source of provable guarantees~\citep{schatzki2024theoretical}.
Finally, tying allows \emph{transfer}: parameters trained on small graphs can be reused on larger ones, which \citet{raj2026scalable} use for pre-training.

\begin{table}[t]
\centering
\caption{Trainability evidence for QGNN families.}
\label{tab:train}
\renewcommand{\arraystretch}{1.3}
\footnotesize
\begin{tabular}{@{}p{2.8cm}p{3.6cm}p{3.6cm}p{3.2cm}@{}}
\toprule
\textbf{Family} & \textbf{Mechanism} & \textbf{Evidence} & \textbf{Caveat} \\
\midrule
Equivariant / tied Hamiltonian & symmetry restricts the algebra & theory for permutation-equivariant circuits~\citep{schatzki2024theoretical} & restriction may allow classical simulation~\citep{cerezo2025does} \\
Sampled / ego-graph & circuit width independent of graph & numerical gradient variance without plateau as qubits grow~\citep{faria2025inductive} & small qubit counts; simulation \\
In-circuit message passing & trainable dynamics do not scale with graph size & Lie-algebra argument and simulation~\citep{raj2026scalable} & preprint \\
Variational convolutional & shallow, local circuits & plateau analysis of the QSGCN~\citep{zheng2024quantum}; trainability study with dequantization~\citep{lo2026quantum} & mixed conclusions \\
Recurrent & residual connections & empirical improvement~\citep{kaldari2025resqgrnn} & simulation \\
\bottomrule
\end{tabular}
\end{table}

\paragraph{Initialization.}
The generic remedies of \cref{sec:trainability-bg}, near-identity initialization~\citep{grant2019initialization} and layer-wise training~\citep{skolik2021layerwise}, apply directly.
Hamiltonian layers are near-identity for small evolution times, so identity-like initialization is natural there; a related remedy for recurrent models is the residual connection of \citet{kaldari2025resqgrnn}.
Beyond initialization, hybrid designs sidestep the problem by keeping the quantum part small (\cref{sec:tax-hybrid}).

\paragraph{Takeaway.}
Positive trainability results in \cref{tab:train} come largely from restricting the circuit, which is also what raises the simulability question.
No study yet shows a QGNN that is provably trainable at graph scale and not classically simulable; this is carried into \cref{sec:ch-train,sec:ch-theory}.

\subsection{Noise, Decoherence, and Error Mitigation}\label{sec:noise}
% ===========================================================================
% Section 6.3  Noise, Decoherence, and Error Mitigation
% (included from main.tex via \input; heading and \label{sec:noise}
%  live in main.tex)
% ===========================================================================

\begin{figure}[t]
\centering
\begin{tikzpicture}[every node/.style={font=\footnotesize}]
  \draw[->] (0,0) -- (8.5,0) node[right] {$G$};
  \draw[->] (0,0) -- (0,4.3) node[above] {$F$};
  \foreach \x/\l in {0/0,2/250,4/500,6/750,8/1000}{\draw (\x,0.06) -- (\x,-0.06) node[below] {\l};}
  \foreach \y/\l in {0/0,2/0.5,4/1}{\draw (0.06,\y) -- (-0.06,\y) node[left] {\l};}
  \draw[thick,blue!70!black,domain=0:1000,samples=60] plot ({\x*0.008},{4*exp(-0.01*\x)});
  \draw[thick,orange!80!black,domain=0:1000,samples=60] plot ({\x*0.008},{4*exp(-0.003*\x)});
  \draw[thick,green!50!black,domain=0:1000,samples=60] plot ({\x*0.008},{4*exp(-0.001*\x)});
  \draw[dashed,gray] (0,1.4715) -- (8.4,1.4715) node[right,gray] {$1/e$};
  \node[blue!70!black] at (1.9,0.75) {$\epsilon=10^{-2}$};
  \node[orange!80!black] at (4.1,1.0) {$\epsilon=3\cdot10^{-3}$};
  \node[green!50!black] at (6.5,2.5) {$\epsilon=10^{-3}$};
\end{tikzpicture}
\caption{Estimated circuit fidelity $F\approx e^{-\epsilon G}$ against gate count $G$ for three average gate errors (illustration of the estimate of \cref{sec:trainability-bg}, not measured data). The dashed line marks $F=1/e$, reached at $G=1/\epsilon$.}
\label{fig:fidelity}
\end{figure}
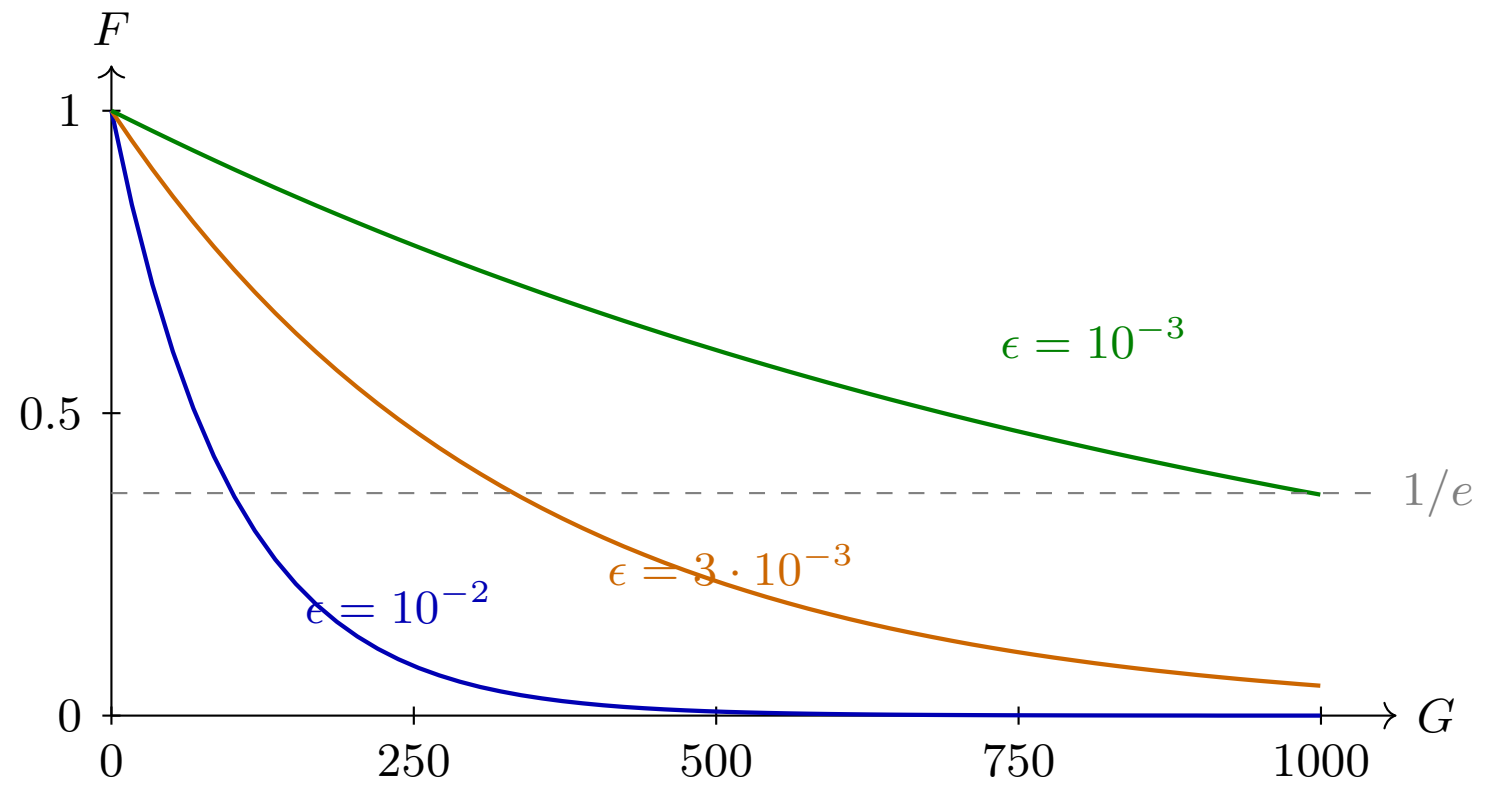

\Cref{sec:trainability-bg} gave the basic estimate: a circuit of $G$ gates with average error $\epsilon$ keeps a fidelity of roughly $e^{-\epsilon G}$, so the useful depth is $\mathcal{O}(1/\epsilon)$ (\cref{fig:fidelity}).
For a node--qubit QGNN the gate count is $G\approx L(n+m)$ (\cref{sec:notation-cost}), so the graph that fits a given error budget satisfies $n+m\lesssim 1/(\epsilon L)$.
As an order-of-magnitude example, with $L=3$ layers and $\epsilon=10^{-2}$ this gives $n+m\lesssim 33$, about 11 nodes at average degree 4, and with $\epsilon=10^{-3}$ it gives $n+m\lesssim 333$, about 110 nodes.
These figures ignore routing overhead and treat all gates as equally noisy, which is pessimistic for single-qubit gates and optimistic for dense graphs on sparse coupling maps (\cref{sec:gates}).

\paragraph{Noise-aware QGNN designs.}
The few works that address noise do so by reducing gate count.
QuanGCN~\citep{zhou2022quangcn} sparsifies node connections to cut the number of crossing-qubit gates and adds a skip connection that augments the quantum output with classical node features, and is evaluated both in simulation and on a real IBM quantum device.
\citet{zeng2025spatial} run their aggregation and classification circuits on a superconducting processor (\cref{sec:conv-var}), and the Lorentz-equivariant QGNN uses only four qubits and is presented as noise resilient~\citep{jahin2025lorentz}.
Pooling a graph to a fixed small number of nodes before the quantum part, as in QuanGCN, serves both the noise and the qubit budget.

\paragraph{Mitigation and correction.}
Zero-noise extrapolation and probabilistic error cancellation~\citep{cai2023quantum} reduce bias at the cost of additional executions that grow with circuit size, which multiplies the shot cost of \cref{eq:epoch-cost}; among the works reviewed here we found no QGNN-specific study of error mitigation, and error-corrected QGNNs are a future direction (\cref{sec:fut-ft}).
Noise also breaks exact equivariance and, through noise-induced plateaus~\citep{wang2021noise}, trainability, so shallow, sparse, symmetric designs are favoured on both counts.

\subsection{Scalability: Qubit Counts, Graph-Size Limits, and Quantum-Walk Encodings}\label{sec:scalability}
% ===========================================================================
% Section 6.4  Scalability: Qubit Counts, Graph-Size Limits, and Quantum-Walk Encodings
% (included from main.tex via \input; heading and \label{sec:scalability}
%  live in main.tex)
% ===========================================================================

\begin{figure}[t]
\centering
\begin{tikzpicture}[every node/.style={font=\footnotesize}]
  \draw[->] (0,0) -- (7.4,0) node[right] {$n$};
  \draw[->] (0,0) -- (0,3.9) node[above] {$n_q$};
  \foreach \x/\l in {0/0,2.8/20,5.6/40}{\draw (\x,0.06) -- (\x,-0.06) node[below] {\l};}
  \foreach \y/\l in {0/0,1.75/50,3.5/100}{\draw (0.06,\y) -- (-0.06,\y) node[left] {\l};}
  \draw[dashed,gray] (0,3.5) -- (7,3.5) node[right,gray] {budget};
  \draw[thick,red!70!black,domain=0:25,samples=2] plot ({\x*0.14},{4*\x*0.035});
  \draw[thick,blue!70!black,domain=0:50,samples=2] plot ({\x*0.14},{\x*0.035});
  \draw[thick,orange!80!black,domain=1:50,samples=40] plot ({\x*0.14},{ln(4*\x)/ln(2)*0.035});
  \draw[thick,green!50!black] (0,0.14) -- (7,0.14);
  \node[red!70!black,anchor=south west] at (3.5,3.3) {angle: $nF$};
  \node[blue!70!black,anchor=south west] at (5.2,1.9) {node--qubit: $n$};
  \node[orange!80!black,anchor=south west] at (4.4,0.5) {amplitude: $\log nF$};
  \node[green!50!black,anchor=south west] at (0.3,0.22) {sampled / hybrid: $F$};
\end{tikzpicture}
\caption{Schematic scaling of qubit count with graph size for $F=4$ features per node (class-level formulas of \cref{tab:encodings}; the dashed line is an illustrative device budget of about 100 qubits, not a hardware specification).}
\label{fig:qubit-scaling}
\end{figure}
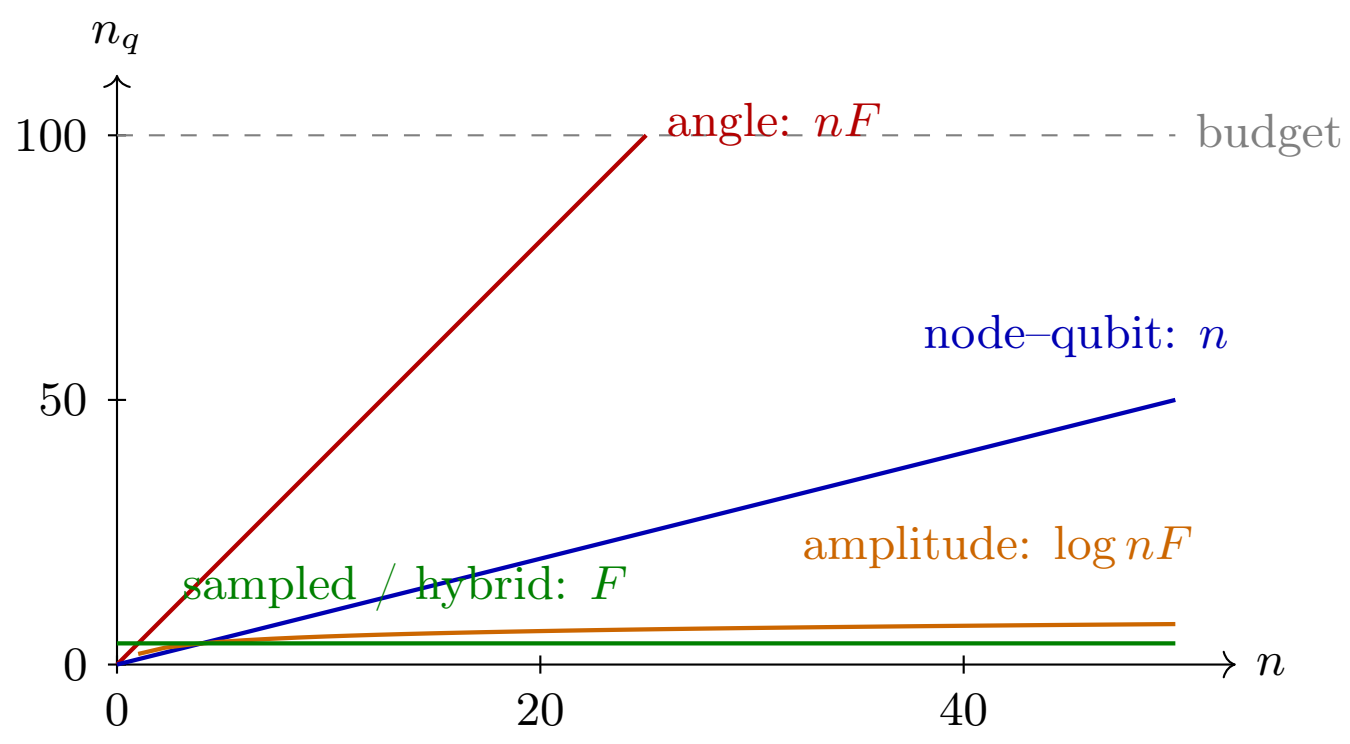

With the node--qubit mapping, $n_q=n$ and the noise budget of \cref{sec:noise} limit graphs to tens or a few hundred nodes; \cref{fig:qubit-scaling} contrasts how the main strategies scale.
Five strategies are used to go further (\cref{tab:scale}).

\begin{table}[t]
\centering
\caption{Strategies for scaling QGNNs to larger graphs.}
\label{tab:scale}
\renewcommand{\arraystretch}{1.3}
\footnotesize
\begin{tabular}{@{}p{2.5cm}p{3.4cm}p{2.4cm}p{3.6cm}@{}}
\toprule
\textbf{Strategy} & \textbf{Mechanism and examples} & \textbf{Qubits} & \textbf{Trade-off} \\
\midrule
Sampling, ego-graphs & process local subgraphs~\citep{ai2024ego,faria2025inductive,giang2026scalable} & set by $F$ & runs grow with nodes; local view \\
Graph compression, pooling & reduce nodes and features first~\citep{casals2025guided,zhou2022quangcn,liu2026rotation} & small, fixed & information loss; classical model needed \\
Compressed encodings & amplitude encoding~\citep{zheng2021quantum,ray2023hybrid} & $\mathcal{O}(\log nF)$ & state-preparation cost \\
Block / walk encodings & block-encode $\A$~\citep{liao2024graph,ye2025spectral} & $\mathcal{O}(\log n)$ plus ancillas & data access; fault tolerance \\
Size-independent dynamics & shared parameters; pre-training on small graphs~\citep{raj2026scalable} & grows with graph in simulation & preprint \\
\bottomrule
\end{tabular}
\end{table}

\paragraph{Compression before the quantum part.}
\citet{casals2025guided} observe that hardware limits the data that can be encoded and that existing studies simplify datasets by hand or use artificial graphs; their guided graph compression trains a graph autoencoder that reduces both the number of nodes and the feature dimension jointly with a downstream classifier, quantum or classical, on quark--gluon jet tagging, and reports that this outperforms using the autoencoder as a separate preprocessing step.
The approach makes realistic datasets usable but moves part of the learning to a classical model, so it belongs with the hybrid designs of \cref{sec:tax-hybrid}.

\paragraph{Quantum-walk and block encodings.}
Amplitude encoding stores $\A$ in $\mathcal{O}(\log n)$ qubits but at the preparation cost of \cref{sec:vqc}.
Quantum-walk and block-encoding representations of $\A$ or $\Lap$ implement graph operators as unitaries, which underlies the algorithmic convolutions of \cref{sec:conv-alg}~\citep{liao2024graph,ye2025spectral}; quantum walks with learned coins were used earlier in quantum-inspired models~\citep{dernbach2019quantum}.
Their speedups assume efficient data access and, for phase estimation, fault tolerance, so they are not near-term scalability tools.

\paragraph{Takeaway.}
Near-term scaling comes from decoupling circuit size from graph size (sampling, compression), at the price of locality or classical preprocessing; scaling with the whole graph in one circuit is available only in simulation or under fault-tolerance assumptions.
This is the reason why the evidence in \cref{sec:tax-compare} is concentrated on small graphs, and the issue returns as an open problem in \cref{sec:ch-scale}.

% ===========================================================================
\section{Applications}\label{sec:applications}
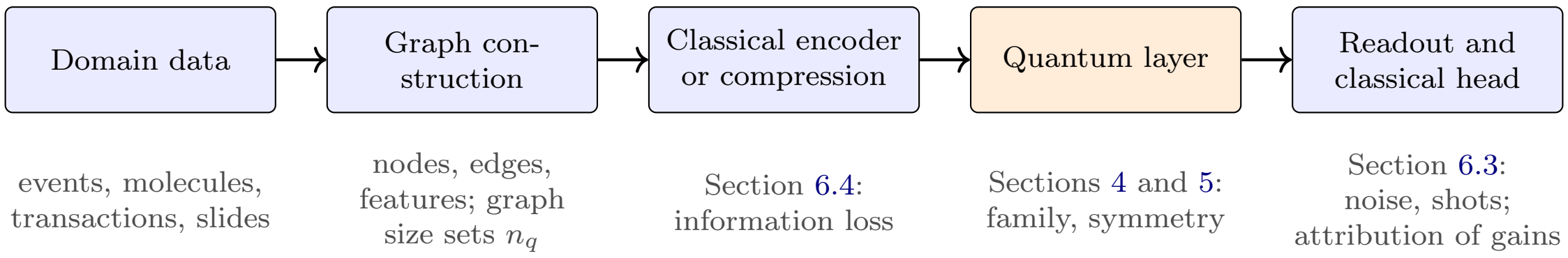
\begin{figure}[t]
\centering
\begin{tikzpicture}[every node/.style={font=\scriptsize},
  box/.style={draw,rounded corners=2pt,fill=blue!8,text width=2.35cm,minimum height=1.0cm,align=center,inner sep=3pt},
  note/.style={text width=2.6cm,align=center,gray!60!black}]
  \node[box] (a) at (0,0) {Domain data};
  \node[box] (b) at (3.05,0) {Graph construction};
  \node[box] (c) at (6.1,0) {Classical encoder or compression};
  \node[box,fill=orange!15] (d) at (9.15,0) {Quantum layer};
  \node[box] (e) at (12.2,0) {Readout and classical head};
  \foreach \f/\t in {a/b,b/c,c/d,d/e}{\draw[->,thick] (\f.east) -- (\t.west);}
  \node[note] at (0,-1.35) {events, molecules, transactions, slides};
  \node[note] at (3.05,-1.35) {nodes, edges, features; graph size sets $n_q$};
  \node[note] at (6.1,-1.35) {\cref{sec:scalability}: information loss};
  \node[note] at (9.15,-1.35) {\cref{sec:taxonomy,sec:symmetry}: family, symmetry};
  \node[note] at (12.2,-1.35) {\cref{sec:noise}: noise, shots; attribution of gains};
\end{tikzpicture}
\caption{Typical application pipeline of the QGNN studies reviewed in this section, with the places where the design choices of earlier sections enter.}
\label{fig:app-pipeline}
\end{figure}

% ===========================================================================
Application studies show where QGNNs are actually used and what kind of evidence exists for each domain.
In almost all of them the graph is built from domain data, a classical stage reduces it to what a small circuit can accept, and the quantum part is a compact layer or classifier (\cref{fig:app-pipeline}); the design choices of \cref{sec:taxonomy,sec:symmetry,sec:expr-train-scale} enter at the corresponding steps.
\Cref{tab:apps} at the end of the section summarizes the works reviewed.

\subsection{High-Energy Physics: Track Reconstruction and Jet Tagging}\label{sec:app-hep}
% ===========================================================================
% Section 7.1  High-Energy Physics: Track Reconstruction and Jet Tagging
% (included from main.tex via \input; heading and \label{sec:app-hep}
%  live in main.tex)
% ===========================================================================

High-energy physics is the most developed application of QGNNs, because its data are naturally graphs (detector hits and track segments, or the constituents of a jet) and the planned high-luminosity upgrade of the Large Hadron Collider multiplies the combinatorial load.

\paragraph{Track reconstruction.}
\citet{tuysuz2020quantum} replace the edge and node networks of a graph-based tracking model with quantum circuits, and report an iterative model that improves on an earlier, over-simplified tree-tensor-network version whose accuracy converged poorly.
The follow-up of \citet{tuysuz2021hybrid} converts the graph network into a hybrid model with variational quantum layers, shows that it performs similarly to the classical approach, and compares parametrized circuits of different expressibility and entangling capacity to see what training behaviour to expect; this is the hybrid pattern of \cref{sec:tax-hybrid}.

\paragraph{Jet discrimination and tagging.}
\citet{chen2025jet} propose the quantum complete graph neural network, a variational model for complete graphs in which an index register of $\lceil\log_2N\rceil$ qubits addresses the $N$ particles of a jet; they argue a polynomial speedup over classical counterparts from quantum parallelism and benchmark it against classical GNNs on the Top (binary) and JetNet (multi-class) datasets.
The symmetry-based designs of \cref{sec:symmetry} come from this domain: the comparison of \citet{forestano2024comparison} and the Lorentz-equivariant model of \citet{jahin2025lorentz}.
Because jets are large for current devices, graph compression before the quantum classifier~\citep{casals2025guided} (\cref{sec:scalability}) is used to reach realistic data.

\paragraph{Assessment.}
Results are mostly from simulation, on small subsets, with accuracy at or near the classical level; the claimed speedup of \citet{chen2025jet} is an argument about quantum parallelism rather than a measured wall-clock gain.

\subsection{Chemistry, Materials, and Drug Discovery}\label{sec:app-chem}
% ===========================================================================
% Section 7.2  Chemistry, Materials, and Drug Discovery
% (included from main.tex via \input; heading and \label{sec:app-chem}
%  live in main.tex)
% ===========================================================================

Molecules and crystals are graphs of atoms and bonds, and property prediction on QM9, perovskite, and similar datasets is the standard test bed.
\citet{ryu2023quantum} discuss a QGNN to predict chemical and physical properties of molecules and materials, together with a review of classical GNNs for materials and of existing QGNNs.
\citet{vitz2024hybrid} build a gradient-free hybrid quantum--classical convolutional GNN for the formation energy of perovskites; the model is competitive with a classical GNN and with XGBoost, although the classical methods are slightly better.
\citet{piperno2025study} design a QGNN for water molecules that predicts intramolecular forces and total energy, with an embedding update analogous to classical GNNs, layers devoted to interactions of different orders, and SWAP gates that emulate the symmetry of the problem; the model gives similar outcomes to earlier models with fewer parameters, which makes it interpretable.
In drug discovery, \citet{villalba2025quantum} combine a QGNN, with attention layers and self-distillation, and a QAOA-inspired ranking built on variational quantum eigensolver outputs to identify candidate serine neutralizers in QM9, and \citet{lu2025quantum} introduce quantum-embedded GNN models in a hybrid framework for molecular property prediction.
The sampled aggregator of \citet{faria2025inductive} is also evaluated on QM9 (\cref{sec:mp-ego}).

\paragraph{Assessment.}
The datasets contain small molecules because of the qubit budget, the 3D symmetries of \cref{sec:sym-rot} are rarely enforced, and the reported advantage is parameter efficiency or competitiveness, not better accuracy.

\subsection{Finance}\label{sec:app-finance}
% ===========================================================================
% Section 7.3  Finance
% (included from main.tex via \input; heading and \label{sec:app-finance}
%  live in main.tex)
% ===========================================================================

Financial fraud detection treats accounts and transactions as a graph in which fraudulent activity appears as unusual connectivity.
\citet{innan2024financial} use a QGNN with variational quantum circuits and compare it with classical GNNs on a real-world fraud dataset, reporting an AUC of 0.85 and an advantage over the classical models.
The result is a single-dataset, simulation-level study of a heavily imbalanced problem, so it should be read as feasibility evidence; the lack of standard financial graph benchmarks is part of the evaluation issues of \cref{sec:ch-fair}.

\subsection{Healthcare and Digital Pathology}\label{sec:app-health}
% ===========================================================================
% Section 7.4  Healthcare and Digital Pathology
% (included from main.tex via \input; heading and \label{sec:app-health}
%  live in main.tex)
% ===========================================================================

In digital pathology, a tissue slide is modelled as a graph of cells whose neighbourhoods carry diagnostic information.
\citet{ray2023hybrid} combine a GNN with a variational quantum classifier for breast-cancer subtyping and find that amplitude encoding of the embedding outperforms classical compression, that end-to-end training of both parts improves over a fixed GNN, and that the hybrid model is on par with state-of-the-art classical GNNs (\cref{sec:tax-hybrid}).
This is a representative case of the hybrid pattern: the graph processing stays classical, and the quantum part is a compact classifier whose benefit remains to be isolated from that of the encoding (\cref{sec:ch-fair}).

\subsection{Earth Observation and Remote Sensing}\label{sec:app-eo}
% ===========================================================================
% Section 7.5  Earth Observation and Remote Sensing
% (included from main.tex via \input; heading and \label{sec:app-eo}
%  live in main.tex)
% ===========================================================================

Climate and earth-observation data are graph-structured through spatial grids and teleconnections.
\citet{mauro2024hybrid} forecast the Oceanic Ni\~no Index, a key indicator of El Ni\~no, with a hybrid network in which a graph convolutional part supplies global dependencies and a quantum multilayer perceptron with amplitude encoding replaces the dense head, motivated by the limits of quantum convolutional networks for global-scale features (\cref{sec:tax-hybrid}).
Climate series are long and high-dimensional, which makes compressed encodings attractive, but the evidence is again limited to simulation.

\subsection{Traffic and Transportation}\label{sec:app-traffic}
% ===========================================================================
% Section 7.6  Traffic and Transportation
% (included from main.tex via \input; heading and \label{sec:app-traffic}
%  live in main.tex)
% ===========================================================================

Road networks are graphs with time-varying signals, so traffic prediction needs temporal and spatial modelling together.
\citet{qu2023temporal} propose a temporal--spatial quantum graph convolutional network for congestion prediction: a closed-form solution of the problem in the Schr\"odinger picture provides temporal features, which are then fed into a quantum graph convolutional network that captures the spatial structure.
It is a convolutional QGNN (\cref{sec:tax-conv}) with a physics-motivated temporal front end.
The setting also suits recurrent models (\cref{sec:mp-recurrent}), but we found no hardware-scale study; the compute budget of citywide networks is far above the graph sizes of \cref{sec:noise}.

\subsection{Computer Vision, IoT, and Other Domains}\label{sec:app-cv}
% ===========================================================================
% Section 7.7  Computer Vision, IoT, and Other Domains
% (included from main.tex via \input; heading and \label{sec:app-cv}
%  live in main.tex)
% ===========================================================================

Beyond the domains above, QGNNs have been proposed for communication networks and genomics, and the graph view of images and point clouds underlies the 3D designs of \cref{sec:sym-rot}.
In wireless and IoT systems, a network of devices and links is a graph: \citet{giang2026scalable} decompose it into subgraphs processed by a shared-parameter circuit with node and edge attributes (\cref{sec:mp-ego}), and the survey of \citet{tung2025graph} lists the integration of GNNs with quantum computing among the open directions for next-generation IoT.
In genomics, \citet{ahmadkhaniha2026edge} propose a fully quantum graph convolution trained without labels, with edge-local message passing that reduces the qubit requirement from $\mathcal{O}(Nn)$ to $\mathcal{O}(n)$ for $N$ nodes and $n$-qubit feature registers, and evaluate it on Cora and a large single-nucleotide-polymorphism dataset, where it is competitive with prior quantum and hybrid approaches.
Compared with the physics-centred domains, these settings give the clearest case for qubit-efficient designs, and they are still at the proposal stage with no head-to-head comparison against strong classical baselines.

\paragraph{Cross-domain summary.}
\begin{table}[t]
\centering
\caption{Application studies of QGNNs. ``Outcome'' is what the authors report, in simulation unless stated otherwise; the family refers to \cref{sec:taxonomy}.}
\label{tab:apps}
\renewcommand{\arraystretch}{1.3}
\footnotesize
\begin{tabular}{@{}p{2.1cm}p{2.5cm}p{3.1cm}p{2.0cm}p{3.6cm}@{}}
\toprule
\textbf{Domain} & \textbf{Graph} & \textbf{Works} & \textbf{Family} & \textbf{Outcome (as reported)} \\
\midrule
HEP tracking & hits, segments & \citet{tuysuz2020quantum,tuysuz2021hybrid} & hybrid & similar to the classical GNN \\
HEP jets & jet constituents & \citet{chen2025jet,forestano2024comparison,jahin2025lorentz,casals2025guided} & in-circuit, EQGC, hybrid & vs.\ classical GNNs: speedup argued; higher AUC; efficiency gains \\
Chemistry, materials & atoms, bonds & \citet{ryu2023quantum,vitz2024hybrid,piperno2025study,villalba2025quantum,lu2025quantum} & hybrid and fully quantum & competitive; fewer parameters \\
Finance & accounts, transactions & \citet{innan2024financial} & VQC-based QGNN & AUC 0.85, above classical GNNs \\
Healthcare & cells in tissue & \citet{ray2023hybrid} & hybrid & on par with classical GNNs \\
Earth observation & grid, teleconnections & \citet{mauro2024hybrid} & hybrid & ONI forecasting \\
Traffic & road network & \citet{qu2023temporal} & convolutional & congestion prediction \\
Genomics, IoT & SNP data; devices & \citet{ahmadkhaniha2026edge,giang2026scalable} & edge-local; ego-graph & competitive with prior quantum and hybrid models \\
\bottomrule
\end{tabular}
\end{table}
Three patterns stand out.
Hybrid designs dominate, because they keep the circuit small; fully quantum designs appear where symmetry or qubit efficiency is the goal.
Outcomes are, at best, parity with classical models or parameter efficiency, and no domain yet shows an end-to-end advantage that survives a parameter-matched classical baseline (\cref{sec:ch-fair}).
Finally, only the HEP studies adopt symmetry-based designs systematically (\cref{sec:symmetry}); in chemistry, 3D symmetry is a clear open opportunity (\cref{sec:sym-rot}).

% ===========================================================================
\section{Datasets, Benchmarks, and Evaluation}\label{sec:datasets}
% ===========================================================================
The evidence for QGNNs is only as strong as the data and protocols behind it.
This section surveys the datasets used in the works reviewed, the evaluation practices and where they fall short, and what is needed to reproduce and compare results.

\subsection{Graph Datasets Used}\label{sec:datasets-used}
% ===========================================================================
% Section 8.1  Graph Datasets Used
% (included from main.tex via \input; heading and \label{sec:datasets-used}
%  live in main.tex)
% ===========================================================================

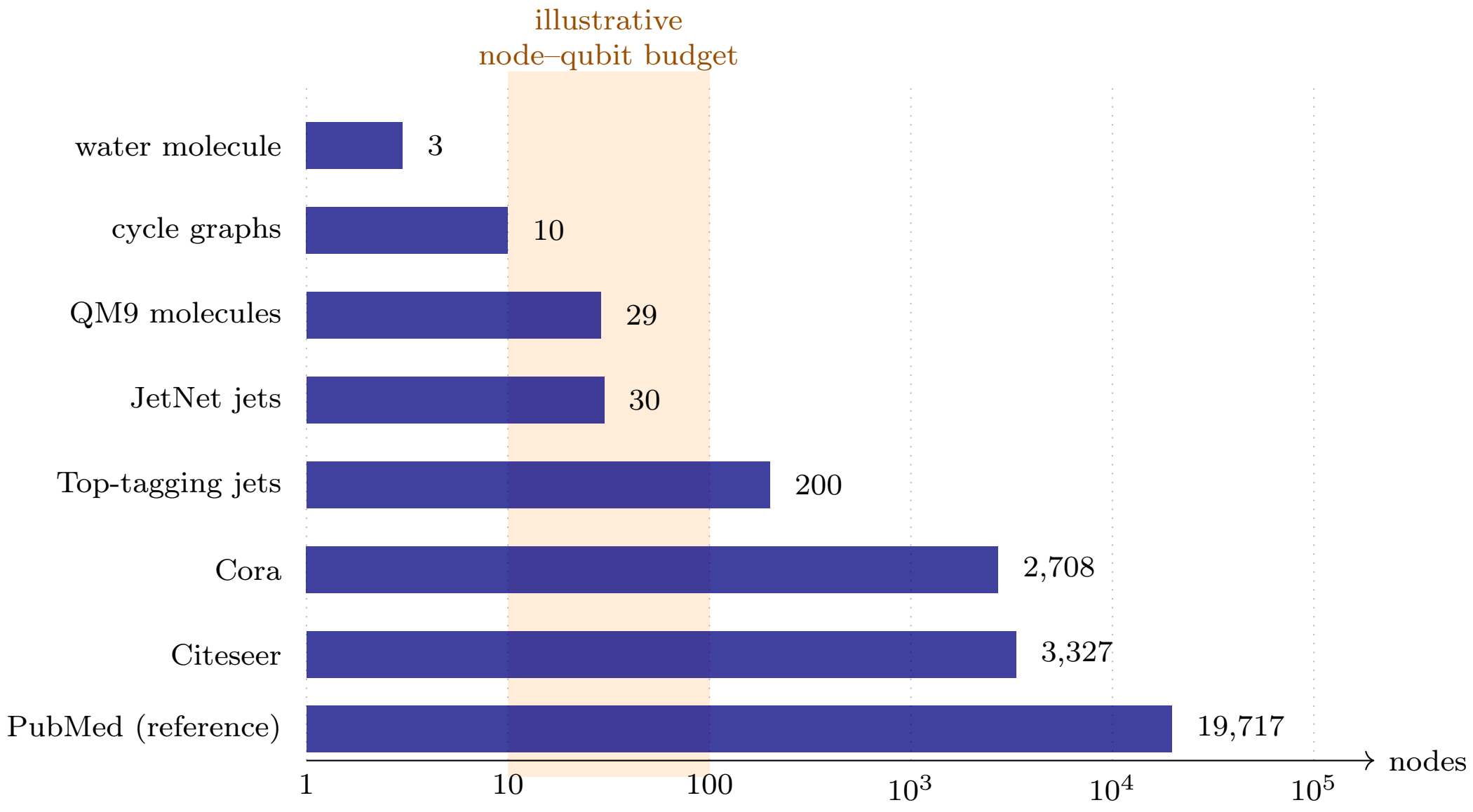
\begin{figure}[t]
\centering
\begin{tikzpicture}[every node/.style={font=\scriptsize}]
  \def\s{1.9}
  \fill[orange!15] ({1*\s},-0.35) rectangle ({2*\s},6.3);
  \node[orange!60!black,align=center] at ({1.5*\s},6.6) {illustrative\\node--qubit budget};
  \foreach \d/\l in {0/1,1/10,2/100,3/$10^3$,4/$10^4$,5/$10^5$}{
    \draw[gray!50,dotted] ({\d*\s},-0.2) -- ({\d*\s},6.2);
    \node[below] at ({\d*\s},-0.2) {\l};}
  \draw[->] (0,-0.2) -- ({5.3*\s},-0.2) node[right] {nodes};
  \foreach \y/\n/\lg/\v in {
    5.6/{water molecule}/0.477/{3},
    4.8/{cycle graphs}/1.0/{10},
    4.0/{QM9 molecules}/1.462/{29},
    3.2/{JetNet jets}/1.477/{30},
    2.4/{Top-tagging jets}/2.301/{200},
    1.6/{Cora}/3.433/{2{,}708},
    0.8/{Citeseer}/3.522/{3{,}327},
    0.1/{PubMed (reference)}/4.295/{19{,}717}}{
    \fill[blue!50!black,opacity=0.75] (0,{\y-0.22}) rectangle ({\lg*\s},{\y+0.22});
    \node[anchor=east] at (-0.1,\y) {\n};
    \node[anchor=west] at ({\lg*\s+0.1},\y) {\v};}
\end{tikzpicture}
\caption{Maximum number of nodes per graph (molecules, jets, toy graphs) or total number of nodes (citation networks) of datasets used in the QGNN studies reviewed here, on a logarithmic scale; PubMed is shown for reference only. The shaded band marks an illustrative node--qubit budget of tens to a hundred nodes (\cref{sec:noise}). Citation graphs enter quantum studies as subsets, subgraphs, or sampled neighbourhoods.}
\label{fig:graph-size}
\end{figure}

QGNN studies do not share a benchmark.
Each application community brings its own data, and the choice is dominated by one constraint: the graph must fit a circuit, or be reduced until it does.
\Cref{tab:datasets} lists the datasets used in the works reviewed in this survey, and \cref{fig:graph-size} compares their size with the node--qubit budget estimated in \cref{sec:noise}.

\paragraph{Molecules, materials, and drug--target data.}
QM9 is the most common molecular benchmark: \citet{faria2025inductive} use it for node-level regression with sampled aggregators, and \citet{villalba2025quantum} for ionization-potential and binding-energy prediction.
Smaller, task-specific sets are used elsewhere: perovskite formation energies~\citep{vitz2024hybrid} and configurations of the water molecule, where the model predicts intramolecular forces and energy~\citep{piperno2025study}.
A recent hybrid study evaluates on the DAVIS kinase benchmark for drug--target affinity, reporting the concordance index and RMSE~\citep{hqgnn2026representational}.
Molecules are small graphs, and QM9 molecules have at most 29 atoms, so they are the one real-world class that approaches the qubit budget without subsampling.

\paragraph{High-energy physics.}
Tracking studies use simulated collision events turned into hit graphs~\citep{tuysuz2020quantum,tuysuz2021hybrid}; jet studies use the Top and JetNet datasets~\citep{chen2025jet} and quark--gluon and electron--photon jet samples~\citep{jahin2025lorentz,forestano2024comparison}.
Jets contain tens to hundreds of constituents, so even these graphs are usually truncated, compressed~\citep{casals2025guided}, or evaluated on small subsets.

\paragraph{Citation and benchmark graphs.}
Node-classification benchmarks are used to compare with classical GNNs: Cora subsets~\citep{hu2022design}, Citeseer subgraphs~\citep{zeng2025spatial}, Cora together with a genomic dataset~\citep{ahmadkhaniha2026edge}, and five open benchmark graphs in the semi-supervised setting~\citep{lo2026quantum}.
These graphs have thousands of nodes, so quantum studies must subsample or process local neighbourhoods; different subsets in different papers make results hard to compare.

\paragraph{Synthetic graphs for expressivity.}
Expressivity claims are tested on purpose-built graphs: one-cycle versus two-cycle graphs of 6 to 10 nodes~\citep{mernyei2022equivariant}, and Cai--F\"urer--Immerman graphs together with QM9 and the travelling-salesperson problem~\citep{raj2026scalable}.
These sets are small enough for exact simulation and informative about the WL hierarchy, but they are not representative of practical tasks.

\begin{table}[t]
\centering
\caption{Datasets used in the QGNN works reviewed in this survey. Sizes are maximum nodes per graph (or total nodes) where known from the standard descriptions of the datasets; ``--'' means not specified in the works we read.}
\label{tab:datasets}
\renewcommand{\arraystretch}{1.3}
\footnotesize
\begin{tabular}{@{}p{2.2cm}p{3.5cm}p{2.0cm}p{4.2cm}@{}}
\toprule
\textbf{Domain} & \textbf{Dataset} & \textbf{Size} & \textbf{Used in} \\
\midrule
Molecules, materials & QM9 & $\le 29$ atoms & \citet{faria2025inductive,villalba2025quantum} \\
 & perovskite formation energies & -- & \citet{vitz2024hybrid} \\
 & water-molecule configurations & 3 atoms & \citet{piperno2025study} \\
 & DAVIS kinase (affinity) & -- & \citet{hqgnn2026representational} \\
HEP & tracking events & -- & \citet{tuysuz2020quantum,tuysuz2021hybrid} \\
 & Top, JetNet jets & $\le 200$; $\le 30$ & \citet{chen2025jet} \\
 & quark--gluon, electron--photon jets & -- & \citet{jahin2025lorentz,forestano2024comparison} \\
Citation graphs & Cora, Citeseer (subsets) & 2,708; 3,327 & \citet{hu2022design,zeng2025spatial,ahmadkhaniha2026edge,lo2026quantum} \\
Synthetic & cycle graphs; CFI graphs; TSP & 6--10 nodes; -- & \citet{mernyei2022equivariant,raj2026scalable} \\
Other & fraud graph; ONI series; SNP data & -- & \citet{innan2024financial,mauro2024hybrid,ahmadkhaniha2026edge} \\
\bottomrule
\end{tabular}
\end{table}

\paragraph{Observations.}
Only toy graphs, molecules, and jets sit near the budget of \cref{fig:graph-size}; every other benchmark requires subsampling, sampling, or compression, which changes the task.
Standardized resources exist on the classical side, such as the Open Graph Benchmark with fixed splits~\citep{hu2020open} and the more recent GraphBench suite with strong baselines~\citep{stoll2025graphbench}, but QGNN papers rarely adopt their splits, so a common, size-aware benchmark is missing (\cref{sec:ch-fair}).

\subsection{Evaluation Protocols and Metrics}\label{sec:eval}
% ===========================================================================
% Section 8.2  Evaluation Protocols and Metrics
% (included from main.tex via \input; heading and \label{sec:eval}
%  live in main.tex)
% ===========================================================================

\paragraph{Task metrics.}
Quality is measured with the standard metrics of each task: accuracy and AUC for classification (jets, fraud, pathology), MAE, RMSE, and $R^2$ for molecular regression, and the concordance index for drug--target affinity.
These metrics alone say little about a quantum model, because what distinguishes it is the cost of reaching a given score.
We therefore recommend reporting alongside them the resource quantities of \cref{sec:notation-cost}: qubits $n_q$, circuit depth and gate count, trainable parameters $P$, shots $S$ per estimate, and total circuit executions as in \cref{eq:epoch-cost}.

\paragraph{Baselines and fairness.}
Most reported gains are parity or modest improvements over classical models (\cref{tab:apps}), so the quality of the baseline decides the conclusion.
Three practices help.
First, parameter matching: \citet{faria2025inductive} build a classical GNN with the same number of learnable parameters as the quantum model, and find that the classical model trains better while the quantum model generalizes better as molecules grow.
Second, encoding controls: the same authors add classical baselines that use the same Fourier feature encoding, to rule out that the encoding explains the gap; similarly, \citet{ray2023hybrid} compare amplitude encoding with classical compression at equal embedding size.
Third, strong and tuned baselines: studies of classical GNNs show that under a uniform protocol and a fair hyperparameter search, simple structure-agnostic baselines are competitive on several graph-classification datasets~\citep{errica2020fair}, and large-scale benchmarks of quantum machine learning find that simple classical models often outperform their quantum counterparts~\citep{bowles2024better}.

\paragraph{Statistics.}
Single-run, single-baseline comparisons are common; the fraud-detection study reports an AUC of 0.85 against 0.77 for a GraphSAGE baseline~\citep{innan2024financial}, which is informative but cannot support firm conclusions without seeds and intervals.
A recent hybrid study shows how much multi-seed evaluation can change the picture: the apparent edge over a dimension-matched classical bottleneck on a kinase-affinity benchmark is close to random ranking, a full classical model is consistently better, and the authors note that this is consistent with near-barren-plateau behaviour of the circuit at four qubits~\citep{hqgnn2026representational}.
Reporting means with confidence intervals over several seeds, with paired tests across datasets, is the standard now being pushed on the classical side~\citep{stoll2025graphbench}.

\paragraph{Execution mode.}
Results should state whether circuits were run on an exact statevector simulator, a shot-based simulator, a noisy simulator, or hardware, because these give different answers to the same question.
Hardware evidence is rare: \citet{zeng2025spatial} run their circuits on a superconducting processor, and QuanGCN is evaluated on a small IBM device~\citep{zhou2022quangcn}.

\begin{table}[t]
\centering
\caption{Minimum reporting protocol proposed for QGNN studies, with supporting examples from the works reviewed.}
\label{tab:protocol}
\renewcommand{\arraystretch}{1.3}
\footnotesize
\begin{tabular}{@{}p{2.6cm}p{5.0cm}p{5.0cm}@{}}
\toprule
\textbf{Item} & \textbf{What to report} & \textbf{Why / example} \\
\midrule
Model resources & $n_q$, depth, gates, $P$, encoding, shots & lets cost be compared (\cref{sec:notation-cost}) \\
Execution mode & statevector, shots, noise model, or hardware & hardware runs are rare~\citep{zeng2025spatial} \\
Baselines & parameter-matched, equally tuned; encoding-matched controls & \citet{faria2025inductive} \\
Seeds and statistics & several seeds, mean and interval, paired tests & multi-seed changes conclusions~\citep{hqgnn2026representational} \\
Data protocol & fixed splits, subset indices, sampling procedure, graph size vs.\ $n_q$ & subsets differ across papers (\cref{tab:datasets}) \\
Ablations & replace the quantum layer by a classical layer of equal width & compression and encoding checks~\citep{ray2023hybrid} \\
Cost & circuit executions, wall-clock, simulator used & \cref{eq:epoch-cost} \\
Artefacts & code, seeds, library versions, circuit definitions & \cref{sec:repro} \\
\bottomrule
\end{tabular}
\end{table}

\subsection{Reproducibility and Available Codebases}\label{sec:repro}
% ===========================================================================
% Section 8.3  Reproducibility and Available Codebases
% (included from main.tex via \input; heading and \label{sec:repro}
%  live in main.tex)
% ===========================================================================

\paragraph{Software stack.}
Most QGNN studies combine a quantum-circuit library with a classical graph-learning library.
On the quantum side, differentiable-programming frameworks such as PennyLane~\citep{bergholm2018pennylane} provide circuits with automatic differentiation, parameter-shift gradients (\cref{eq:psr}), and interfaces to classical deep-learning libraries and to hardware backends; the hybrid designs of \cref{sec:tax-hybrid} typically pair such a framework with a graph library such as PyTorch Geometric~\citep{fey2019fast}.
We do not tabulate which paper uses which library, because the information is not given uniformly; it is itself part of the reporting problem.

\paragraph{What limits reproducibility.}
Four issues recur in the studies reviewed.
(i) \emph{Simulation cost}: exact simulation scales exponentially with the number of qubits, so results come from a small number of qubits, and conclusions from four-qubit models (as in the study of \citet{hqgnn2026representational}) may not transfer to larger ones.
(ii) \emph{Hidden degrees of freedom}: the encoding, the data subset, the number of shots, the noise model, and the initialization all change results, and are not always specified; the subset differences of \cref{tab:datasets} are an example.
(iii) \emph{Preprint status}: a number of the most recent results~\citep{raj2026scalable,ahmadkhaniha2026edge,lo2026quantum} are preprints whose code and settings may still change.
(iv) \emph{Asymmetric effort}: quantum models are often tuned carefully while classical baselines are run with default settings, which inflates the reported benefit.

\paragraph{Recommendations.}
Beyond the protocol of \cref{tab:protocol}, we suggest a small, shared benchmark suite spanning the scales at which QGNNs are credible: exact toy tasks for expressivity (cycle and CFI graphs), molecular tasks with a size cap (QM9 subsets), a jet or tracking task from HEP, and a citation subgraph with a fixed neighbourhood-sampling protocol, all with fixed splits and seeds.
Such a suite would turn the isolated results of \cref{sec:applications} into comparable evidence and would make the claims of \cref{sec:tax-compare} testable; we return to it in \cref{sec:ch-fair}.

% ===========================================================================
\section{Challenges and Open Problems}\label{sec:challenges}
\begin{figure}[t]
\centering
\begin{tikzpicture}[every node/.style={font=\footnotesize},
  box/.style={draw,rounded corners=2pt,fill=blue!8,minimum height=7mm,text width=2.5cm,align=center,inner sep=3pt},
  ctr/.style={draw,rounded corners=2pt,fill=orange!15,minimum height=8mm,text width=2.8cm,align=center,inner sep=3pt},
  lab/.style={font=\scriptsize,gray!50!black,fill=white,inner sep=1pt}]
  \node[box] (B) at (5,2.5) {Trainability};
  \node[box] (A) at (0.3,0.6) {Hardware and noise};
  \node[box] (C) at (9.7,0.6) {Expressivity};
  \node[box] (E) at (1.6,-2.4) {Scalability};
  \node[box] (D) at (8.4,-2.4) {Classical simulability};
  \node[ctr] (X) at (5,-0.3) {Evaluation and claims};
  \draw (A) -- (B) node[lab,midway] {noise-induced plateaus};
  \draw (A) -- (E) node[lab,midway] {depth budget};
  \draw (B) -- (C) node[lab,midway] {Lie-algebra size};
  \draw (B) -- (D) node[lab,pos=0.55] {restriction};
  \draw (E) -- (D) node[lab,midway] {classical pre-processing};
  \foreach \n in {A,B,C,D,E}{\draw[dotted,gray] (X) -- (\n);}
\end{tikzpicture}
\caption{Interactions among the challenges of this section. Solid links name the mechanism that couples two challenges; the dotted links indicate that every claim must be evaluated against all of them.}
\label{fig:challenge-map}
\end{figure}
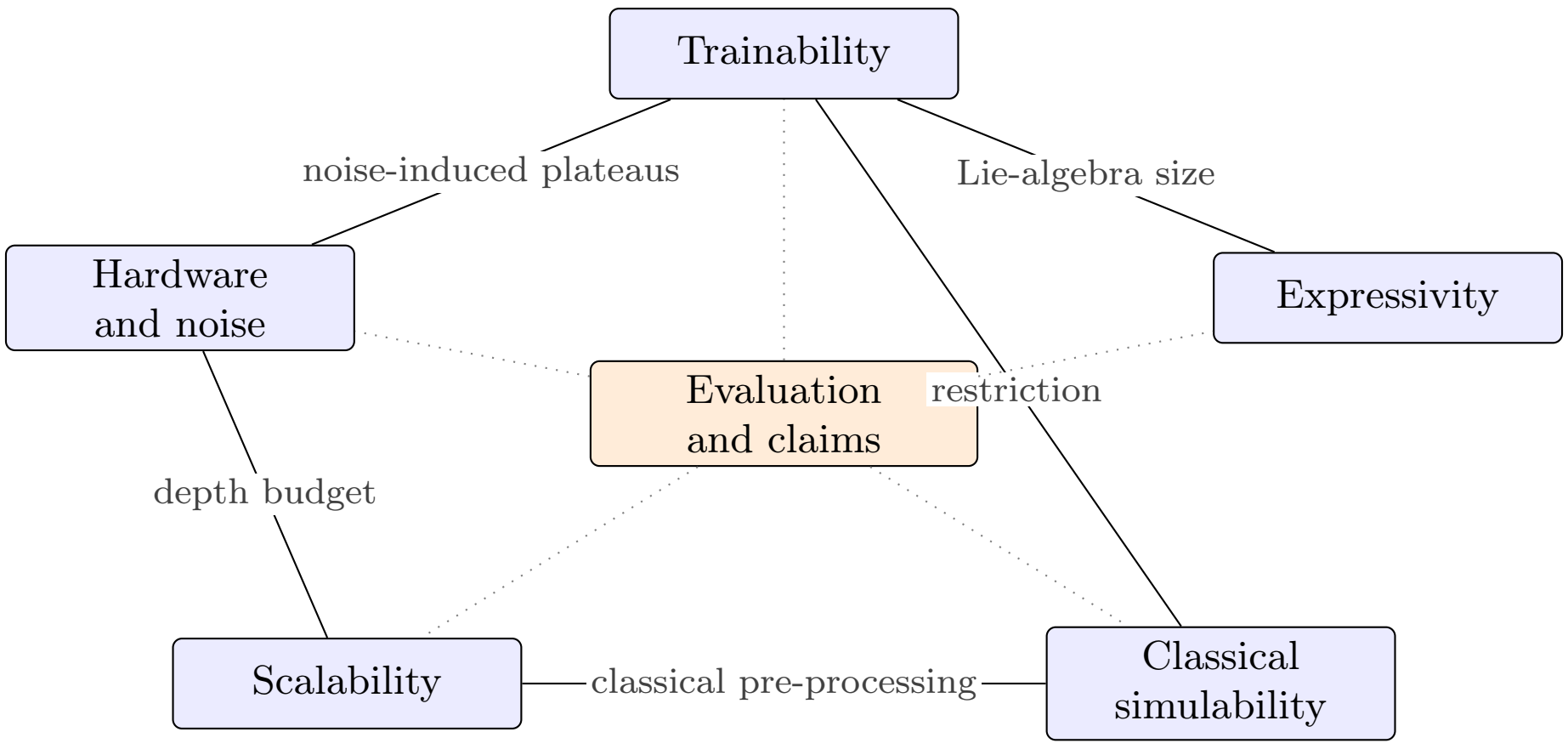

% ===========================================================================
The evidence of \cref{sec:taxonomy,sec:symmetry,sec:expr-train-scale,sec:applications,sec:datasets} shows that QGNNs work on small graphs in simulation; it does not yet show that they work at scale, on hardware, or better than classical models.
The gap has five parts, which map to the subsections below and which are the general difficulties of quantum machine learning~\citep{cerezo2022challenges} seen through graph structure.
They are not independent (\cref{fig:challenge-map}): noise limits depth and flattens landscapes, the restrictions that make circuits trainable also tend to make them simulable, and the compromises needed for scale shrink the quantum part of the model.
\Cref{tab:challenges} summarizes what is known and what is missing.

\begin{table}[t]
\centering
\caption{Summary of the open challenges for QGNNs.}
\label{tab:challenges}
\renewcommand{\arraystretch}{1.3}
\footnotesize
\begin{tabular}{@{}p{2.3cm}p{4.6cm}p{4.6cm}p{1.2cm}@{}}
\toprule
\textbf{Challenge} & \textbf{Evidence so far} & \textbf{Missing} & \textbf{Section} \\
\midrule
Hardware, noise & gate-budget estimates; two small hardware runs & hardware-aware graph embedding; symmetry under noise; logical-qubit estimates & \ref{sec:ch-hardware} \\
Trainability & numerical absence of plateaus vs.\ qubits; symmetry-based guarantees & graph-dependent Lie-algebra analysis; scaling by graph, not qubits & \ref{sec:ch-train} \\
Scaling & sampling, compression, in-circuit dynamics in simulation & information-loss accounting; end-to-end resource estimates & \ref{sec:ch-scale} \\
Fair comparison & parameter-matched and encoding-controlled baselines in a few studies & shared, size-aware benchmark; cost-normalized reporting & \ref{sec:ch-fair} \\
Theory & universality, WL-level, QPE results & separation at equal cost; size generalization; advantage without data-access assumptions & \ref{sec:ch-theory} \\
\bottomrule
\end{tabular}
\end{table}

\subsection{Hardware Constraints and NISQ Noise}\label{sec:ch-hardware}
% ===========================================================================
% Section 9.1  Hardware Constraints and NISQ Noise
% (included from main.tex via \input; heading and \label{sec:ch-hardware}
%  live in main.tex)
% ===========================================================================

\Cref{sec:noise} gave the gate budget, $G\approx L(n+m)\lesssim 1/\epsilon$, and the examples of \cref{fig:fidelity}.
That budget is only the first constraint, and it hides four problems that are specific to graph models.

\paragraph{The graph and the device have different topologies.}
The input graph is arbitrary, whereas the coupling map of a device is fixed; dense or irregular graphs therefore pay a SWAP overhead that inflates depth (\cref{sec:gates}).
Among the designs reviewed, only QuanGCN addresses this explicitly, by sparsifying node connections to reduce the number of crossing gates~\citep{zhou2022quangcn}.
Analog arrays of neutral atoms suggest the opposite strategy, since their programmable geometry realizes graph-shaped interactions directly; the quantum evolution kernel of \citet{henry2021quantum} was proposed for this setting.
A systematic study of hardware-aware graph embedding (ordering, sparsification, partition) and of its effect on accuracy is missing.

\paragraph{Training cost, not circuit cost, is the bottleneck.}
By \cref{eq:epoch-cost}, one epoch with the parameter-shift rule needs $2PKSN$ circuit executions.
As an illustration, a tied model with $P=100$ parameters, one measurement basis, $S=10^3$ shots, and $N=10^4$ graphs needs $2\times10^9$ executions per epoch, more than two days at $10^4$ executions per second, before queueing and compilation overheads.
Parameter tying keeps $P$ small, but the shot and dataset factors remain; this is why almost all results are obtained in simulation (\cref{tab:datasets}).

\paragraph{Noise acts like a node-dependent perturbation.}
Exact permutation equivariance (\cref{prop:qgnn-equiv}) is a statement about noiseless expectation values.
On hardware, error rates differ between qubits and couplers, so relabeling the nodes changes which noisy components are used; the symmetry that underlies the trainability arguments of \cref{sec:sym-perm} holds only approximately.
Noise also flattens the landscape independently of the ansatz~\citep{wang2021noise}.
We found no study that quantifies the loss of equivariance, or the resulting accuracy spread over qubit assignments, for a QGNN on a real device.

\paragraph{Mitigation and correction are not yet in reach for QGNNs.}
Error mitigation reduces bias at a sampling cost that grows exponentially with circuit size in the worst case~\citep{cai2023quantum}, and the largest mitigated experiments concern tasks other than learning~\citep{kim2023evidence}.
Error correction has reached the stage of improving logical error rates below threshold~\citep{acharya2025quantum}, but early fault-tolerant machines are expected to offer few logical qubits with tight limits on connectivity and latency~\citep{palsberg2026computer}.
Hardware evidence for QGNNs is correspondingly thin: a hardware run of a spatial QGCN~\citep{zeng2025spatial} and a five-qubit-class device for QuanGCN~\citep{zhou2022quangcn}.

\paragraph{Open problems.}
(i) Hardware-aware co-design of graph embedding and circuit layout, with reported accuracy-versus-SWAP trade-offs.
(ii) Noise sweeps and hardware runs as a standard part of QGNN evaluation (\cref{tab:protocol}).
(iii) Symmetry-aware error mitigation that preserves or restores equivariance.
(iv) Resource estimates for QGNN layers on a few logical qubits, which are the subject of \cref{sec:fut-ft}.

\subsection{Trainability (Barren Plateaus) at Graph Scale}\label{sec:ch-train}
% ===========================================================================
% Section 9.2  Trainability (Barren Plateaus) at Graph Scale
% (included from main.tex via \input; heading and \label{sec:ch-train}
%  live in main.tex)
% ===========================================================================

\Cref{tab:train} showed that positive trainability evidence for QGNNs comes largely from restricting the circuit.
The challenge is to know whether that evidence survives at the graph scales that matter.

\paragraph{What the evidence covers.}
The numerical studies vary the \emph{qubit count} with synthetic or sampled inputs.
\citet{faria2025inductive} sample random points of the loss landscape for circuits with increasing numbers of qubits, built from synthetic node features, and find sub-exponential variance scaling for correlated and uncorrelated ans\"atze; \citet{lo2026quantum} do the same on randomly sampled subgraphs of the Karate and Wisconsin datasets, observe a gentle decay or a constant variance, and state that analytical results are still required.
On the theoretical side, permutation-equivariant circuits provably avoid plateaus under suitable conditions~\citep{schatzki2024theoretical}, and the in-circuit message-passing model of \citet{raj2026scalable} relies on a Lie-algebra argument.
None of these results varies the \emph{graph}: its size, degree, and regularity.

\paragraph{Graph-structured ans\"atze can have large Lie algebras.}
A recent preprint shows that the dynamical Lie algebra of QAOA for MaxCut grows exponentially for almost all graphs (proved for the unweighted case), a growth associated with barren plateaus~\citep{mao2025qaoa}.
The edge layer of \cref{eq:edgelayer} with a node mixer is exactly this structure, and the QGCNN of \citet{verdon2019quantum} belongs to the same quantum alternating operator family (\cref{sec:ham-variants}).
The result is a warning, not a theorem for QGNNs, because node-feature encodings, additional gates, and local readouts change the algebra; but it shows that tying parameters over edges does not by itself ensure trainability, and that trainability depends on the topology of the input graph.
This agrees with the expectation of \citet{skolik2023equivariant} that plateaus in their weighted-graph circuits depend on connectivity, weights, and features, and it makes highly symmetric graph families, where algebras are small, the places where guarantees are most likely.

\paragraph{Plateaus are not the only obstacle.}
Landscapes can be dominated by poor local minima even when gradients do not vanish~\citep{anschuetz2022beyond}, and training can be NP-hard in the worst case~\citep{bittel2021training}.
Studies that report absence of plateaus therefore say little about final accuracy at scale.

\paragraph{Trainability and classical simulation are coupled.}
Many mechanisms that avoid plateaus also admit classical surrogates~\citep{cerezo2025does,schreiber2023classical}, and the relation between trainability and dequantization of variational learning models is being formalized~\citep{gilfuster2025trainability}.
For QGNNs the same issue appears concretely: the dequantization argument of \citet{lo2026quantum} for quantum graph convolutions rests on low-rank structure.
A QGNN family that is trainable at scale but for which no classical surrogate is known would be a meaningful positive result; none has been demonstrated.

\paragraph{Open problems.}
(i) Graph-dependent analysis of the dynamical Lie algebra for the QGNN families of \cref{sec:taxonomy}, including the encoder and the observables, for bounded-degree and tree-like graph classes.
(ii) Gradient-variance studies that scale the graph, not only the qubit count, and report results per graph family.
(iii) Theory for transfer from small to large graphs, since parameter tying makes pre-training~\citep{raj2026scalable} possible but unexplained.
(iv) Joint reporting of trainability and a classical-surrogate test (\cref{sec:ch-theory}).

\subsection{Scaling to Large Graphs and Qubit--Node Mapping}\label{sec:ch-scale}
% ===========================================================================
% Section 9.3  Scaling to Large Graphs and Qubit--Node Mapping
% (included from main.tex via \input; heading and \label{sec:ch-scale}
%  live in main.tex)
% ===========================================================================

Every way of fitting a large graph into a small device moves the cost somewhere else (\cref{tab:cost-shift}); the open question is whether a quantum advantage can survive the move.

\begin{table}[t]
\centering
\caption{Where the cost goes when a large graph is fitted to a small device.}
\label{tab:cost-shift}
\renewcommand{\arraystretch}{1.3}
\footnotesize
\begin{tabular}{@{}p{3.0cm}p{3.6cm}p{4.9cm}@{}}
\toprule
\textbf{Route} & \textbf{Cost moves to} & \textbf{What remains to be shown} \\
\midrule
Node--qubit mapping & qubits and depth ($n_q=n$) & any graph beyond the budget of \cref{fig:graph-size} \\
Index or amplitude registers & loading and readout; possibly QRAM & speedup net of state preparation and dequantization \\
Sampling, ego-graphs & circuit executions grow with nodes & value of the quantum part relative to the classical sampler \\
Compression, pooling & a classical encoder & how much task-relevant structure is lost \\
In-circuit, size-independent dynamics & classical simulation (so far) & behaviour beyond simulated sizes \\
\bottomrule
\end{tabular}
\end{table}

\paragraph{Index registers and data access.}
Compressed encodings need $\mathcal{O}(\log n)$ qubits, but generic state preparation costs $\mathcal{O}(2^{n_q})$ gates unless a quantum random-access memory is assumed (\cref{sec:vqc}), and the real costs of such a memory are themselves a subject of critique~\citep{jaques2023qram}.
The algorithmic convolutions of \cref{sec:conv-alg}~\citep{liao2024graph,ye2025spectral} inherit this assumption, and low-rank structure, which makes data loading easy, also makes classical simulation easy~\citep{tang2019quantum,lo2026quantum}.

\paragraph{Sampling and compression: scalable, but how quantum?}
Ego-graph and sampled aggregators~\citep{ai2024ego,faria2025inductive} scale because the quantum circuit sees only a local piece.
The total work still grows linearly with the number of nodes: processing every node of Cora (2,708 nodes) with one circuit per node, $P=100$, one basis, and $S=10^3$ shots gives about $5.4\times10^8$ circuit executions per epoch without mini-batching, the same linear scaling as a classical GNN with a much larger constant.
Compression moves part of the learning to a classical encoder~\citep{casals2025guided}, and the information discarded is rarely measured.
In both cases the \emph{fraction} of the computation that is quantum shrinks as the graph grows, which weakens any argument that the quantum part is essential.

\paragraph{Graph-learning bottlenecks are inherited.}
Classical GNNs suffer from over-smoothing and over-squashing (\cref{sec:failure-modes}), and the light-cone argument of \cref{prop:lightcone} shows that an $L$-layer circuit with local gates has an $L$-hop receptive field, just like message passing.
Whether long-range Hamiltonian evolution or entanglement relieves these bottlenecks, or only adds parameters, has not been tested against classical remedies such as rewiring or graph transformers; we found no study that does so.

\paragraph{Open problems.}
(i) Partition and compression schemes that report the information lost and its effect on accuracy.
(ii) Encodings of adjacency that avoid generic state preparation.
(iii) End-to-end fault-tolerant resource estimates (logical qubits, T-count, memory) for the algorithmic layers, instead of asymptotics.
(iv) Tasks on which the quantum part cannot be replaced by classical pre-processing, so that scaling arguments are about the quantum model itself.

\subsection{Fair Comparisons with Classical GNNs and the Absence of Standard Benchmarks}\label{sec:ch-fair}
% ===========================================================================
% Section 9.4  Fair Comparisons with Classical GNNs and the Absence of Standard Benchmarks
% (included from main.tex via \input; heading and \label{sec:ch-fair}
%  live in main.tex)
% ===========================================================================

\Cref{sec:eval} listed good practices; here we ask why they are not yet routine and what would make claims comparable.

\paragraph{Structural obstacles.}
First, the scale gap: classical benchmarks such as the Open Graph Benchmark are large and standardized~\citep{hu2020open}, whereas quantum studies work on subsets and toy graphs (\cref{fig:graph-size}), so the comparison is made where classical models are essentially solved.
Second, resources: ``fewer parameters'' is a statement about $P$, but cost also includes shots, depth, and simulation time (\cref{eq:epoch-cost}); a model with fewer parameters can be far more expensive.
Third, baseline effort: studies of classical graph learning show that conclusions depend on the evaluation protocol and the hyperparameter search~\citep{errica2020fair}, and recent suites such as GraphBench were built because task-specific datasets and inconsistent protocols hinder comparison~\citep{stoll2025graphbench}.
Quantum machine learning has the same problem: simple classical models often outperform quantum ones in large-scale benchmarks~\citep{bowles2024better}, and quantum advantage is not obviously the right criterion for judging such models~\citep{schuld2022quantum}.

\paragraph{Claims and the evidence they need.}
\Cref{tab:claims} lists the claims made in the QGNN literature and what a convincing test would add.

\begin{table}[t]
\centering
\caption{Typical claims in QGNN papers and the evidence needed to support them.}
\label{tab:claims}
\renewcommand{\arraystretch}{1.3}
\footnotesize
\begin{tabular}{@{}p{2.6cm}p{4.3cm}p{5.2cm}@{}}
\toprule
\textbf{Claim} & \textbf{What is typically shown} & \textbf{What is needed} \\
\midrule
Fewer parameters & lower $P$ at similar score & total cost: shots, depth, simulation time; a classical model of equal accuracy \\
Better generalization & lower gap at small data or larger molecules~\citep{faria2025inductive} & learning curves, several seeds, out-of-distribution splits, encoding controls \\
Noise robustness & small circuits; few noisy runs & noise sweeps and hardware runs (\cref{tab:protocol}) \\
Expressivity beyond 1-WL & separation of small synthetic graphs & comparison with classical random-initialization and subgraph GNNs at equal cost \\
Speedup & asymptotic argument, assumes data access & end-to-end cost including loading and readout; dequantization check \\
Advantage & AUC or accuracy above one baseline~\citep{innan2024financial} & strong tuned baselines; several datasets; paired statistics \\
\bottomrule
\end{tabular}
\end{table}

\paragraph{What the field needs.}
A shared benchmark suite with size caps and fixed splits (\cref{sec:repro}) is the first step, but it will not be sufficient on its own: comparisons should be \emph{cost-normalized}, plotting accuracy against total circuit executions or against classical compute, and negative or null results should be published, as the multi-seed outcome of \citet{hqgnn2026representational} shows how much they teach.
The broader debate on how to judge claims of quantum advantage in learning points the same way~\citep{eisert2025mind}.

\subsection{Theoretical Gaps: Expressivity, Generalization Bounds, and Provable Advantage}\label{sec:ch-theory}
% ===========================================================================
% Section 9.5  Theoretical Gaps: Expressivity, Generalization Bounds, and Provable Advantage
% (included from main.tex via \input; heading and \label{sec:ch-theory}
%  live in main.tex)
% ===========================================================================

The theory of QGNNs is far behind their architectures.
The general challenges of quantum machine learning (trainability, data encoding, generalization, and the search for advantage~\citep{cerezo2022challenges}) all apply, and graph structure adds its own questions.

\paragraph{Expressivity.}
Known results are existence statements: universality through the correspondence with message-passing networks with random node initialization~\citep{mernyei2022equivariant,mernyei2023constructions}, a position above 1-WL in the Weisfeiler--Leman hierarchy~\citep{raj2026scalable}, and expressive quantum positional encodings~\citep{thabet2024quantum}, to be read against the classical limits of \cref{sec:equivariance}.
What is missing is a statement at \emph{equal cost}: which graph properties can a QGNN of polynomial size compute that a classical GNN of comparable cost cannot, for example the counting of cycles or other substructures?
Without it, expressivity results show that QGNNs are not weaker than the classical randomized class, not that they are stronger (\cref{sec:universal}).

\paragraph{Generalization.}
Generalization bounds for quantum models typically scale with the number of trainable gates or the encoding~\citep{caro2022generalization}, and standard arguments may fail to explain why trained quantum models generalize~\citep{gilfuster2024understanding}; recent work gives stability-based bounds~\citep{yang2025stability}.
For graphs there are two specific gaps.
Size generalization, from training on small graphs to testing on larger ones, is observed (the sampled QGNN of \citet{faria2025inductive} generalizes better as molecules grow, and \citet{raj2026scalable} pre-train on small graphs) but has no theory for tied-parameter circuits.
Kernel analyses, such as the quantum neural tangent kernel for graph data~\citep{tang2022graphqntk}, offer one analytical route, but they have not been extended to the architectures of \cref{sec:taxonomy}.

\paragraph{Provable advantage.}
Rigorous quantum advantages in learning exist, but for special settings: a speedup under the discrete-logarithm assumption on contrived data~\citep{liu2021rigorous}, and exponential savings when the data come from quantum experiments and the learner has quantum memory~\citep{huang2022quantum}.
For graphs, the clearest advantage is algorithmic and not about learning: the quantum walk on glued trees is exponentially faster than any classical algorithm in the oracle model~\citep{childs2003exponential}.
No advantage is known for a GNN task on classical graph data, and the algorithmic designs that promise one depend on data-access assumptions and face dequantization (\cref{sec:ch-scale}).
In our reading, the most plausible directions are those where the graph or the data are themselves quantum, such as the graph-structured quantum data of \citet{beer2023quantum} and learning Hamiltonian dynamics on a graph with the QGRNN (\cref{sec:ham-variants}), and fault-tolerant spectral methods whose data-access costs are made explicit (\cref{sec:fut-advantage}).

\paragraph{Trainability versus simulability.}
This coupling (\cref{sec:ch-train}) gives a concrete requirement: a claim that a QGNN family is trainable should state whether a classical surrogate exists, in the sense of~\citet{schreiber2023classical}, and cite the demonstrations of efficient surrogates on real devices~\citep{liao2025demonstration}.

\begin{table}[t]
\centering
\caption{Open theoretical questions for QGNNs.}
\label{tab:open}
\renewcommand{\arraystretch}{1.3}
\footnotesize
\begin{tabular}{@{}p{4.6cm}p{4.0cm}p{3.6cm}@{}}
\toprule
\textbf{Question} & \textbf{Why it matters} & \textbf{Possible route} \\
\midrule
Which graph properties are efficiently computable by QGNNs but not by classical GNNs of equal cost? & turns universality into separation & substructure counting; WL-hierarchy arguments \\
How does the dynamical Lie algebra of QGNN layers depend on graph structure? & decides trainability by graph class & extend the QAOA analysis~\citep{mao2025qaoa} to encoders and readouts \\
What are size-generalization bounds for tied-parameter QGNNs? & justifies pre-training on small graphs & stability and kernel methods~\citep{yang2025stability,tang2022graphqntk} \\
Is there a trainable QGNN family with no classical surrogate? & minimal condition for advantage & surrogate construction~\citep{schreiber2023classical} \\
Where do quantum inputs give provable advantage on graphs? & the most plausible regime & quantum-data graphs~\citep{beer2023quantum,huang2022quantum} \\
\bottomrule
\end{tabular}
\end{table}

% ===========================================================================
\section{Future Directions}\label{sec:future}
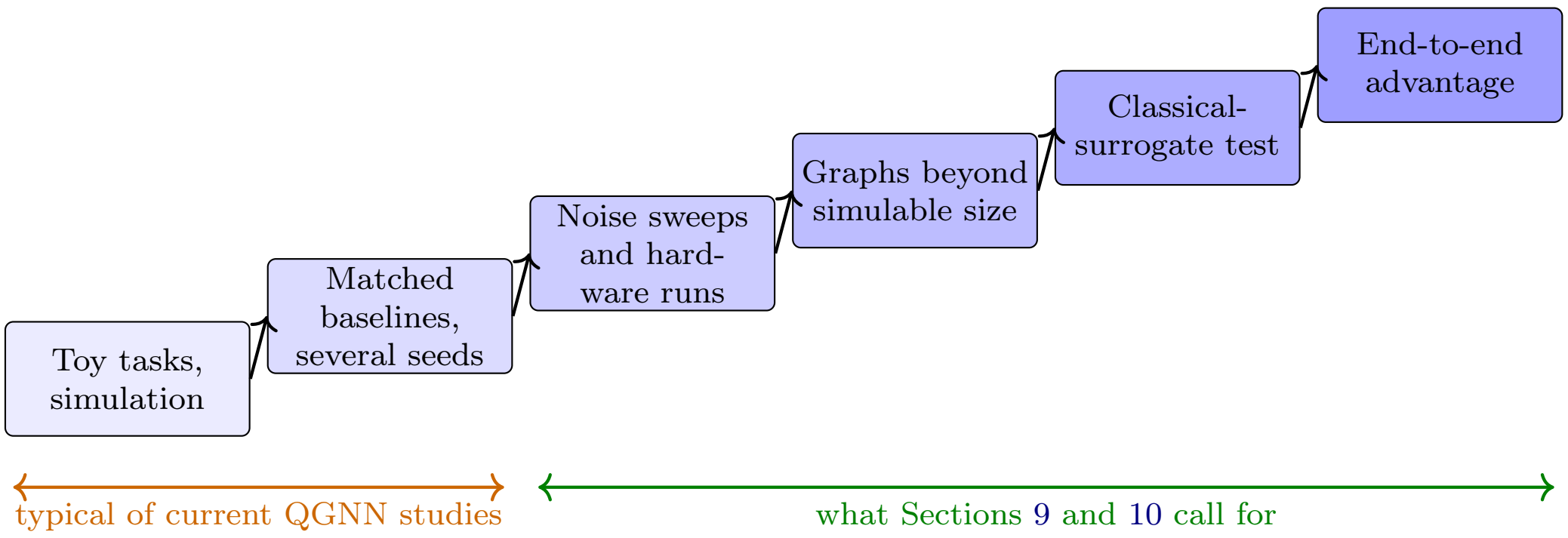
\begin{figure}[t]
\centering
\begin{tikzpicture}[every node/.style={font=\scriptsize},
  st/.style={draw,rounded corners=2pt,text width=2.0cm,align=center,minimum height=1.0cm,inner sep=2pt}]
  \foreach \i/\t/\c in {1/{Toy tasks, simulation}/8,2/{Matched baselines, several seeds}/14,3/{Noise sweeps and hardware runs}/20,4/{Graphs beyond simulable size}/26,5/{Classical-surrogate test}/32,6/{End-to-end advantage}/38}{
    \node[st,fill=blue!\c] (n\i) at ({(\i-1)*2.3},{(\i-1)*0.55}) {\t};}
  \foreach \i/\j in {1/2,2/3,3/4,4/5,5/6}{\draw[->,thick] (n\i.east) -- (n\j.west);}
  \draw[<->,thick,orange!80!black] (-1.0,-0.95) -- (3.3,-0.95);
  \node[orange!80!black,below] at (1.15,-0.95) {typical of current QGNN studies};
  \draw[<->,thick,green!50!black] (3.6,-0.95) -- (12.5,-0.95);
  \node[green!50!black,below] at (8.05,-0.95) {what \cref{sec:challenges,sec:future} call for};
\end{tikzpicture}
\caption{Evidence ladder for claims about QGNNs, from toy simulations to an end-to-end advantage. Each rung adds a test that earlier rungs do not provide.}
\label{fig:evidence-ladder}
\end{figure}

% ===========================================================================
The challenges of \cref{sec:challenges} suggest four directions: finding regimes where an advantage is plausible and provable, extending the symmetry principle that already works, preparing for fault-tolerant hardware, and building the benchmarks and software that make progress measurable.
Across all four, the same standard applies: a result should climb the ladder of \cref{fig:evidence-ladder} before it is called an advantage, and most current QGNN results sit on its lowest rungs.
\Cref{tab:roadmap} gives an indicative roadmap by hardware horizon.

\begin{table}[t]
\centering
\caption{Indicative roadmap for the four directions by hardware horizon (our assessment, based on the works cited in this section).}
\label{tab:roadmap}
\renewcommand{\arraystretch}{1.3}
\footnotesize
\begin{tabular}{@{}p{2.4cm}p{3.7cm}p{3.7cm}p{3.1cm}@{}}
\toprule
\textbf{Direction} & \textbf{Now (NISQ, simulation)} & \textbf{Early fault tolerance} & \textbf{Fault tolerance} \\
\midrule
Advantage & quantum-data and generative tasks on tens of qubits & dynamics learning on graphs & spectral layers with data access \\
Equivariant, geometric & scalar-input and small equivariant circuits & symmetry-preserving mitigation; $E(3)$ layers & large equivariant dynamics \\
Fault tolerance & per-layer resource estimates & Hamiltonian layers on few logical qubits & block-encoded layers \\
Benchmarks, software & shared tiers and reporting protocol & hardware-in-the-loop leaderboards & resource-based end-to-end benchmarks \\
\bottomrule
\end{tabular}
\end{table}

\subsection{Provable Quantum Advantage in Graph Learning}\label{sec:fut-advantage}
% ===========================================================================
% Section 10.1  Provable Quantum Advantage in Graph Learning
% (included from main.tex via \input; heading and \label{sec:fut-advantage}
%  live in main.tex)
% ===========================================================================

\Cref{sec:ch-theory} found no known advantage for GNN tasks on classical graph data.
The constructive question is where to look.
From the requirements collected in \cref{sec:challenges}, a credible advantage claim needs a model that is hard to simulate classically, efficiently trainable, free of hidden data-access costs, and compared with strong classical baselines~\citep{eisert2025mind}.
Recent results show that the first two requirements can coexist.
\citet{huang2025generative} introduce families of generative quantum models that are hard to simulate classically, efficiently trainable, free of barren plateaus and proliferating local minima, and demonstrate them on a 68-qubit superconducting processor; training is classically efficient, but inference requires a quantum computer.
The same ``train classically, deploy on quantum'' idea has been scaled to circuits of a thousand qubits~\citep{recioarmengol2025train}.
These results concern distributions and circuits, not graphs, but they define a template that QGNN research can test.

\begin{table}[t]
\centering
\caption{Candidate regimes for advantage in graph learning, with the main obstacle for each. The first two rows and the last are the most developed.}
\label{tab:fut-adv}
\renewcommand{\arraystretch}{1.3}
\footnotesize
\begin{tabular}{@{}p{3.0cm}p{4.4cm}p{3.3cm}p{2.4cm}@{}}
\toprule
\textbf{Regime} & \textbf{Why advantage is plausible} & \textbf{Support} & \textbf{Obstacle} \\
\midrule
Quantum data on graphs & the learner can use quantum memory; classical learners cannot & \citet{beer2023quantum,huang2022quantum} & graph-structured task with a proof \\
Graph-structured generative models & hard-to-sample, classically trainable circuits & \citet{huang2025generative,recioarmengol2025train} & not yet shown for graph data \\
Dynamics learning on graphs & data generated by quantum dynamics on an interaction graph & QGRNN~\citep{verdon2019quantum} & classical surrogates \\
Fault-tolerant spectral layers & block encodings and phase estimation & \citet{liao2024graph,ye2025spectral} & data access; dequantization \\
Walk-based graph algorithms & exponential speedup in the oracle model & \citet{childs2003exponential} & not a learning problem \\
\bottomrule
\end{tabular}
\end{table}

\paragraph{Quantum data on graphs.}
In the setting of \citet{beer2023quantum}, a quantum state sits on every vertex, and the target is, for instance, an unobserved vertex label.
Advantages in learning from quantum experiments are rigorous when the learner can use quantum memory~\citep{huang2022quantum}, which makes this regime the most defensible, and it matches practical needs such as characterizing noisy devices.
The missing step is a graph-structured task, with a stated correlation structure, for which the separation can be proved.

\paragraph{Graph-structured generative models.}
Circuits whose interaction pattern is given by a graph are natural candidates for the generative template above, since graph states and edge-layer circuits (\cref{eq:graphstate,eq:edgelayer}) are exactly such circuits.
In our reading, a research programme would (i) identify a family of graph-structured circuits that is classically hard to sample from yet trainable by classical methods, (ii) define a graph-generation or graph-property task on which it provably beats classical models, and (iii) demonstrate it on tens of qubits.
We are not aware of this having been done for graph data.

\paragraph{Evidence required.}
Whatever the regime, the claim should be accompanied by the tests of \cref{sec:ch-train,sec:ch-fair}: a statement on classical surrogates, parameter- and cost-matched baselines, and a resource account that includes data loading and readout.
The ladder of \cref{fig:evidence-ladder} makes these steps explicit.

\subsection{Equivariant and Geometric QGNNs as the Next Frontier}\label{sec:fut-equiv}
% ===========================================================================
% Section 10.2  Equivariant and Geometric QGNNs as the Next Frontier
% (included from main.tex via \input; heading and \label{sec:fut-equiv}
%  live in main.tex)
% ===========================================================================

Symmetry is the best-supported idea in the field: it yields circuits that are correct by construction, provable trainability under conditions (\cref{sec:sym-perm}), and the most compact application-specific models (\cref{sec:sym-rot,sec:sym-lorentz}).
We see four ways to extend it.

\paragraph{From relabeling to full geometry.}
The 3D models of \cref{sec:sym-rot} treat rotations together with permutations~\citep{liu2025rotation,park2026hyqurp}, and the Lorentz model of \citet{jahin2025lorentz} works with invariant scalars.
What is missing is equivariant handling of continuous coordinates and of vector- and tensor-valued outputs, such as forces on atoms in molecules, where the output must rotate with the input; the water-molecule model of \citet{piperno2025study} predicts forces and encodes the symmetry of that system through its circuit structure.
Gate-level equivariance for $E(3)$ would also reduce reliance on the invariant-input strategy and its loss of directional information (\cref{tab:sym-compare}).

\paragraph{Beyond pairwise structure.}
Graphs model pairwise relations only.
Quantum simplicial networks apply an Ising-inspired quantum layer to simplicial complexes and are the first quantum topological deep learning model; on synthetic classification tasks they are reported to beat classical simplicial models in accuracy and efficiency~\citep{piperno2025quantum}.
Equivariance for higher-order structures (hypergraphs, cell complexes), its trainability, and real-data evaluation are open.

\paragraph{Equivariance together with scale.}
Exact equivariance in the node--qubit mapping costs $n$ qubits (\cref{tab:tax-compare}), which is the opposite of what scale requires.
In-circuit message passing with size-independent trainable dynamics and a proved WL position~\citep{raj2026scalable} is the first design to aim at both; the open points are behaviour beyond simulated sizes, noise, and the combination with sampling.

\paragraph{Symmetry that survives hardware and has theory beyond $\Sym$.}
Noise breaks equivariance on real devices (\cref{sec:ch-hardware}), so symmetry-preserving error mitigation is needed.
On the theory side, the representation-theoretic toolbox~\citep{larocca2022group,meyer2023exploiting,nguyen2024theory,schatzki2024theoretical} should be extended to the groups above and, following \citet{mao2025qaoa}, made sensitive to the topology of the input graph.
A useful target is a QGNN family with a proof that it is trainable at scale for a stated graph class, which would be both a trainability and a simulability test (\cref{sec:ch-theory}).

\subsection{Fault-Tolerant QGNNs and Error-Corrected Architectures}\label{sec:fut-ft}
% ===========================================================================
% Section 10.3  Fault-Tolerant QGNNs and Error-Corrected Architectures
% (included from main.tex via \input; heading and \label{sec:fut-ft}
%  live in main.tex)
% ===========================================================================

Hardware is moving from physical to logical qubits: error correction has been shown to improve logical error rates as code distance grows~\citep{acharya2025quantum}, and logical processors have been demonstrated on reconfigurable atom arrays~\citep{bluvstein2024logical}.
Early fault-tolerant machines are expected to offer few logical qubits with tight limits on connectivity and control~\citep{katabarwa2024early,palsberg2026computer}.
For QGNNs, the questions are which layers are worth running on such machines, and what they cost.

\paragraph{Candidate layers.}
\Cref{tab:fut-ft} lists the layer types of this survey that could be implemented on logical qubits, with the cost driver of each.
The Hamiltonian family is the best aligned with early fault-tolerant hardware, because time evolution under a graph Hamiltonian is the canonical fault-tolerant workload~\citep{lloyd1996universal}; the algorithmic convolutions of \cref{sec:conv-alg} offer the largest potential gains but also the largest assumptions.

\begin{table}[t]
\centering
\caption{Layer types for fault-tolerant QGNNs and their main cost drivers (our assessment from the cited works).}
\label{tab:fut-ft}
\renewcommand{\arraystretch}{1.3}
\footnotesize
\begin{tabular}{@{}p{3.3cm}p{3.2cm}p{4.3cm}p{2.3cm}@{}}
\toprule
\textbf{Layer} & \textbf{Primitive} & \textbf{Dominant cost} & \textbf{Status} \\
\midrule
Block-encoded convolution, attention, message passing & block encoding of graph matrices & data access and normalization~\citep{liao2024graph} & complexity analysis \\
Spectral convolution & phase estimation and swap test & phase-estimation depth; data loading~\citep{ye2025spectral} & no numerics \\
Hamiltonian-evolution layer & product formulas, qubitization & evolution time; gates growing with $n+m$ & small simulations \\
Variational layer on logical qubits & parametrized rotations & logical-gate and magic-state overhead & not studied for QGNNs \\
\bottomrule
\end{tabular}
\end{table}

\paragraph{Data access is the first bottleneck.}
For dense graph data, the resources to block-encode a classical matrix are large: even optimistic constructions that reach polylogarithmic depth need on the order of $N^2$ parallel T gates, and a quantum random-access memory is seen as the bottleneck of fault-tolerant implementations~\citep{clader2022quantum}; the cost of QRAM may erase asymptotic gains~\citep{jaques2023qram}.
For graphs given implicitly (lattices, or molecules whose bonds follow from geometry) or sparse graphs with efficient row access, cheaper oracles may exist, which is why the graph class should be part of any resource statement; we regard this as the most promising route, but it has not been worked out for the QGNN layers above.

\paragraph{Resource estimates are the missing deliverable.}
End-to-end analyses, from input to output, have been done for quantum interior-point methods~\citep{dalzell2023endtoend} and for linear-system solvers including energy~\citep{tu2025towards}; no equivalent exists for a QGNN layer.
Such an analysis should report logical qubits, T-count, and loading and readout costs as functions of $n$, $m$, maximum degree, and precision, and compare them with a classical GNN of equal accuracy.

\paragraph{Open problems.}
(i) End-to-end estimates for the layers of \cref{tab:fut-ft}.
(ii) Variational and Hamiltonian layers designed for logical qubits, including their symmetry (\cref{sec:fut-equiv}).
(iii) Hybrid pipelines in which a classical model handles data access and a small fault-tolerant core handles the quantum step.

\subsection{Cross-Domain Benchmarks and Unified Software Frameworks}\label{sec:fut-bench}
% ===========================================================================
% Section 10.4  Cross-Domain Benchmarks and Unified Software Frameworks
% (included from main.tex via \input; heading and \label{sec:fut-bench}
%  live in main.tex)
% ===========================================================================

\Cref{sec:datasets} and \cref{sec:ch-fair} showed that QGNN claims are hard to compare.
Fixing this needs a benchmark and a software layer designed together, because the benchmark must report resources that only the framework can measure.
\Cref{tab:fut-bench} lists the components and what exists today.

\begin{table}[t]
\centering
\caption{Components of a unified QGNN benchmark and framework, existing resources, and gaps.}
\label{tab:fut-bench}
\renewcommand{\arraystretch}{1.3}
\footnotesize
\begin{tabular}{@{}p{2.6cm}p{5.0cm}p{5.2cm}@{}}
\toprule
\textbf{Component} & \textbf{Existing resource} & \textbf{Gap} \\
\midrule
Graph data & standardized classical benchmarks with fixed splits~\citep{hu2020open,stoll2025graphbench}; QM9 through PyTorch Geometric~\citep{fey2019fast} & size tiers matched to qubit budgets; shared subsets \\
Expressivity & BREC pairs, used by \citet{raj2026scalable} & comparison at equal cost with classical randomized and subgraph GNNs \\
Quantum-ML tasks & curated QML benchmark datasets~\citep{bowles2024better}, hosted in PennyLane & graph tasks and graph-specific baselines \\
Protocol and statistics & minimum reporting protocol (\cref{tab:protocol}); paired tests in the classical suites & adoption; cost-normalized leaderboards \\
Software & differentiable circuit libraries~\citep{bergholm2018pennylane} paired with graph libraries; the release of \citet{raj2026scalable} with a vectorized PyTorch backend and a PennyLane reference backend & a common QGNN layer interface, automatic resource counters, and hardware back-ends \\
\bottomrule
\end{tabular}
\end{table}

\paragraph{A layered benchmark.}
We suggest tiers that follow the scales at which QGNNs are credible: (i) tiny exact tasks for expressivity, with BREC and CFI pairs; (ii) molecular tasks with a size cap, for instance QM9 subsets; (iii) a high-energy-physics task; (iv) a citation or sampled-neighbourhood task with a fixed sampling protocol.
Every entry reports the quantities of \cref{tab:protocol} and plots accuracy against total circuit executions (\cref{eq:epoch-cost}) and against classical compute, so that parity at lower cost and parity at higher cost are distinguishable.

\paragraph{A unified framework.}
The unified definition of \cref{def:qgnn} suggests an interface with four swappable parts: encoder, graph-conditioned circuit, observables, and classical head.
A framework built on it could count qubits, depth, shots, and executions automatically, run the same model on a statevector simulator, a noisy simulator, and hardware, and expose the symmetry checks of \cref{def:qequiv} as tests.
The dual-backend design of \citet{raj2026scalable} is an example of the first steps in this direction.

\paragraph{Governance.}
Benchmarks degrade when they are tuned against; held-out test tiers, versioned splits, and a public record of negative results (\cref{sec:ch-fair}) are as important as the datasets themselves.

% ===========================================================================
\section{Conclusion}\label{sec:conclusion}
% ===========================================================================
% ===========================================================================
% Section 11  Conclusion
% (included from main.tex via \input; heading and \label{sec:conclusion}
%  live in main.tex)
% ===========================================================================

This survey reviewed quantum graph neural networks: models that process graph-structured data with parametrized quantum circuits.
We started from a unified definition (\cref{def:qgnn}) that describes every architecture as an encoder, a graph-conditioned circuit, observables, and an optional classical head, and used it to trace the field from the Hamiltonian-based model of \citet{verdon2019quantum} and its quantum-inspired, circuit-based, and quantum-data contemporaries, through a taxonomy of five architecture families, to symmetry, trainability, scalability, applications, and evaluation.
The overall picture is of a young field with a rich design space and well-chosen ideas, but with evidence that is still thin.

\paragraph{What the survey established.}
First, the design space is coupled (\cref{tab:tax-compare}): exact permutation equivariance comes with the node--qubit mapping and its linear qubit cost, scalable designs obtain size-independent circuits by treating the graph classically and giving up global entanglement, and hybrid designs, which keep the quantum part small, dominate practice.
Second, symmetry is the best-developed inductive bias: equivariant circuits are correct by construction~\citep{mernyei2022equivariant}, they carry the clearest trainability guarantees~\citep{schatzki2024theoretical}, and they have been extended to rotations and the Lorentz group.
Third, theory consists of existence statements, namely universality, positions in the Weisfeiler--Leman hierarchy, and trainability for restricted circuits, and the restrictions that make training feasible also raise the question of classical simulability~\citep{cerezo2025does}; the Lie-algebra analysis of QAOA-type graph layers shows that trainability depends on the topology of the input graph~\citep{mao2025qaoa}.
Fourth, the hardware budget is tight: with $L$ layers and error rate $\epsilon$, a node--qubit model fits graphs with $n+m\lesssim 1/(\epsilon L)$, which is tens to about a hundred nodes today, so larger graphs require sampling or compression, which shifts cost rather than removing it (\cref{tab:cost-shift}).

\paragraph{State of the evidence.}
QGNNs have been applied in high-energy physics, chemistry and materials, finance, healthcare, earth observation, transportation, genomics, and IoT (\cref{tab:apps}).
Reported outcomes are parity with classical models or parameter efficiency, almost always in simulation, on small graphs or subsets of standard datasets, and often against a single baseline.
Hardware demonstrations are rare, no known result shows an advantage on a GNN task over classical graph data, and multi-seed evaluation can reverse an apparent gain~\citep{hqgnn2026representational}.
The binding constraint of the field is therefore not the supply of architectures but the quality of evidence.

\begin{table}[t]
\centering
\caption{Questions addressed in this survey and the short answers.}
\label{tab:takeaways}
\renewcommand{\arraystretch}{1.3}
\footnotesize
\begin{tabular}{@{}p{3.6cm}p{8.1cm}p{1.8cm}@{}}
\toprule
\textbf{Question} & \textbf{Answer} & \textbf{Where} \\
\midrule
What is a QGNN? & encoder, graph-conditioned circuit, observables, optional classical head; equivariance as a circuit-level condition & \cref{sec:notation} \\
How are QGNNs organized? & five families: convolutional, equivariant, message-passing/ego-graph/recurrent, Hamiltonian, hybrid & \cref{sec:taxonomy} \\
Which inductive bias works best? & symmetry, from permutations to rotations and Lorentz transformations & \cref{sec:symmetry} \\
Can they be trained, scaled, and run? & yes on small graphs in simulation; limited by noise, graph-dependent trainability, and qubit counts & \cref{sec:expr-train-scale} \\
Where are they used, and how good is the evidence? & many domains; parity or parameter efficiency, mostly simulated, few baselines & \cref{sec:applications,sec:datasets} \\
Is there a quantum advantage? & not yet shown for classical graph data; quantum-data and generative settings are most plausible & \cref{sec:ch-theory,sec:fut-advantage} \\
\bottomrule
\end{tabular}
\end{table}

\paragraph{Recommendations.}
For authors, we recommend reporting resources alongside accuracy, parameter- and cost-matched baselines, several seeds, noise sweeps or hardware runs, and a statement on classical surrogates (\cref{tab:protocol}).
For the community, we recommend a shared, size-aware benchmark with fixed splits and a software layer that counts resources automatically (\cref{sec:fut-bench}).
For theory, the priorities are separations from classical GNNs at equal cost, graph-dependent trainability analysis, and bounds for generalization from small to large graphs (\cref{sec:ch-theory}).
For applications, the most promising directions are tasks where the data or the graph is itself quantum, generative models that can be trained classically and deployed on quantum hardware, Hamiltonian layers for early fault-tolerant machines, and the extension of symmetry to full geometry and higher-order structure (\cref{sec:future}).

\paragraph{Outlook.}
At present, QGNNs are best understood as a framework for studying how graph structure and symmetry can be built into quantum models, not as a demonstrated alternative to classical GNNs.
Whether they become the latter depends on climbing the evidence ladder of \cref{fig:evidence-ladder}: hardware runs, graphs beyond simulable size, tests against classical surrogates, and, ultimately, an end-to-end advantage on a task that matters.
We hope that the unified formalism, taxonomy, and reporting protocol of this survey help the community make that progress measurable.

% ===========================================================================
\section*{Declarations}
% ===========================================================================

\paragraph{Funding.}
This research received no specific grant from any funding agency in the public, commercial, or not-for-profit sectors. 

\paragraph{Competing Interests.}
The authors declare that they have no known competing financial interests or personal relationships that could have appeared to influence the work reported in this paper.

\paragraph{Data Availability.}
Data sharing is not applicable to this article as no new datasets were generated or analyzed during this survey study.

\paragraph{Author Contributions.}
\textbf{Z. Sharifi:} Conceptualization, Methodology, Investigation, Visualization, Writing -- Original Draft, and Writing -- Review \& Editing. \textbf{H. Shirazi:} Supervision.

\bibliographystyle{unsrtnat}
\bibliography{references}

\end{document}